\pdfoutput=1

\documentclass[lettersize,journal]{IEEEtran}
\usepackage{amsmath,amsfonts}
\usepackage{algorithmic}
\usepackage{algorithm}
\usepackage{array}
\usepackage[caption=false,font=normalsize,labelfont=sf,textfont=sf]{subfig}
\usepackage{textcomp}
\usepackage{stfloats}
\usepackage{url}
\usepackage{verbatim}
\usepackage{graphicx}
\usepackage{cite}
\usepackage{multirow}
\graphicspath{{Figures/}}
\usepackage{url}
\usepackage{xcolor}

\begin{document}

\title{A Deeper Dive into the Irreversibility of PolyProtect:\\ Making Protected Face Templates Harder to Invert}

\author{Vedrana~Krivoku\'ca~Hahn, J\'er\'emy Maceiras, S\'ebastien~Marcel,~\IEEEmembership{Senior Member,~IEEE}
\thanks{All authors are with Idiap Research Institute in Martigny, Switzerland.}}

\maketitle

\begin{abstract}
This work presents a deeper analysis of the ``irreversibility'' property of PolyProtect, a biometric template protection method initially proposed for securing face embeddings.  PolyProtect transforms embeddings into protected templates via multivariate polynomials, whose coefficients and exponents are distinct for each subject enrolled in the face recognition system.  A polynomial is applied to consecutive sets of elements from a given embedding, where the amount of overlap between the sets is a tunable parameter.  We begin our irreversibility analysis by demonstrating that PolyProtected templates are easier to invert using a numerical solver based on cosine distance, as opposed to Euclidean distance (used in the earlier PolyProtect work).  To make this inversion more difficult, we then propose a ``key selection algorithm'', which tries to choose ``keys'' (coefficients and exponents of the PolyProtect polynomial) that enhance the irreversibility of PolyProtected templates, compared to when the keys are purely random.  Our experiments show that this algorithm is effective at generating PolyProtected templates that are significantly more difficult to invert, and that it approximately equalises the irreversibility of PolyProtected templates generated using different ``overlap'' parameters.  This allows for better control of the known irreversibility versus accuracy trade-off across different overlaps.  We also show that accuracy in the PolyProtected domain can be affected by the range in which the embedding elements lie, but that this can be improved by normalizing the embeddings prior to applying PolyProtect.  This work is reproducible using our open-source code: \url{https://github.com/idiap/BTPBench/blob/main/paper/polyprotect-irreversibility/}.        
\end{abstract}


\section{Introduction}

\IEEEPARstart{O}{ur} faces are becoming indispensable tools for proving our identities in various applications (e.g., unlocking smartphones, accessing bank accounts, verifying passports at electronic gates).  The convenience of this form of authentication is evident, as is the heightened security and identity assurance offered by the uniqueness of our faces.  There is, however, a hidden cost: our privacy.  As more and more organisations collect our face data for such purposes, concerns arise around how this data is being stored and whether (and with whom) it is being shared.  The problem is that, the more our face data gets distributed, the more likely it is to be misused; e.g., by creating presentation attacks or deepfakes to impersonate us and potentially gain unlawful access to protected resources, or by tracking us (for profiling or stalking purposes) across different applications in which the same data is enrolled.  So, it is crucial that face data never be stored in the clear: it should always be protected, such that, if the databases of the underlying face recognition systems are jeopardised, the original, irreplaceable face information remains inaccessible.  This is the realm of Biometric Template Protection (BTP).  

The aim of BTP is to convert a biometric feature vector, or ``template'', into a protected template, from which it is impossible to recover the original one.  In this paper, we are interested in the application of BTP to face templates.  Modern face recognition systems are based on deep learning architectures that are trained to map face images to fixed-length numerical representations called ``embeddings''.  It has been shown that face embeddings are invertible \cite{bg26}, in that they can be used to recover an approximation of the underlying face image \cite{z16, c17, m19, s22}, and that certain soft biometric attributes (e.g., sex, race, age, hair colour) can be extracted from these representations \cite{f20, t20}.  These findings underline the need to protect embeddings in order to prevent recovery of the original face information, thereby protecting the privacy of the face recognition system users.  This is where BTP comes into play.  A recent survey of face BTP methods \cite{h22}, as well as Part II of the new \textit{Handbook of Biometric Template Protection} \cite{h26}, revealed two broad approaches: applying handcrafted (human-designed) BTP methods to learned face embeddings, and using neural networks to learn the BTP algorithms.

Handcrafted BTP methods include: feature transformations \cite{d26}, which involve transforming a biometric template from its original feature space to a new, protected space (e.g., \cite{d19, kh22}); biometric cryptosystems \cite{r26}, which most commonly involve binding an external key with the biometric template (e.g., \cite{g19, r22}); and homomorphic encryption \cite{b26}, which allows us to compare reference and probe templates in encrypted form (e.g., \cite{m17, b18, e22}).  Learned BTP methods \cite{kh26} include: training a neural network to learn the mapping from a biometric template to a pre-defined random code (e.g., \cite{p16, j18}); or training a neural network to learn its own representation of a protected template (e.g., \cite{t19, p21, m21}).  While learned BTP methods may allow for higher complexity and thus potentially the generation of more secure protected templates, handcrafted methods have two chief advantages that make them easier to adopt in practice: (i) they can more readily be integrated into existing biometric systems (e.g., as a module after the feature extractor), and (ii) they tend to be easier to evaluate, since the algorithms are designed by humans (i.e., not learned by neural networks) so their properties are more explicit.  For these reasons, in our work we adopt a \textit{handcrafted} BTP method.

The context for this work is defined by a project that aims to use BTP to develop a privacy-preserving face identification system for humanitarian aid distribution.  This is similar to the context in \cite{s25}, except that the end goal of our project is different.  Nevertheless, as the global target in both cases is a humanitarian use-case, the criteria for selecting a suitable BTP method were similar.  In particular, for our work, the chosen BTP method should possess the following characteristics: easy and lightweight to implement, modular (can be integrated into an existing face recognition system), satisfies the main BTP criteria (recognition accuracy, irreversibility, unlinkability), and ideally comes with open-source code.  So, similarly to the reasoning outlined in \cite{s25}, these requirements narrowed our selection to \textit{handcrafted}, as opposed to learned, BTP methods.  Then, within this category, biometric cryptosystems were eliminated due to known accuracy issues and envisaged implementation difficulties in our use-case, and homomorphic encryption was deemed unsuitable due to its computational complexity and the need to safely store the decryption key (which cannot be guaranteed in a volatile humanitarian setting).  So, we settled on the \textit{feature transformations} category, in which the PolyProtect method \cite{kh22} was found to most closely align with our project requirements. 

\begin{figure}[!h]
\centering
\includegraphics[width=0.8\columnwidth]{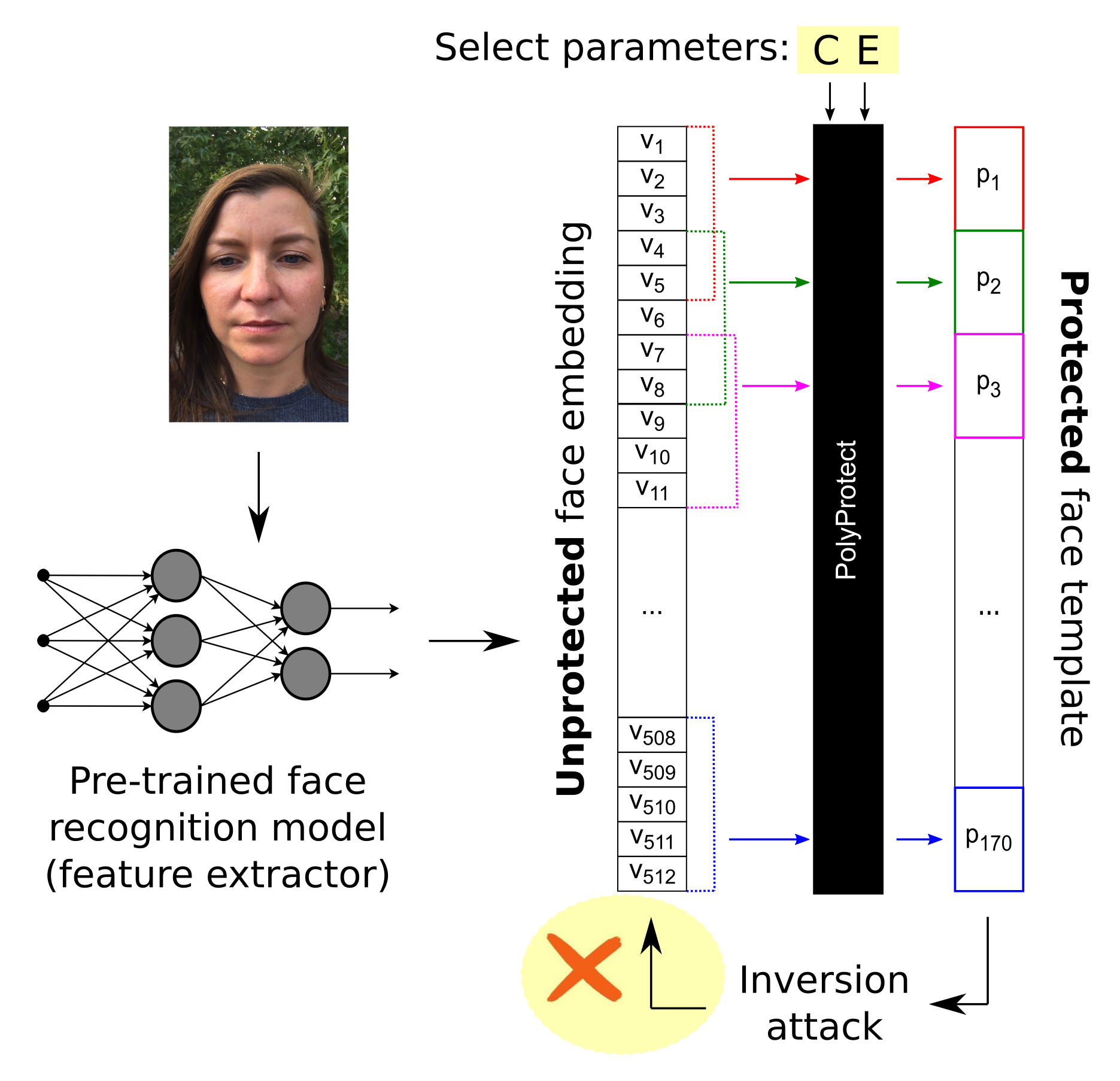}%
\caption{The focus of this work (highlighted in yellow) is on enhancing the irreversibility of PolyProtected templates generated from face embeddings, by selecting the polynomial coefficients, $C$, and exponents, $E$, such that it is more difficult (ideally impossible) to invert a PolyProtected template to recover the unprotected embedding.  (Source of face image: SOTERIA \cite{r24}.)}
\label{fig:teaser}
\end{figure}

PolyProtect transforms face embeddings into protected templates via multivariate polynomials, whose parameters (coefficients, $C$, and exponents, $E$) are unique for each subject enrolled in the face recognition system.  A polynomial is applied to consecutive sets of elements from a given embedding, where the amount of overlap between the sets is a tunable parameter.  PolyProtect has already been shown \cite{kh22, s25} to satisfy the three crucial BTP criteria: recognition accuracy, irreversibility, and unlinkability.  Our goal is to build on this evaluation, with a deeper dive into the ``irreversibility'' criterion.  Our ultimate aim is to make PolyProtected templates more difficult to invert, by selecting the $C$ and $E$ parameters in a smarter (i.e., not purely random) way.  This focus is illustrated in Fig. \ref{fig:teaser}.  

Concretely, our contributions are the following:

\begin{itemize}
	\item Firstly, we evaluate the recognition accuracy attainable by the protected templates when PolyProtect is applied to face embeddings generated by two different face recognition models: iResNet100 and EdgeFace.  We show that, although the two sets of embeddings have comparable accuracy in the unprotected domain, the iResNet100 embeddings result in significantly worse accuracy in the protected domain.  We attribute this to the larger range in which the iResNet100 embedding elements lie, which causes greater intra-class variance when the PolyProtect transform is applied to these embeddings (compared to those generated by the EdgeFace model).  We then propose a simple fix: normalizing the embeddings prior to applying the PolyProtect transform.  This considerably improves the accuracy of the protected templates, making PolyProtect applicable to embeddings in any range.  These findings have an important impact on the suitability of PolyProtect (and potentially other BTP methods) for protecting face embeddings, yet this is the first time they are being reported (to the best of our knowledge).
	\item Secondly, we launch into our analysis of the irreversibility of PolyProtect by challenging the definition of the worst-case attacker from the original work \cite{kh22}.  Specifically, we investigate an alternative numerical solver for recovering a face embedding from its PolyProtected template: one based on cosine distance, as opposed to the original solver based on Euclidean distance.  We show that the cosine-based solver is the more effective inversion tool, so it should replace the Euclidean-based solver in our definition of the capabilities of the worst-case attacker.\footnote{In this paper, ``worst-case attack(er)'' is used to refer to the most powerful type of protected template inversion attack(er) discovered \textit{thus far}.  As we cannot evaluate every possible attack to determine the \textit{absolute} worst-case attack(er) (which, in any case, is likely to be context-dependent), we assume a flexible definition that is subject to updates if stronger attacks are discovered.}  This type of investigation, currently absent from the literature, represents a valuable contribution to the BTP community, because it provides important insight into PolyProtect's template protection potential and encourages similar studies to be conducted on other BTP methods.   
	\item Thirdly, we propose a promising method for significantly improving the irreversibility of PolyProtect against our new, more powerful worst-case attacker.  The method tries to select ``keys'' (i.e., coefficients and exponents of the PolyProtect polynomial) that are more likely (than random keys) to generate ``irreversible'' protected templates.  Use of this method results in an important additional effect: the irreversibility of PolyProtected templates generated using different ``overlap'' parameters is approximately equalised.  This gives us more control over the irreversibility/accuracy trade-off that currently exists across different overlaps \cite{kh22, s25}, by allowing us to tune the overlap to obtain the best possible recognition accuracy while achieving a high (and relatively stable) degree of irreversibility regardless of the selected overlap.
\end{itemize}
          
The remainder of this paper is structured as follows.  Section \ref{sec:set_up} details our experimental set-up, including a description of the PolyProtect BTP method, and the face recognition models and datasets used for the PolyProtect evaluation.  Section \ref{sec:accuracy_eval} evaluates PolyProtect's recognition accuracy, which is an important first step for assessing its practical utility.  Section \ref{sec:irreversibility_eval} then dives into the irreversibility analysis, which focuses on establishing the feasibility of recovering a face embedding from its PolyProtected template using two different numerical solvers.  Section \ref{sec:key_selection} builds on the analysis from Section \ref{sec:irreversibility_eval} by presenting a new key selection algorithm that significantly improves the irreversibility of PolyProtect and helps to effectively balance the known irreversibility versus accuracy trade-off across different overlaps.  Finally, Section \ref{sec:conclusion} concludes this work and presents potential future directions.

\section{Experimental Set-up}
\label{sec:set_up}

This section starts with a brief description of PolyProtect, the BTP method being studied in this work, in Section \ref{subsec:polyprotect}.  Section \ref{subsec:models_datasets} then details the face recognition models selected for the generation of face embeddings, to which PolyProtect will be applied, and the face datasets that will serve as the image sources for these embeddings.

\subsection{PolyProtect: The selected BTP method}
\label{subsec:polyprotect}

PolyProtect, proposed in \cite{kh22} as a BTP method for face embeddings, works as follows.  Let $V = [v_1, v_2, ..., v_n]$ denote an $n$-dimensional face embedding. PolyProtect transforms $V$ into another, lower-dimensional feature vector, $P = [p_1, p_2, ..., p_k]$ (where $k < n$), which is the protected version of $V$.  This is achieved by mapping sets of $m$ (where $m << n$) consecutive elements from $V$ to single elements in $P$ via multivariate polynomials defined by $m$ subject-specific (i.e., distinct for each subject enrolled in the face recognition system) coefficients, $C = [c_1, c_2, ..., c_m]$, and exponents, $E = [e_1, e_2, ..., e_m]$.  

The first $m$ elements in $V$ (i.e., $v_1, v_2, ..., v_m$) are transformed into the first element in $P$ (i.e., $p_1$) via Eq. (1):

\begin{equation}
p_1 = c_{1}v_{1}^{e_1} + c_{2}v_{2}^{e_2} + ... + c_{m}v_{m}^{e_m}
\end{equation}

The elements of $V$ used to generate $p_2$ depend on the chosen amount of \textit{overlap} between successive sets of elements. The minimum overlap is 0, in which case the elements of $V$ in each set would be unique, and the maximum is $m - 1$, in which case successive sets would share $m - 1$ elements. Eqs. (2) and (3) define the mapping from $V$ to $p_2$ for overlaps of 0 and $m - 1$, respectively:

\begin{equation}
p_2 = c_{1}v_{m+1}^{e_1} + c_{2}v_{m+2}^{e_2} + ... + c_{m}v_{m+m}^{e_m}	
\end{equation}

\begin{equation}
p_2 = c_{1}v_{2}^{e_1} + c_{2}v_{3}^{e_2} + ... + c_{m}v_{m+1}^{e_m}	
\end{equation}

The remaining elements in $P$ (i.e., $p_3, ..., p_k$) are generated in a similar manner, until all the elements in $V$ have been used up. If the last set of elements is incomplete because the dimensionality of $V$ is not divisible by the required number of sets (defined by $m$ and the amount of overlap), $V$ is padded by a sufficient number of zeros to complete the last set.

\begin{figure*}[!h]
\centering
\includegraphics[width=0.9\textwidth]{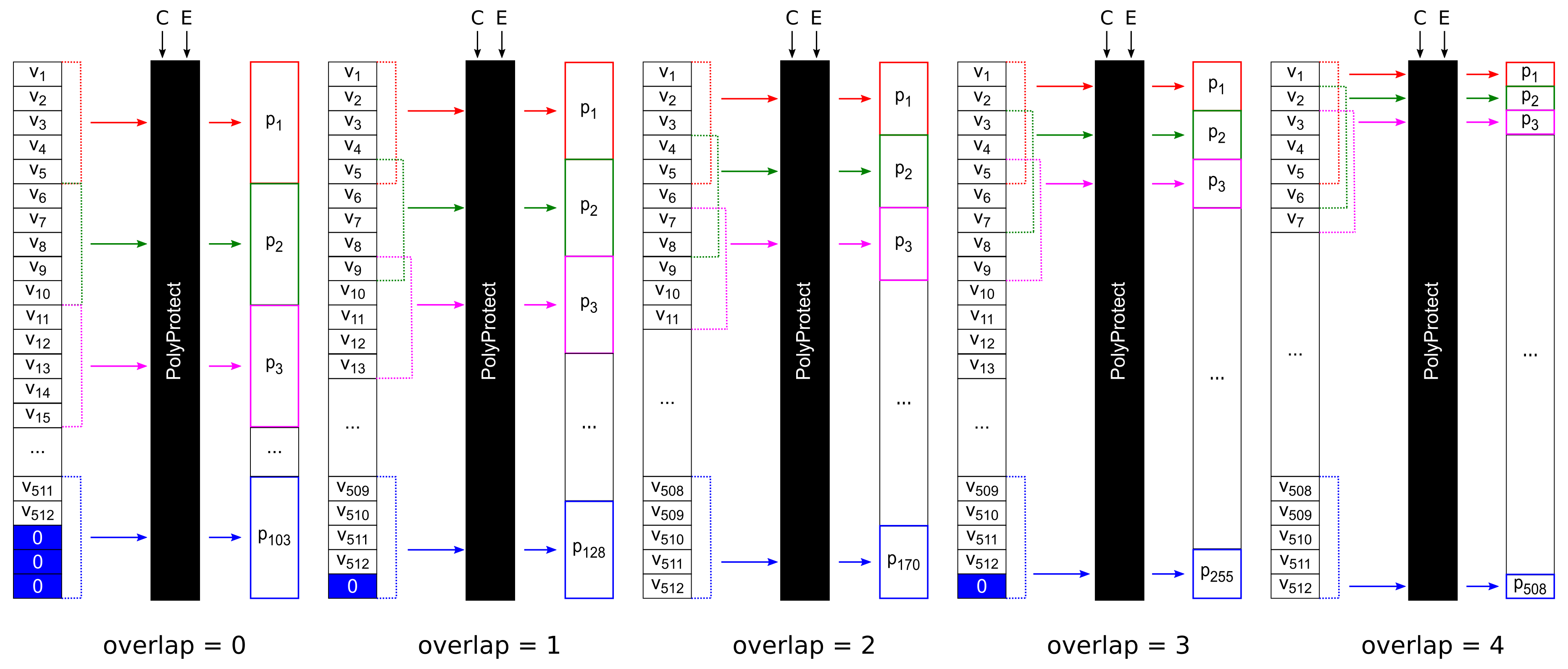}%
\caption{Mapping 512-dimensional $V$ to $P$ via PolyProtect, using $C = [c_1, c_2, ..., c_5]$ and $E = [e_1, e_2, ..., e_5]$, for different amounts of overlap.}
\label{fig:polyprotect}
\end{figure*}

Since the analysis presented in this paper will be based on 512-dimensional face embeddings (see Section \ref{subsec:models_datasets}), Fig. \ref{fig:polyprotect} illustrates the transformation from a 512-dimensional $V$ to $P$, for overlaps 0 -- 4, when $m = 5$.  It is evident that the dimensionality of $P$ is influenced by the amount of overlap used in the $V \rightarrow P$ mapping, i.e., larger overlap $\rightarrow$ larger $P$.  This has been shown to have an effect on the recognition accuracy and irreversibility properties of PolyProtect \cite{kh22, s25}, which will be further investigated in this paper.

For the PolyProtect evaluations presented in the next sections, we chose $m = 5$, as in \cite{kh22}, meaning that each element in $P$ was generated using 5 consecutive elements from $V$, as illustrated for different overlaps in Fig. \ref{fig:polyprotect}.  As explained in \cite{kh22}, this setting was inspired by the Abel-Ruffini theorem, which states that there is no closed-form algebraic expression for solving polynomials of degree 5 or higher with arbitrary coefficients.  Furthermore, as in \cite{kh22}, we also chose not to set $m > 5$, since this would require using exponents larger than 5 in the PolyProtect transform, which may obliterate small embedding elements.  Consequently, the exponents, $E$, were randomly generated, unique integers in the range [1, 5].  Regarding the choice of values for $C$, we used the [-50, 50] range as in \cite{kh22}, so all sets of $C$s consisted of 5 randomly generated, unique, non-zero integers in this range.

\subsection{Selected face recognition models and datasets}
\label{subsec:models_datasets}

Recall that the aim of this paper is to dig deeper into the irreversibility of PolyProtect, and the context is a project focusing on the protection of face templates.  So, in order to perform this analysis, the first step was to generate the face templates to which PolyProtect would be applied.  Since modern face recognition systems employ neural-network-based models for the extraction of face templates (``embeddings''), these were the types of feature extractors we were interested in.  To select the best face recognition model(s) for our study, we evaluated five state-of-the-art models to which we have open-source access: iResNet50 and iResNet100\footnote{From \url{https://github.com/deepinsight/insightface/tree/master/model_zoo}, converted to PyTorch via \url{https://github.com/nizhib/pytorch-insightface}} \cite{h16, dg19}, EdgeFace and EdgeFace-XS\footnote{\url{https://github.com/otroshi/edgeface}} \cite{g24}, and FaceNet\footnote{\url{https://github.com/timesler/facenet-pytorch}} \cite{s15}.  All five models generate 512-dimensional face embeddings.  These models were applied to three different face datasets, to extract the face embeddings from the underlying face images.  The three face datasets used in this study were selected to represent three different image acquisition scenarios.  They include:

\begin{itemize}
	\item Multi-PIE\footnote{\url{https://www.cs.cmu.edu/afs/cs/project/PIE/MultiPie}} \cite{g10}:  Contains face images of 337 subjects.  The images were captured in a very controlled environment, using multiple cameras fixed at different angles.  We randomly selected 10 images per subject across the 3 frontal cameras (14\_0, 05\_1, and 05\_0), which resulted in a total of 3,370 face images.
	\item SOTERIA\footnote{\url{https://www.idiap.ch/en/scientific-research/data/soteria}} \cite{r24}: Contains face videos of 70 subjects.  The (bona-fide) videos were captured in a less controlled environment than Multi-PIE, using the frontal (``selfie'') and main (back) cameras of five different mobile phones (Apple iPhones 6s and 12, Xiaomi Redmi 6 Pro and 9A, and Samsung Galaxy S9), under various lighting conditions.  We randomly selected 10 frontal frames per subject across the five phones, which resulted in a total of 700 face images.
	\item iCarB-Face\footnote{\url{https://www.idiap.ch/en/scientific-research/data/icarb-face}} \cite{kh24}: Contains face videos of 197 subjects.  The videos were captured inside a car, using a near-infrared camera, while the subjects were seated in the driver's seat.  We selected 4 video frames per subject, when the subject wore a neutral facial expression and no accessories: 2 when the car was parked indoors and the other 2 when it was parked outdoors, with the subject's eyes being open in one image and closed in the other.  This resulted in a total of 788 face images.
\end{itemize}  

\begin{figure}[!h]
\centering
\subfloat{\includegraphics[width=0.33\columnwidth]{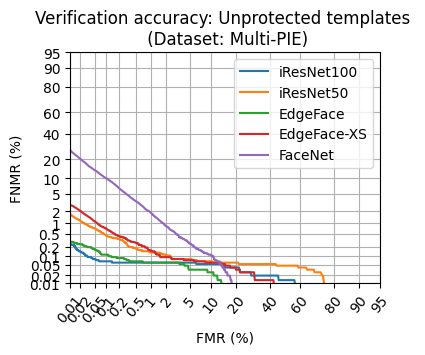}}
\hfil
\subfloat{\includegraphics[width=0.33\columnwidth]{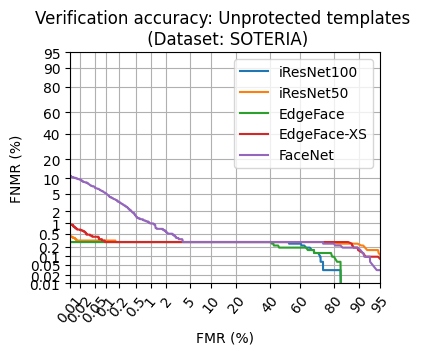}}
\hfil
\subfloat{\includegraphics[width=0.33\columnwidth]{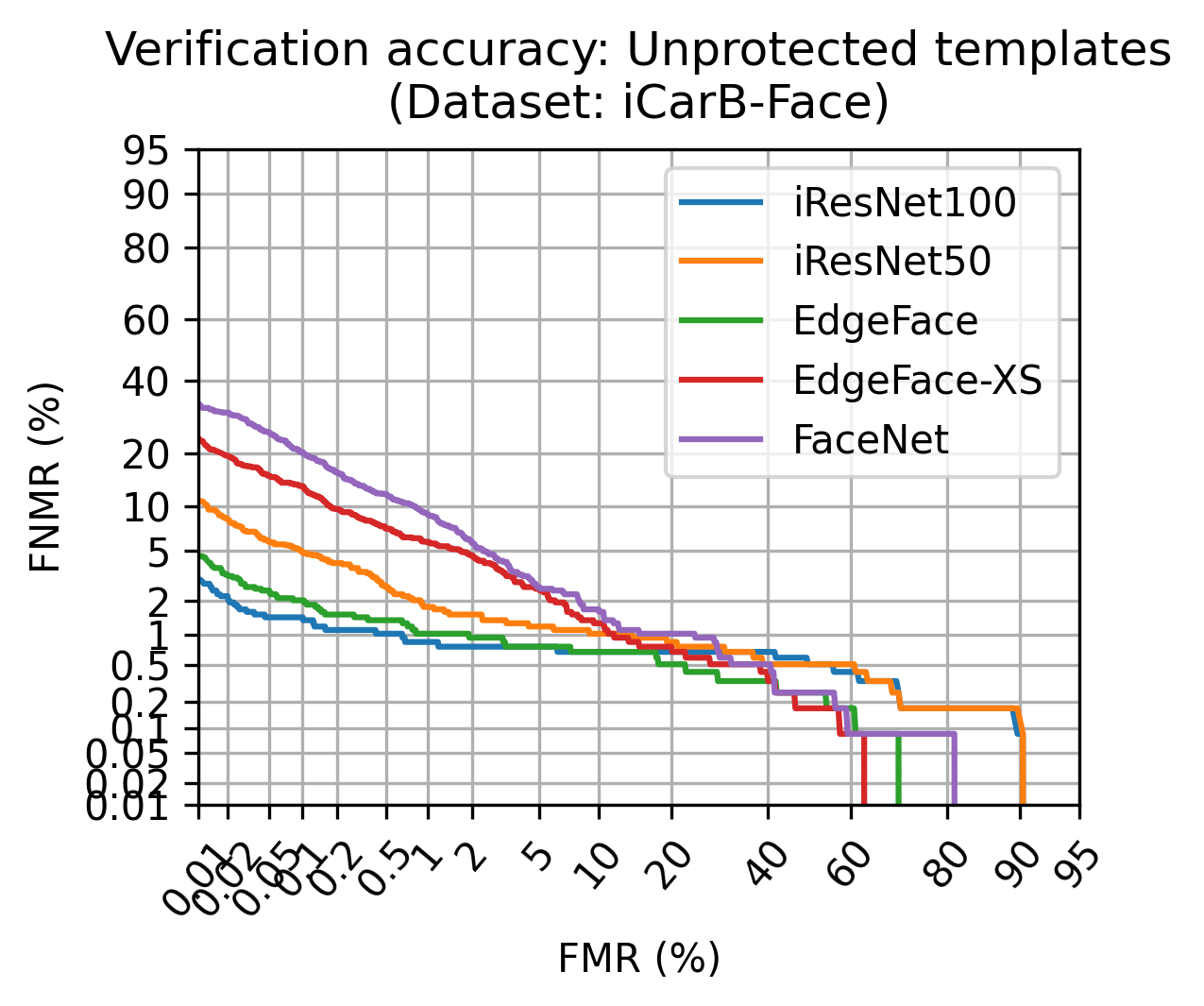}}
\caption{Detection Error Trade-off (DET) plots comparing the verification accuracy across face embeddings (unprotected face templates) generated using five different face recognition models, on three face datasets.}
\label{fig:unprotected_accuracy}
\end{figure}

To select the best face recognition model(s) for our study, the verification accuracy for each of the five sets of extracted face embeddings was computed on each of our three datasets.  Within each set of embeddings, all possible pairs were compared in terms of cosine distance, which resulted in: 15,165 genuine and 5,661,600 impostor scores for Multi-PIE; 3,150 genuine and 241,500 impostor scores for SOTERIA; and 1,182 genuine and 308,896 impostor scores for iCarB-Face.  Fig. \ref{fig:unprotected_accuracy} compares the resulting verification accuracy of the five face recognition models on our three datasets, in terms of the False Non-Match Rate (FNMR) and False Match Rate (FMR).  

From Fig. \ref{fig:unprotected_accuracy}, we see that, overall, iResNet100 achieves the highest accuracy, closely followed by EdgeFace.  So, both iResNet100 and EdgeFace were selected as our baseline face recognition models, whose face embeddings would be used in the first step of the PolyProtect evaluation: recognition accuracy in the protected domain, covered in Section \ref{sec:accuracy_eval}.

\section{PolyProtect Evaluation: Accuracy}
\label{sec:accuracy_eval}

Although the main aim of this paper is to dig deeper into the \textit{irreversibility} of PolyProtect, it is important to first demonstrate that this BTP method allows for acceptable \textit{recognition accuracy} in the protected domain.  This is because, when a BTP method is integrated into a face recognition system in practice, we must ensure that the ability of that system to perform facial recognition is not adversely affected as a result -- otherwise, the irreversibility of the BTP method is of little importance.  So, we begin our analysis of PolyProtect with an evaluation of its recognition accuracy.

\begin{figure}[!h]
\centering
\subfloat{\includegraphics[width=0.33\columnwidth]{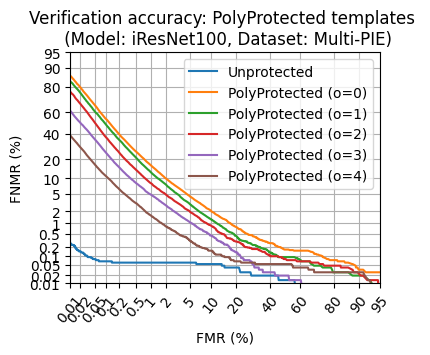}}
\hfil 
\subfloat{\includegraphics[width=0.33\columnwidth]{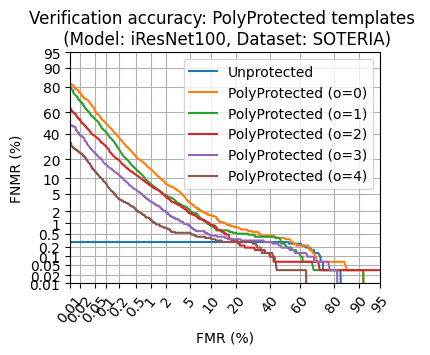}}
\hfil
\subfloat{\includegraphics[width=0.33\columnwidth]{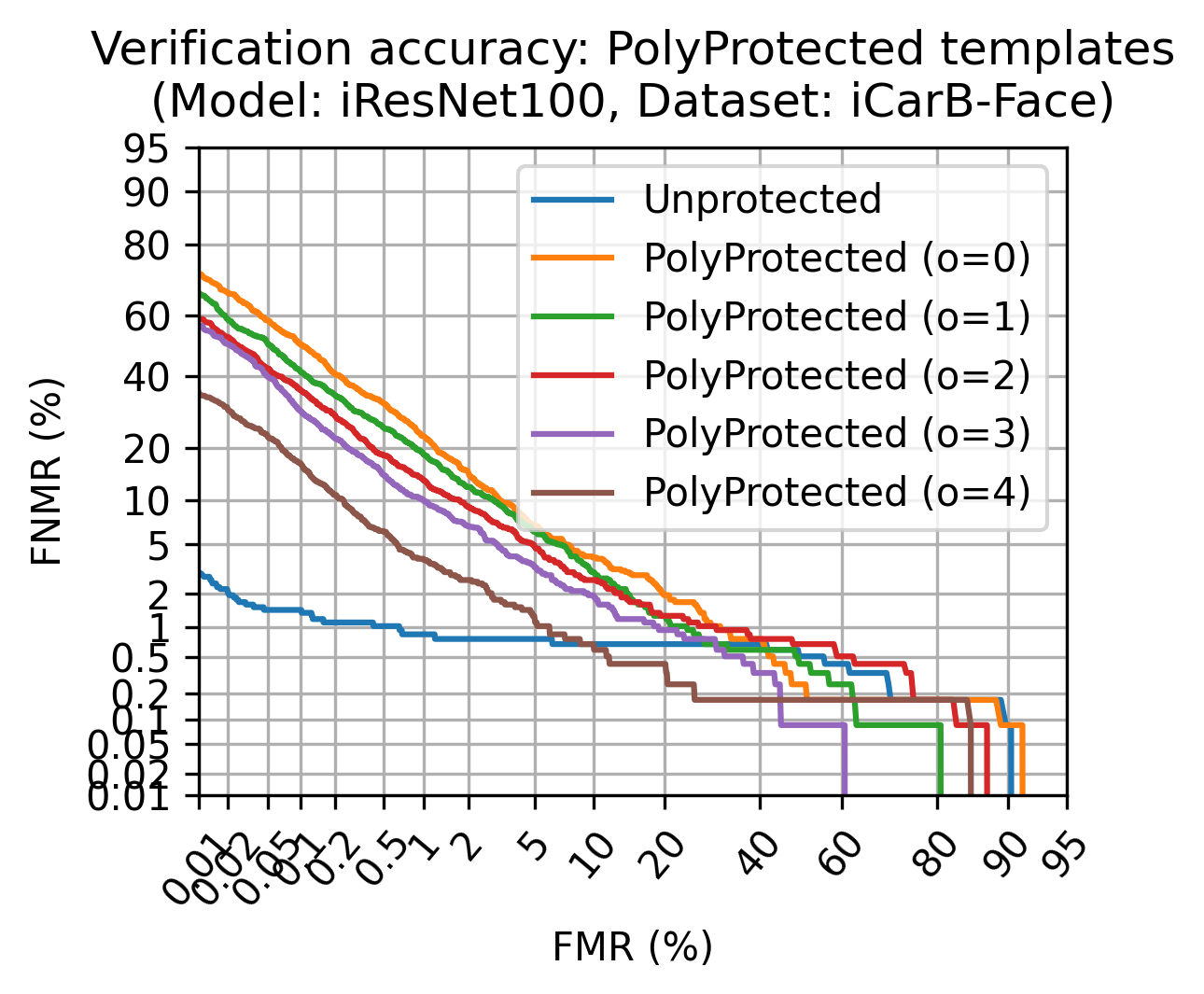}}
\vfil
\subfloat{\includegraphics[width=0.33\columnwidth]{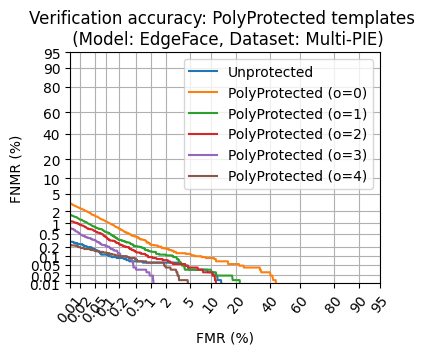}}
\hfil
\subfloat{\includegraphics[width=0.33\columnwidth]{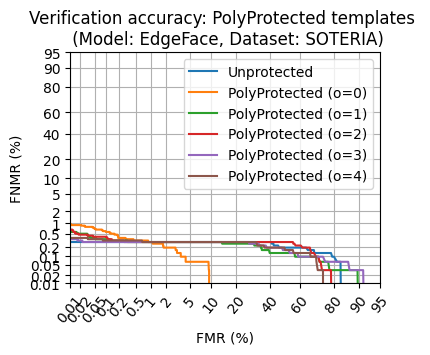}}
\hfil
\subfloat{\includegraphics[width=0.33\columnwidth]{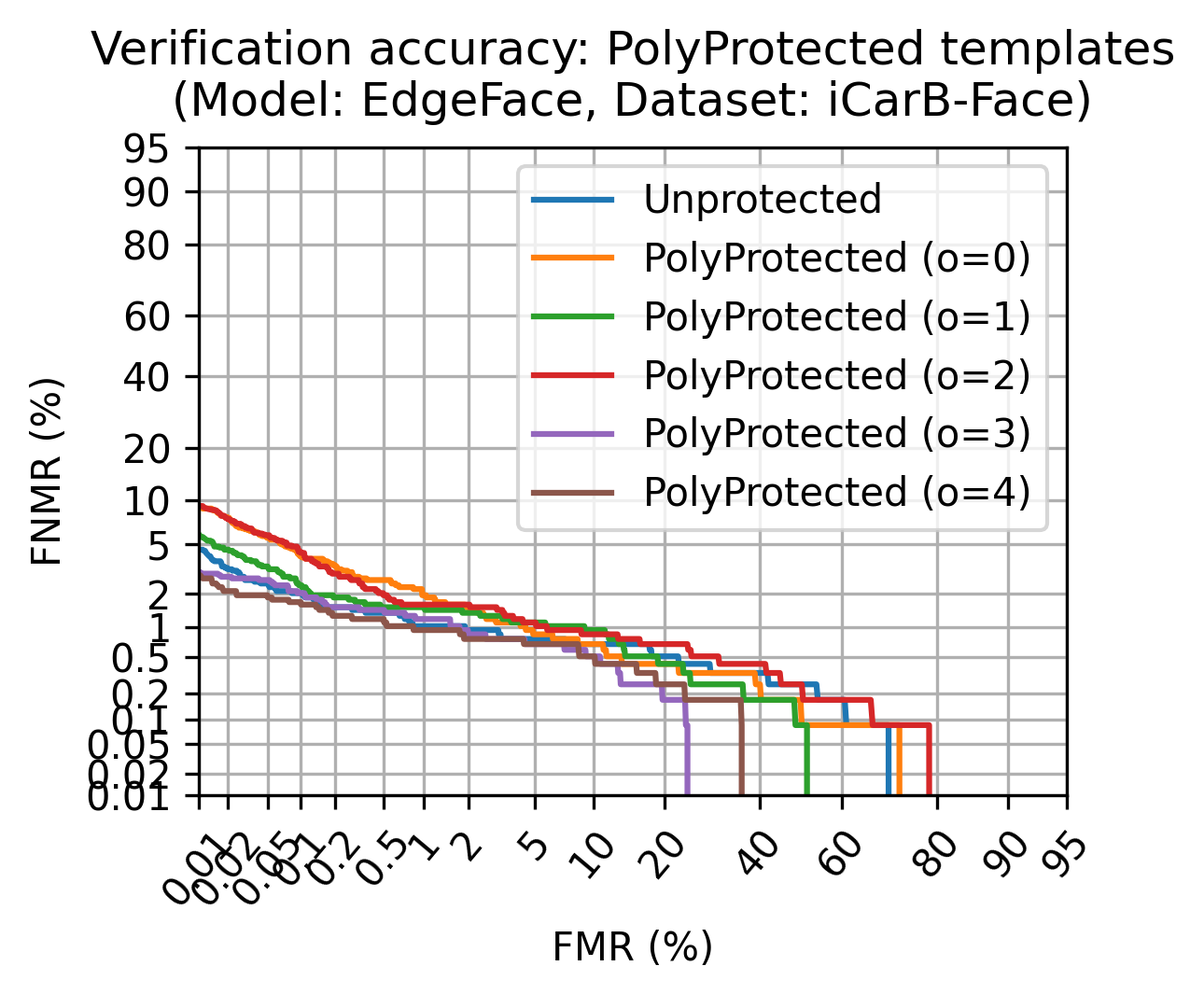}}
\caption{Detection Error Trade-off (DET) plots comparing verification accuracy across PolyProtected templates generated from iResNet100 and EdgeFace face embeddings using different overlaps ($o$).}
\label{fig:protected_accuracy}
\end{figure}

The first step was to apply PolyProtect to our iResNet100 and EdgeFace face embeddings (from Section \ref{subsec:models_datasets}), in order to transform them into \textit{protected templates}.  This transformation was performed using the PolyProtect parameters specified in Section \ref{subsec:polyprotect}, separately for each ``overlap'' in the range $[0, 4]$.  This means that we ended up with a separate set of protected face templates for each of our two sets of face embeddings (iResNet100 and EdgeFace), from each of our three datasets (Multi-PIE, SOTERIA, and iCarB-Face), and for each of the five aforementioned PolyProtect overlaps.  Then, within each of these 30 sets of protected face templates, we computed the cosine distance (comparison score) between every possible pair of templates.  Finally, the resulting scores were used to calculate the verification accuracy in the protected domain, in terms of FMR and FNMR.  Fig. \ref{fig:protected_accuracy} presents the verification accuracy in our different evaluation scenarios, in terms of DET plots.  Each plot compares the accuracy of the unprotected face templates from a particular model and dataset, against the accuracy of the corresponding PolyProtected templates when different amounts of overlap are used for the transform.

There are two important observations from Fig. \ref{fig:protected_accuracy}.  The first one is that, across all evaluation scenarios, the general trend is that accuracy increases as the amount of overlap increases.  The same trend was already observed in the original PolyProtect work \cite{kh22} and later confirmed in \cite{s25}, and it was attributed to the higher dimensionality of PolyProtected templates (and thus more information about the original face embedding being retained) when a larger overlap is used; so, this was expected.  

\begin{figure}[!h]
\centering
\subfloat{\includegraphics[width=0.2\paperwidth]{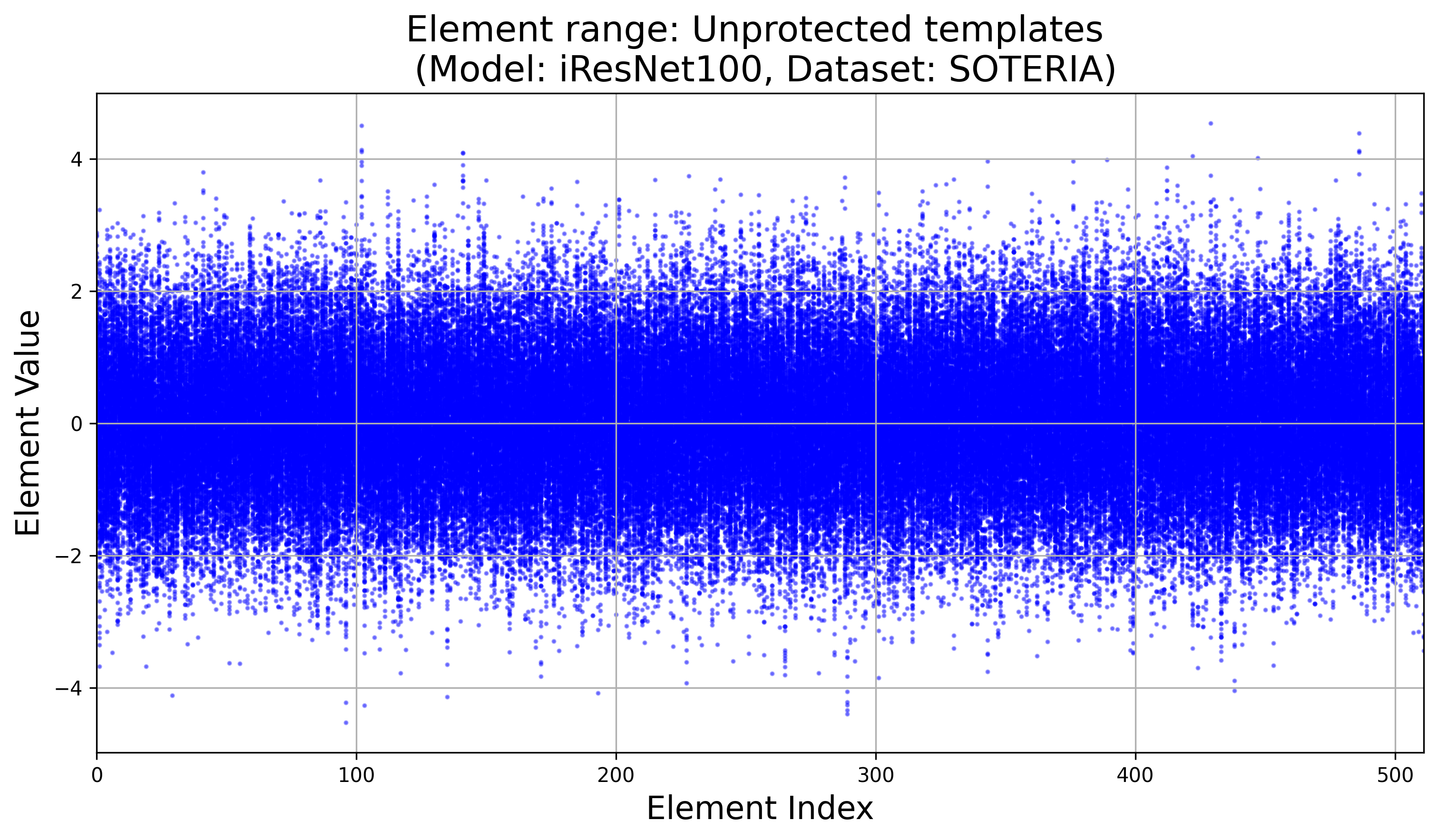}}
\hfil 
\subfloat{\includegraphics[width=0.2\paperwidth]{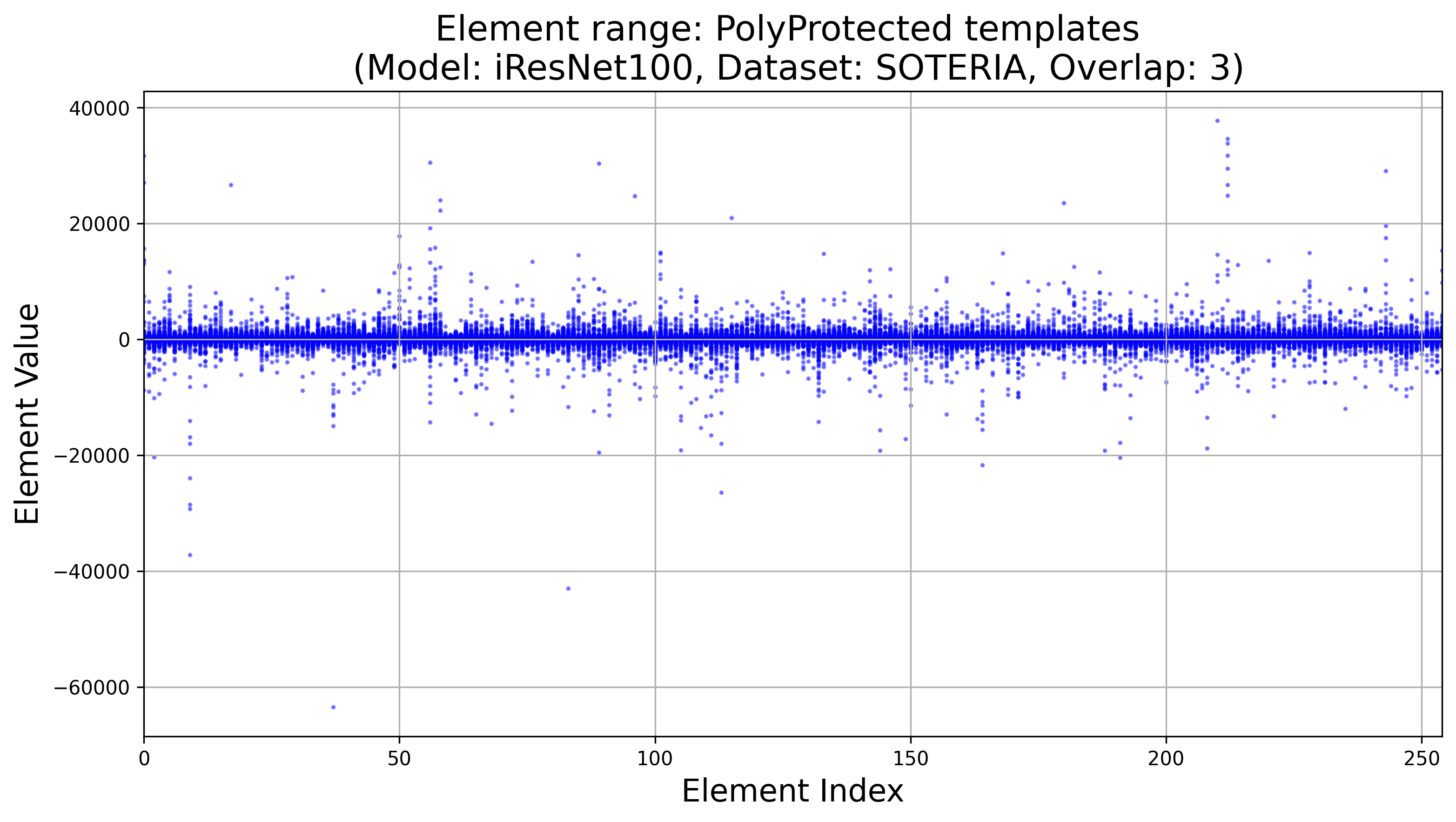}}
\vfil
\subfloat{\includegraphics[width=0.2\paperwidth]{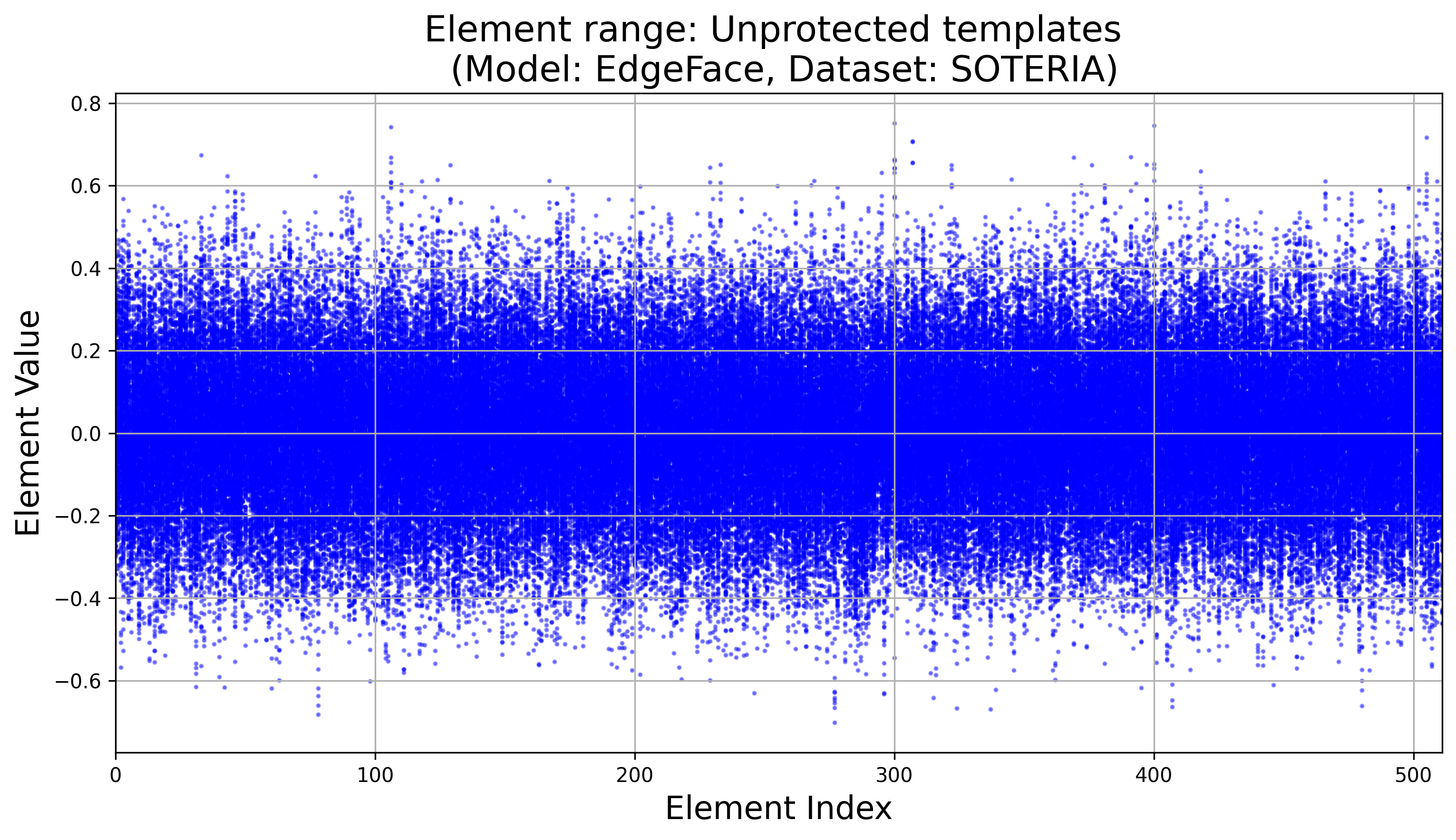}}
\hfil
\subfloat{\includegraphics[width=0.2\paperwidth]{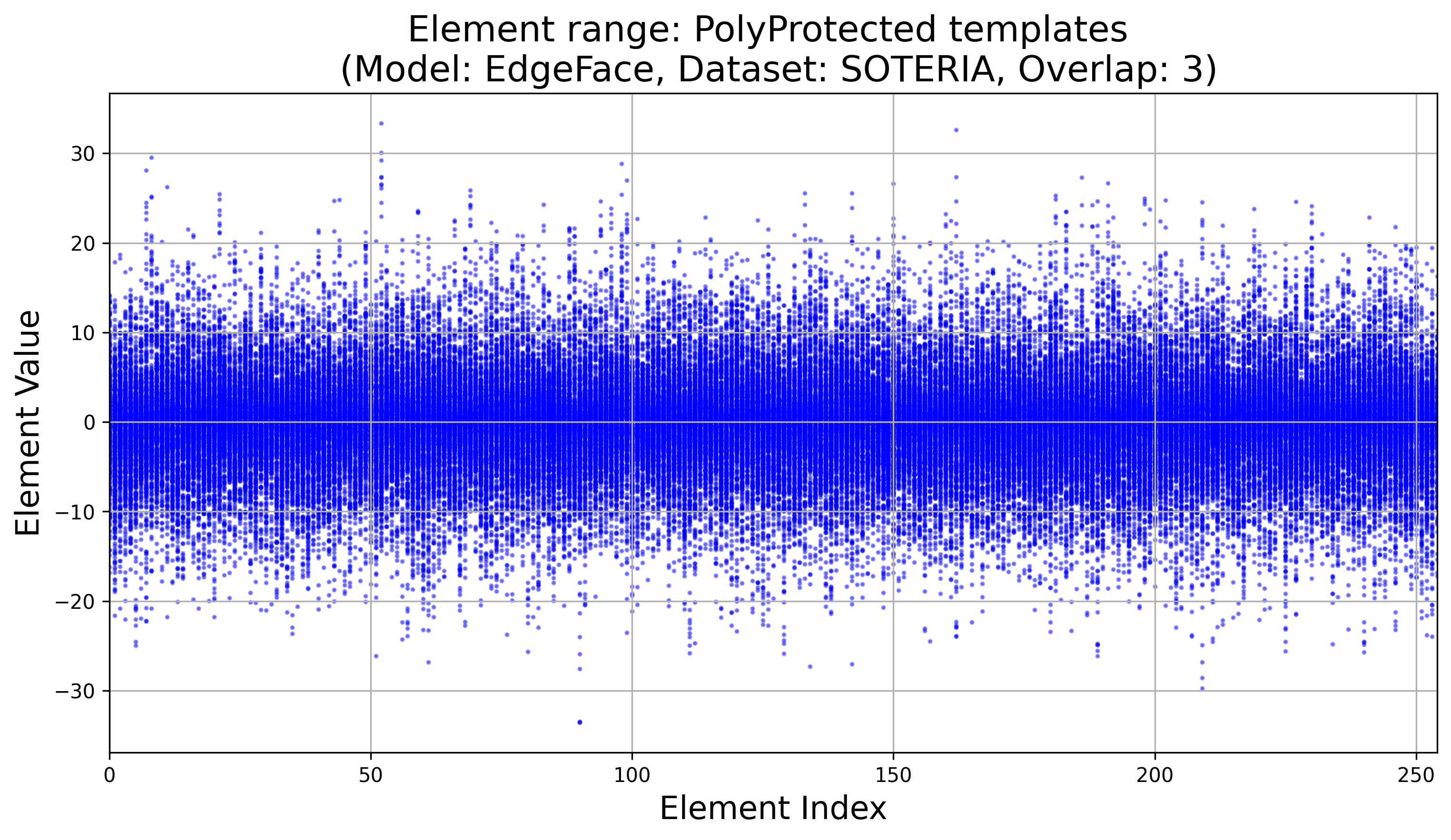}}
\caption{Range of iResNet100 and EdgeFace template elements before and after PolyProtect (overlap = 3), on the SOTERIA dataset.  Approximately the same ranges are observed on the Multi-PIE and iCarB-Face datasets, as well as across the other PolyProtect overlaps (0, 1, 2, 4).}
\label{fig:range}
\end{figure}

\begin{figure}[!h]
\centering
\subfloat{\includegraphics[width=0.4\columnwidth]{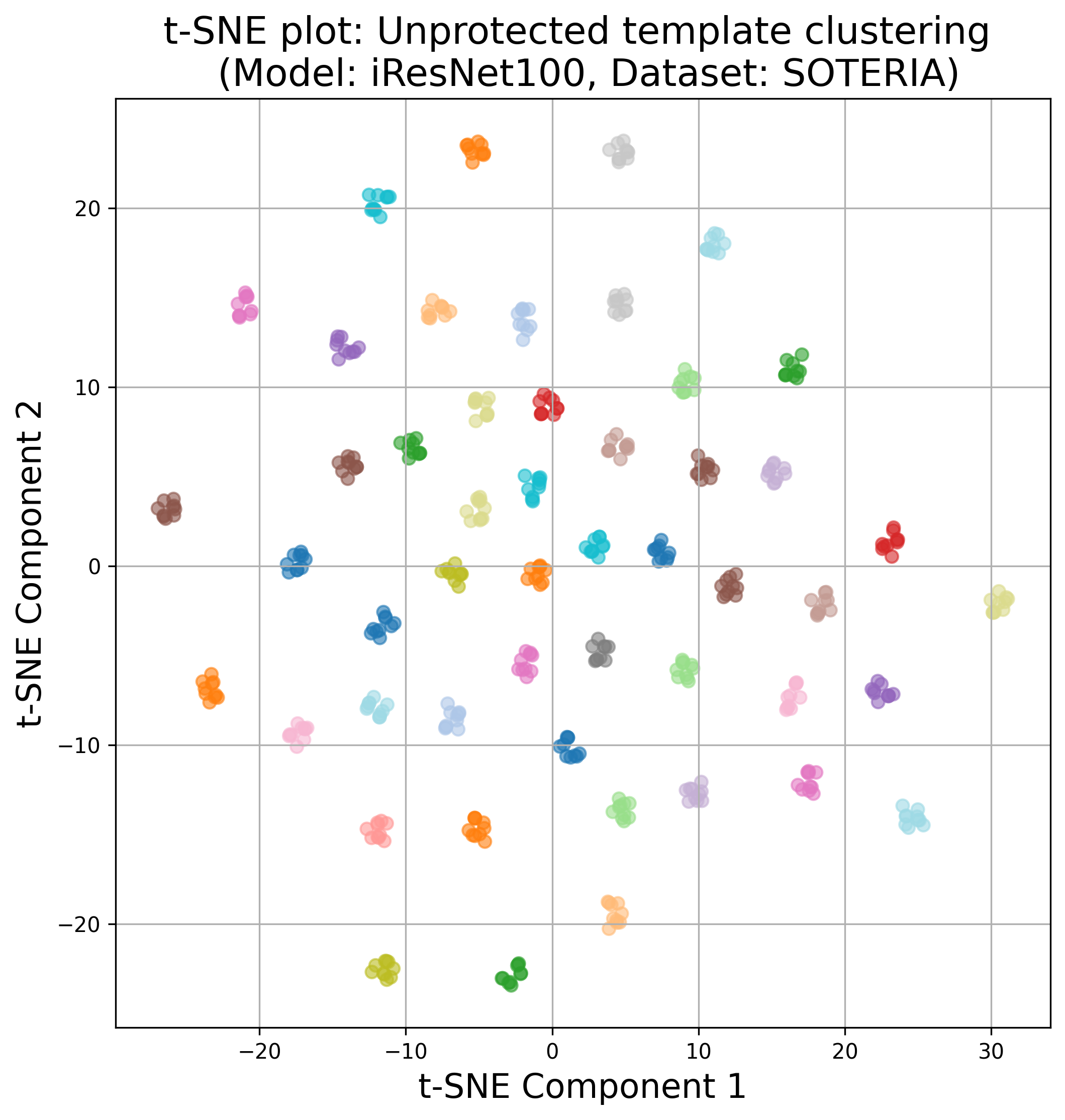}}
\hfil 
\subfloat{\includegraphics[width=0.4\columnwidth]{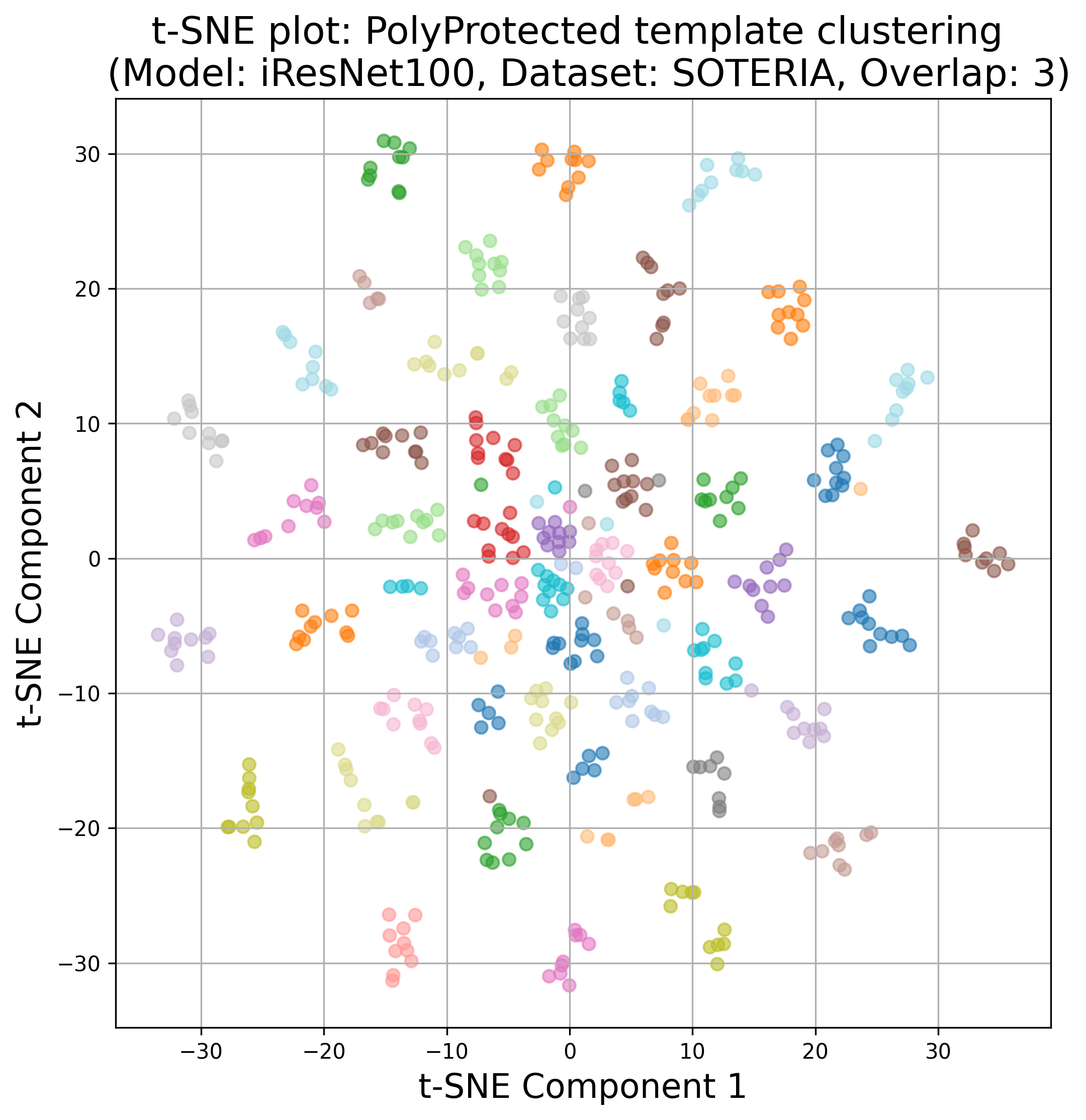}}
\vfil
\subfloat{\includegraphics[width=0.4\columnwidth]{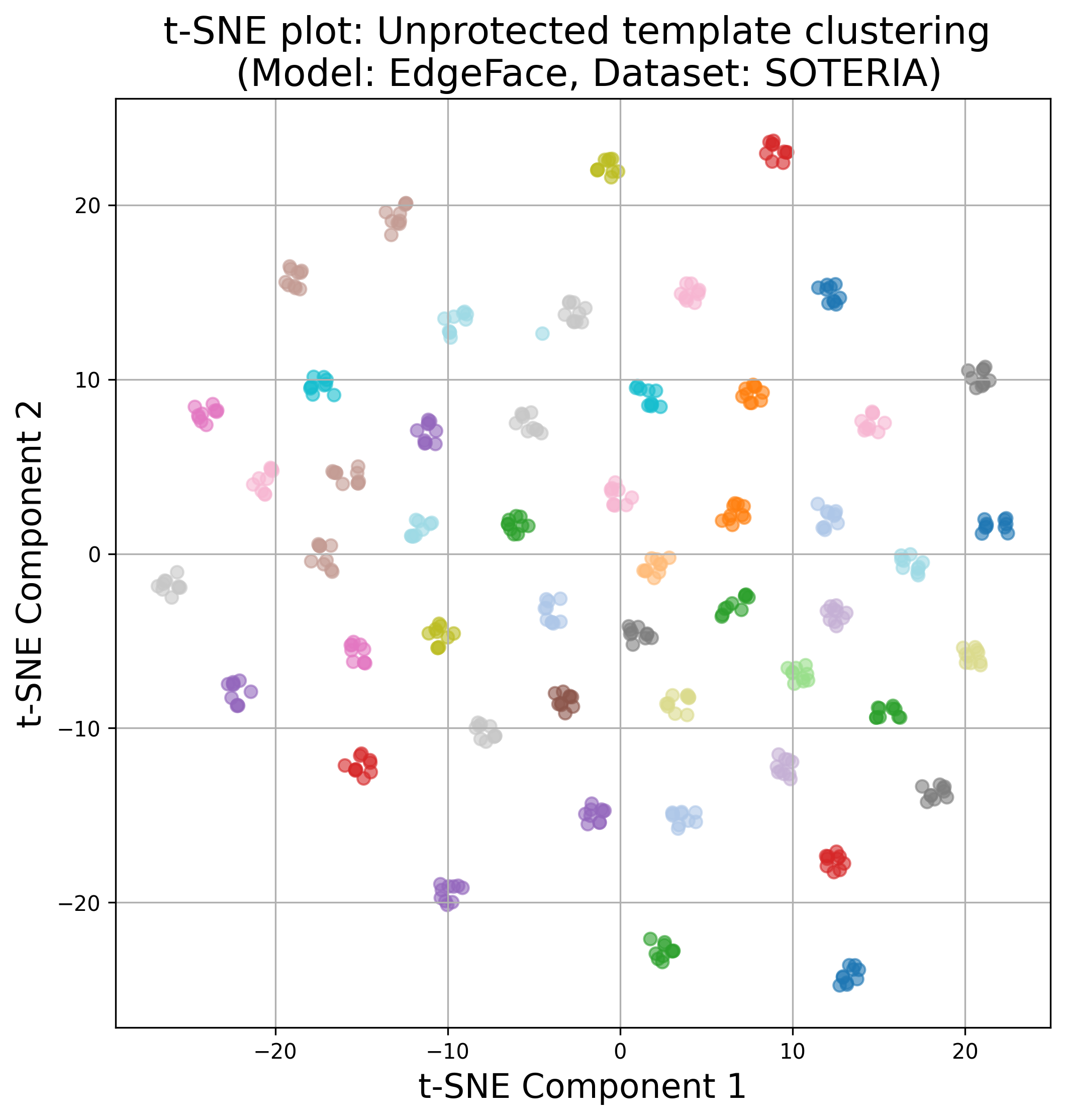}}
\hfil
\subfloat{\includegraphics[width=0.4\columnwidth]{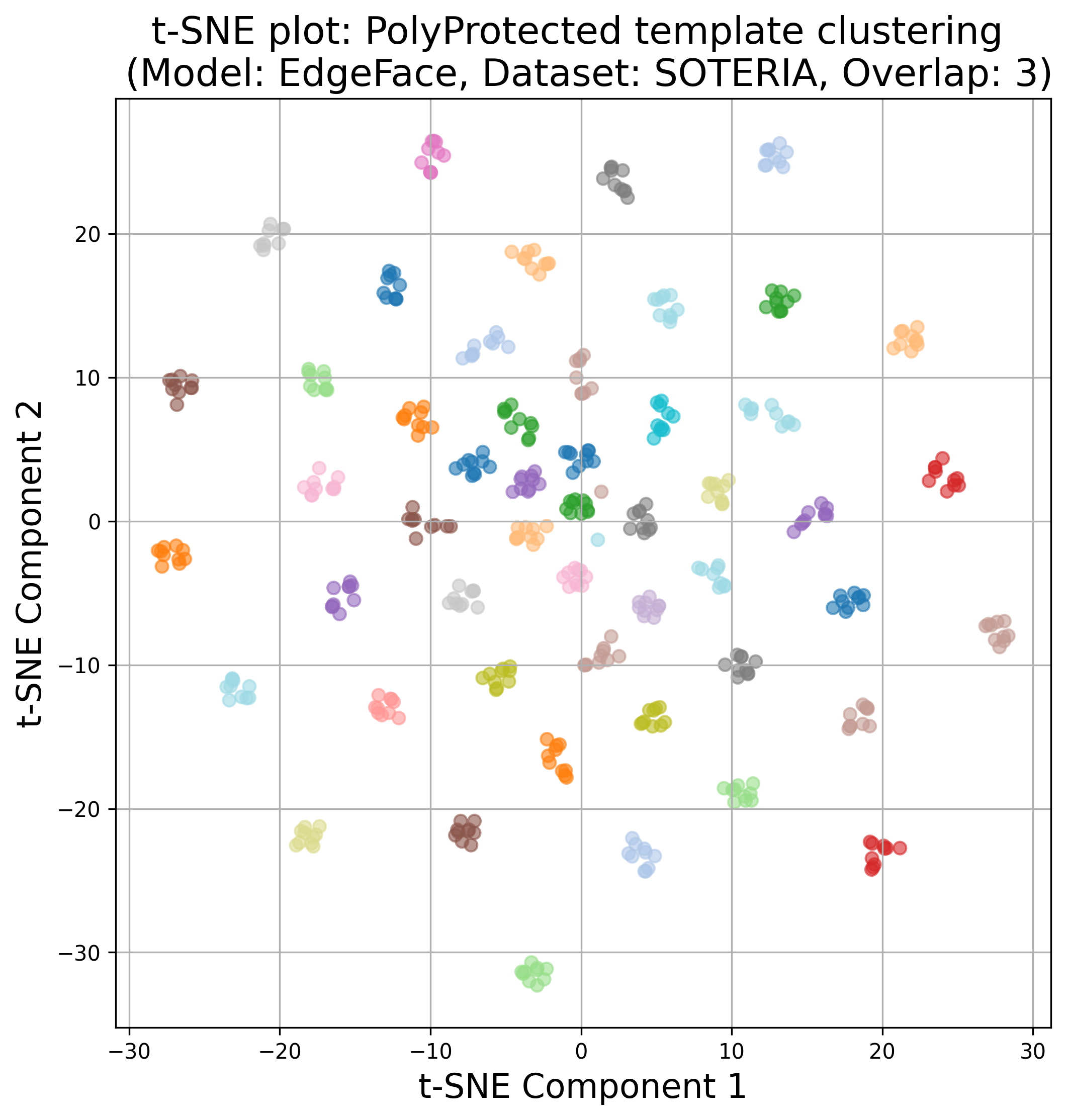}}
\caption{Class (identity) separation provided by iResNet100 and EdgeFace face templates before and after PolyProtect (overlap = 3), on SOTERIA.  Similar observations can be made on the Multi-PIE and iCarB-Face datasets, as well as for the other PolyProtect overlaps (0, 1, 2, 4).}
\label{fig:tsne}
\end{figure}

The second important finding from Fig. \ref{fig:protected_accuracy}, which has not previously been reported, is that accuracy in the PolyProtected domain seems to depend on the model used to generate the underlying face embeddings.  Specifically, although the iResNet100 and EdgeFace embeddings were found to have similar accuracy in the unprotected domain (Fig. \ref{fig:unprotected_accuracy}), in the protected domain the accuracy of the iResNet100 PolyProtected templates is significantly worse (Fig. \ref{fig:protected_accuracy}).  We believe this is because the iResNet100 face embeddings lie in a larger range compared to the EdgeFace embeddings.  So, when the PolyProtect transform is applied, the iResNet100 embeddings are more distorted, leading to greater intra-class variance among the iResNet100 protected templates and thus lower accuracy in the protected domain.  To validate this theory, Fig. \ref{fig:range} compares the range of the iResNet100 versus EdgeFace template (embedding) elements before and after PolyProtect is applied.  Fig. \ref{fig:tsne} then presents t-SNE plots to illustrate the differences in the class (identity) clustering and intra-class variance of the unprotected versus protected iResNet100 and EdgeFace templates.  Due to space constraints, these plots show the results for only the SOTERIA dataset, and only for PolyProtected templates generated using an overlap of 3; however, the same observations can be made for the other datasets and overlaps (i.e., same range regardless of the dataset and overlap, and similar identity clustering behaviour).

From Fig. \ref{fig:range}, we see that the unprotected EdgeFace templates lie in the range of approximately $[-0.7, 0.7]$, while the unprotected iResNet100 templates lie in the larger range of about $[-4, 4]$.  Once these embeddings are transformed via PolyProtect, the range of protected iResNet100 templates explodes to roughly $[-20000, 20000]$, which is 1,000 times larger than the protected EdgeFace range of about $[-20, 20]$.  So, it makes sense to postulate that any intra-class variance present in the face embeddings prior to the PolyProtect transformation would be more exaggerated in the iResNet100 protected templates than in the EdgeFace protected templates.  The t-SNE plots in Fig. \ref{fig:tsne} show that this indeed seems to be the case: while the clustering of the classes (identities) is approximately the same among the unprotected and PolyProtected EdgeFace templates, as well as the unprotected iResNet100 templates, the classes in the PolyProtected iResNet100 domain are more dispersed.  So, the same identities become more difficult to distinguish (separate) when represented using PolyProtected iResNet100 templates.  This can be used to explain why the accuracy in the iResNet100 protected domain is worse than the accuracy in both the iResNet100 unprotected domain and the EdgeFace protected domain (Fig. \ref{fig:protected_accuracy}).

\begin{figure}[!h]
\centering
\subfloat{\includegraphics[width=0.33\columnwidth]{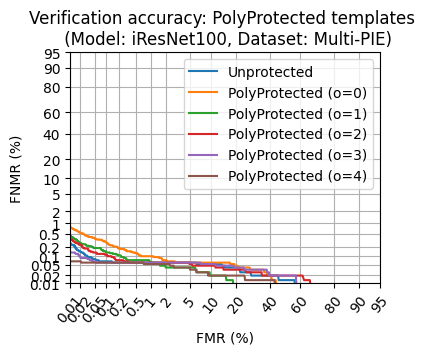}}
\hfil 
\subfloat{\includegraphics[width=0.33\columnwidth]{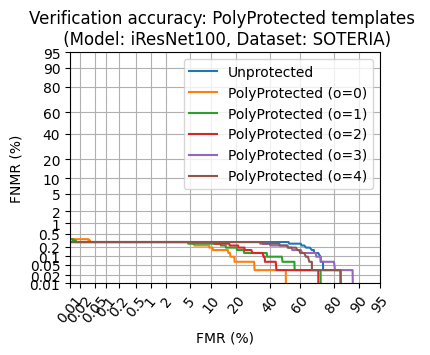}}
\hfil
\subfloat{\includegraphics[width=0.33\columnwidth]{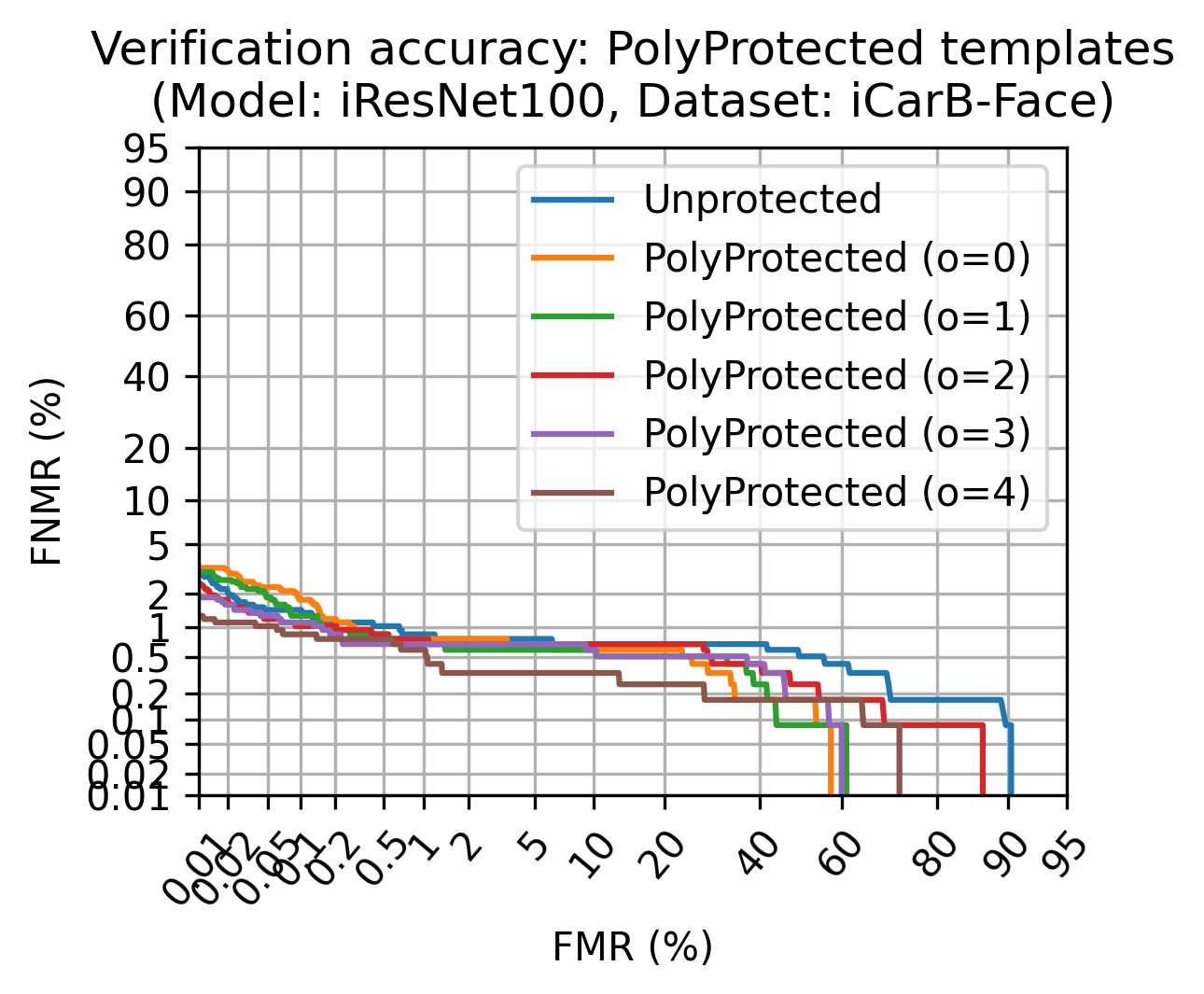}}
\vfil
\subfloat{\includegraphics[width=0.33\columnwidth]{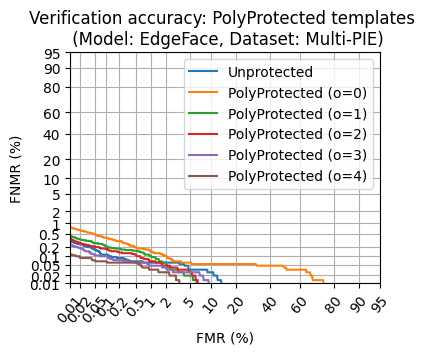}}
\hfil
\subfloat{\includegraphics[width=0.33\columnwidth]{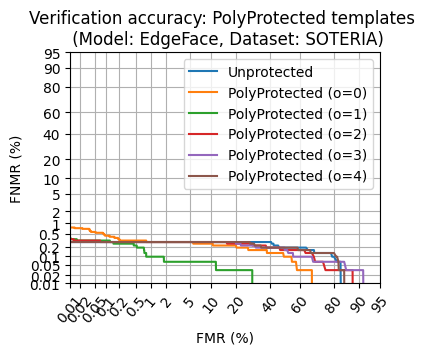}}
\hfil
\subfloat{\includegraphics[width=0.33\columnwidth]{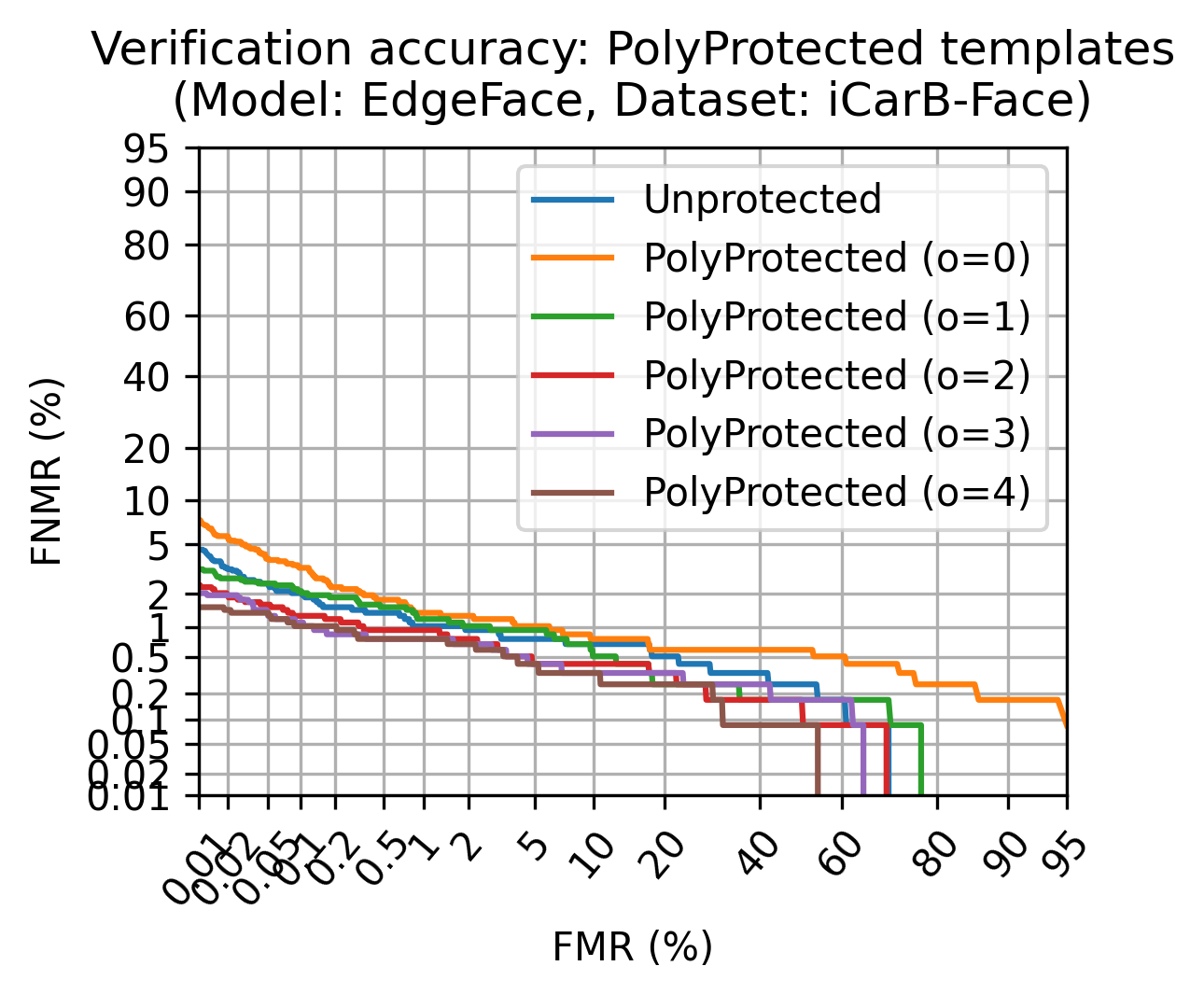}}
\caption{Detection Error Trade-off (DET) plots comparing verification accuracy across PolyProtected templates generated from normalized iResNet100 and EdgeFace face embeddings using different overlaps ($o$).}
\label{fig:protected_accuracy_norm}
\end{figure}

At this stage, we may be tempted to conclude that PolyProtect is not a suitable BTP method for face embeddings generated using the iResNet100 model (or other models that produce embeddings lying in a relatively large range).  However, we found a very simple fix for this issue: \textit{normalize} the embeddings prior to applying PolyProtect.  This way, regardless of the model used, all face embeddings end up lying in the same range, which prevents exaggerated intra-class variance for any one model in the protected domain.  Fig. \ref{fig:protected_accuracy_norm} shows the same DET plots from Fig. \ref{fig:protected_accuracy}, except this time the unprotected face embeddings are \textit{normalized} prior to applying PolyProtect.  To facilitate the ``unnormalized'' versus ``normalized'' accuracy comparison, Table \ref{tab:unnorm_vs_norm} quantifies the results in terms of the FNMR at the commonly used 0.1\% FMR threshold. 

\begin{table}[!h]
\renewcommand{\arraystretch}{1.2}
\caption{Verification accuracy for unnormalized (\textit{U}) VS normalized (\textit{N}) iResNet100 and EdgeFace face embeddings, when they are PolyProtected using different overlaps ($o$).\label{tab:unnorm_vs_norm}} 
\centering
\begin{tabular}{|c|c|c|c|c|c|c|c|}
\hline
\multirow{3}{*}{Model} & \multirow{3}{*}{Templates} & \multicolumn{6}{c|}{FNMR (\%) @ 0.1\% FMR} \\
\cline{3-8}
 & & \multicolumn{2}{c|}{Multi-PIE} & \multicolumn{2}{c|}{SOTERIA} & \multicolumn{2}{c|}{iCarB-Face} \\
\cline{3-8}
 & & \textit{U} & \textit{N} & \textit{U} & \textit{N} & \textit{U} & \textit{N} \\
\hline
\multirow{6}{*}{iResNet100} & Unprotected & 0.1 & 0.1 & 0.3 & 0.3 & 1.4 & 1.4 \\
\cline{2-8}
 & $o$ = 0  & 53.3 & \textbf{0.3} & 49.3 & \textbf{0.3} & 50.6 & \textbf{1.8} \\
 \cline{2-8}
 & $o$ = 1  & 48.2 & \textbf{0.1} & 40.9 & \textbf{0.3} & 41.8 & \textbf{1.3} \\
 \cline{2-8}
 & $o$ = 2  & 36.6 & \textbf{0.1} & 29.0 & \textbf{0.3} & 35.8 & \textbf{1.0} \\
 \cline{2-8}
 & $o$ = 3  & 24.2 & \textbf{0.1} & 16.7 & \textbf{0.3} & 29.5 & \textbf{1.1} \\
 \cline{2-8}
 & $o$ = 4  & 10.7 & \textbf{0.1} & 7.9 & \textbf{0.3} & 16.5 & \textbf{0.8} \\
\hline
\multirow{6}{*}{EdgeFace} & Unprotected & 0.1 & 0.1 & 0.3 & 0.3 & 2.0 & 2.0 \\
\cline{2-8}
 & $o$ = 0  & 1.1 & \textbf{0.4} & 0.6 & \textbf{0.4} & 4.1 & \textbf{3.3} \\
 \cline{2-8}
 & $o$ = 1  & 0.6 & \textbf{0.2} & 0.4 & \textbf{0.3} & 2.4 & \textbf{2.1} \\
 \cline{2-8}
 & $o$ = 2  & 0.4 & \textbf{0.2} & 0.4 & \textbf{0.3} & 4.4 & \textbf{1.3} \\
 \cline{2-8}
 & $o$ = 3  & 0.2 & \textbf{0.1} & 0.3 & 0.3 & 2.0 & \textbf{1.1} \\
 \cline{2-8}
 & $o$ = 4  & 0.1 & 0.1 & 0.3 & 0.3 & 1.7 & \textbf{1.0} \\ 
\hline
\end{tabular}
\end{table} 

Comparing Fig. \ref{fig:protected_accuracy_norm} to Fig. \ref{fig:protected_accuracy}, and based on the results in Table \ref{tab:unnorm_vs_norm}, it is clear that normalization is indeed an effective fix to the issue of poor accuracy in the iResNet100 protected domain.  Now, regardless of the face recognition model, the accuracy in the PolyProtected domain remains relatively close to the corresponding baseline (unprotected) accuracy (with higher PolyProtect overlaps resulting in better accuracy, as before).  Even for EdgeFace, normalization results in slightly better accuracy in the protected domain, compared to when the embeddings are not normalized.  So, we may conclude that, in order to ensure the best possible accuracy in the PolyProtected domain, it is a good idea to normalize the face embeddings prior to applying the PolyProtect transform.

We are now ready to dig into the \textit{irreversibility} of PolyProtect.  Based on the findings in our accuracy analysis, the irreversibility evaluation will be conducted on protected templates generated from \textit{normalized} face embeddings only.  Furthermore, due to space constraints, in the \textit{figures} that follow we will report irreversibility results on iResNet100 only,\footnote{EdgeFace figures are comparable and can be reproduced using our code.} since the accuracy of normalized iResNet100 templates in the protected domain was found to be slightly better than that of EdgeFace (see Table \ref{tab:unnorm_vs_norm}).  However, to demonstrate the generalizability of our findings, the \textit{tables} that follow will show results for EdgeFace as well.

\section{PolyProtect Evaluation: Irreversibility}
\label{sec:irreversibility_eval}

When evaluating the irreversibility of PolyProtect, we are trying to answer the following question: Is it possible to invert the $V \rightarrow P$ transform (i.e., perform the inverse transform, $P \rightarrow V$) to recover a face embedding, $V$, from its protected template, $P$?  In the original work \cite{kh22}, the irreversibility of PolyProtect was evaluated in two ways: theoretically and empirically.  The theoretical analysis showed that the inverse transform is defined by an \textit{underdetermined} system of equations and, therefore, technically does not exist.  This is because there are (theoretically) infinitely many solutions for the elements in $V$ that could produce $P$, so there is \textit{no unique solution}.  So, it was concluded that, in theory, the PolyProtect transform is irreversible (non-invertible).  Since this conclusion is mathematically sound, there is no reason for us to investigate the theoretical irreversibility further.  The only difference is that our face embeddings are 512-dimensional, whereas those in \cite{kh22} were 128-dimensional, so the number of equations considered in our theoretical analysis would differ.  Table \ref{tab:mappings} shows the forward ($V \rightarrow P$) and inverse ($P \rightarrow V$) transforms when PolyProtect is applied to our 512-dimensional face embeddings.  The dimensionality of $P$ indicates the number of equations involved; e.g., when overlap = 0, the $V \rightarrow P$ transform is defined by 103 equations in 512 variables (unknowns), resulting in a 103-dimensional PolyProtected template, $P$. So, the inverse transform, $P \rightarrow V$, cannot be uniquely defined, due to the $512 - 103 = 409$ degrees of freedom.  In other words, it is mathematically impossible to invert the 103-dimensional $P$ to recover the 512-dimensional $V$.  The same may be concluded for the other overlaps, albeit with differing degrees of freedom.  

\begin{table}[!h]
\renewcommand{\arraystretch}{1.3}
\caption{$V \rightarrow P$ and $P \rightarrow V$ transforms for different overlaps.
\label{tab:mappings}} 
\centering
\begin{tabular}{|c|c|c|}
\hline
\textbf{Overlap} & $\mathbf{V \rightarrow P}$ & $\mathbf{P \rightarrow V}$ \\
\hline
0 & $\mathbb{R}^{512} \rightarrow \mathbb{R}^{103}$ & $\mathbb{R}^{103} \rightarrow \mathbb{R}^{512}$ \\
\hline 
1 & $\mathbb{R}^{512} \rightarrow \mathbb{R}^{128}$ & $\mathbb{R}^{128} \rightarrow \mathbb{R}^{512}$ \\
\hline
2 & $\mathbb{R}^{512} \rightarrow \mathbb{R}^{170}$ & $\mathbb{R}^{170} \rightarrow \mathbb{R}^{512}$ \\
\hline
3 & $\mathbb{R}^{512} \rightarrow \mathbb{R}^{255}$ & $\mathbb{R}^{255} \rightarrow \mathbb{R}^{512}$ \\
\hline
4 & $\mathbb{R}^{512} \rightarrow \mathbb{R}^{508}$ & $\mathbb{R}^{508} \rightarrow \mathbb{R}^{512}$ \\
\hline
\end{tabular}
\end{table}  

The second type of irreversibility analysis in \cite{kh22} involved using a numerical solver to try to approximate a solution for $V$ from $P$, even if a unique solution technically does not exist.  This inversion attack was simulated using an open-source numerical solver: Python's \textit{scipy.optimize.root} function with the \textit{lm} method.  This method adopts the Levenberg-Marquardt algorithm, which approximates a solution to a non-linear system of equations using a damped least-squares approach.  So, this solver essentially tries to find some set of $n$ elements representing the $n$-dimensional face embedding, $V$, such that when we apply PolyProtect to those $n$ elements the Euclidean distance between the resulting protected template and the true protected template, $P$, is as small as possible. 

In \cite{kh22}, the success of this inversion attack was quantified as follows.  If, for a particular $P$, the solver converges to a solution, $V^{*}$, calculate the comparison score (i.e., cosine distance) between $V^{*}$ and the true embedding, $V$.  The idea is to determine whether $V^{*}$ is a \textit{close enough} approximation to $V$, such that $V^{*}$ could be used to impersonate the identity represented by $V$ in a face recognition system that stores the \textit{unprotected} $V$ as a reference face embedding.  If the resulting cosine distance is smaller than a pre-defined threshold, the inversion attack is deemed successful; otherwise, the PolyProtected template is considered ``irreversible'' at this threshold.  

Since the success rate of an inversion attack in \cite{kh22} was based on cosine distance, we asked ourselves the following question: Would the inversion attack be more successful if we were to use a numerical solver that tries to minimise the \textit{cosine} distance between $P$ and the protected version of $V^{*}$, as opposed to Euclidean distance?  To answer this question, we performed two inversion attacks: one based on Python's \textit{scipy.optimize.root} function with the \textit{lm} method, as in \cite{kh22}, and the other one based on Python's \textit{scipy.optimize.minimize} function with the \textit{L-BFGS-B} method, which was set up to minimise the cosine distance between the PolyProtected version of the solution, $V^{*}$, and $P$.  

The two solvers were applied to the PolyProtected templates generated from our 512-dimensional, normalized iResNet100 and EdgeFace embeddings, which originated from the three datasets mentioned in Section \ref{subsec:models_datasets}.  For each dataset, we selected only one face image, and thus face embedding, per identity (to represent \textit{reference} embeddings), meaning that we ended up with one PolyProtected template per identity.  In total, this amounted to 337 protected templates from the Multi-PIE dataset, 70 from SOTERIA, and 197 from iCarB-Face.  As in \cite{kh22}, our irreversibility evaluation was based on the worst-case scenario of a fully-informed attacker, corresponding to the \textit{full disclosure} threat model defined in ISO/IEC 30316\footnote{\url{https://www.iso.org/standard/53256.html}} (i.e., the PolyProtect algorithm and all parameters are known by the attacker).  The only difference was that, in our work, the initial guesses for the solvers were drawn from probability distributions estimated on the \textit{same} face embeddings on which the inversion attacks were performed, whereas in \cite{kh22} the distributions were estimated on a different set of face embeddings.  So, ours represents an even \textit{more informed} attacker than that assumed in \cite{kh22}, which should not be encountered in practice but is useful for estimating the worst-case irreversibility of PolyProtect.  

\begin{figure*}[!h]
\centering
\subfloat{\includegraphics[width=0.25\paperwidth]{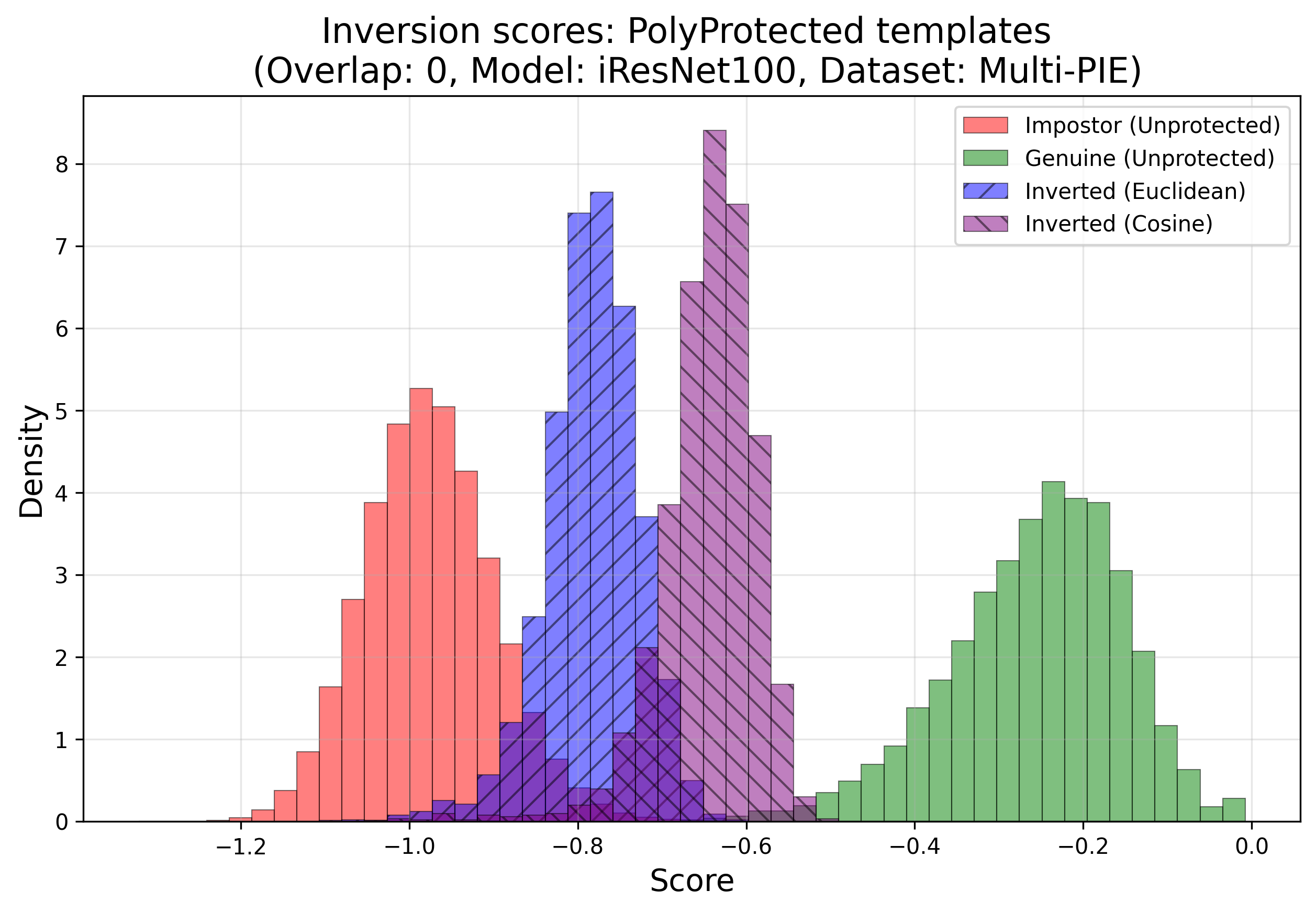}}
\hfil 
\subfloat{\includegraphics[width=0.25\paperwidth]{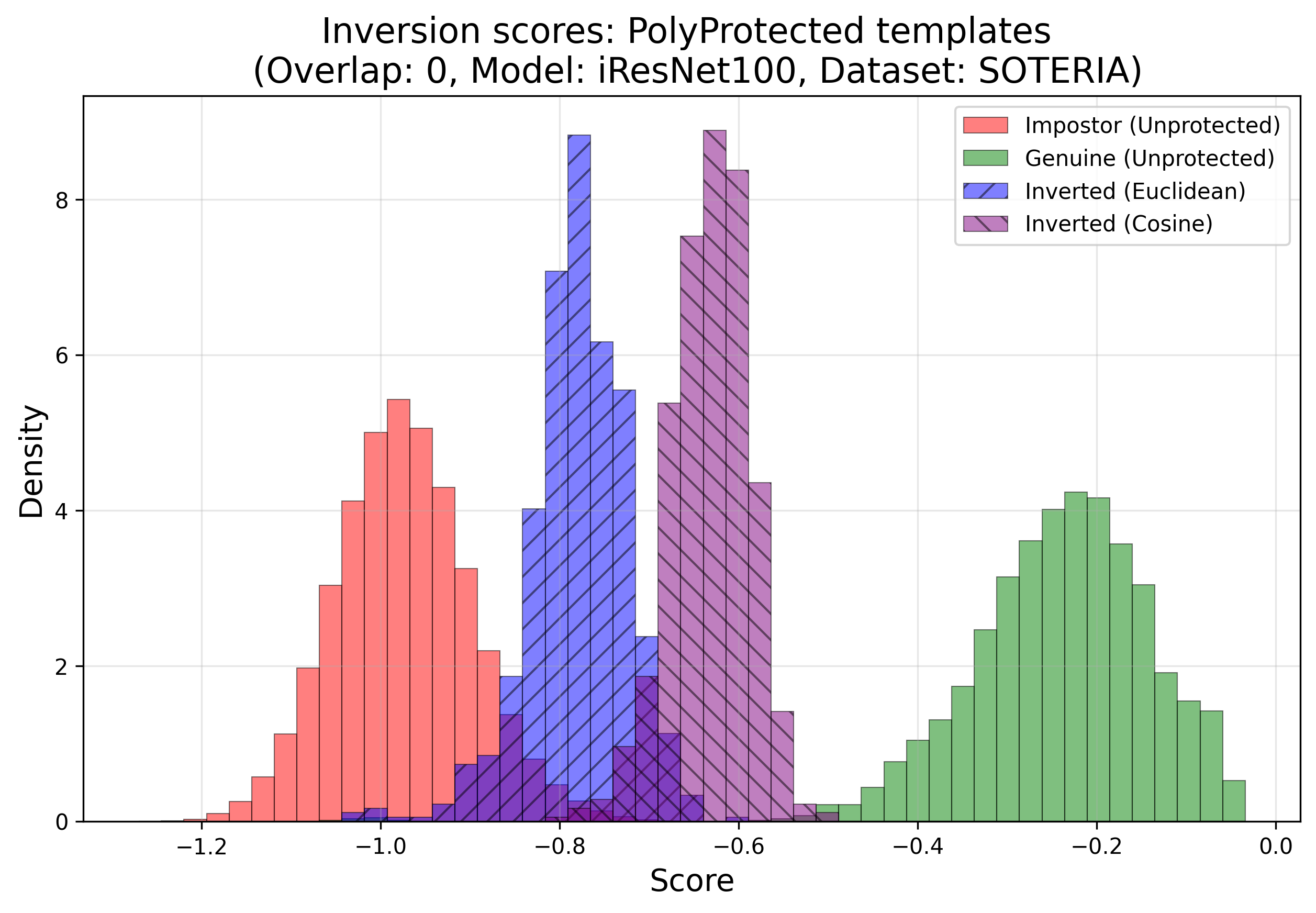}}
\hfil
\subfloat{\includegraphics[width=0.25\paperwidth]{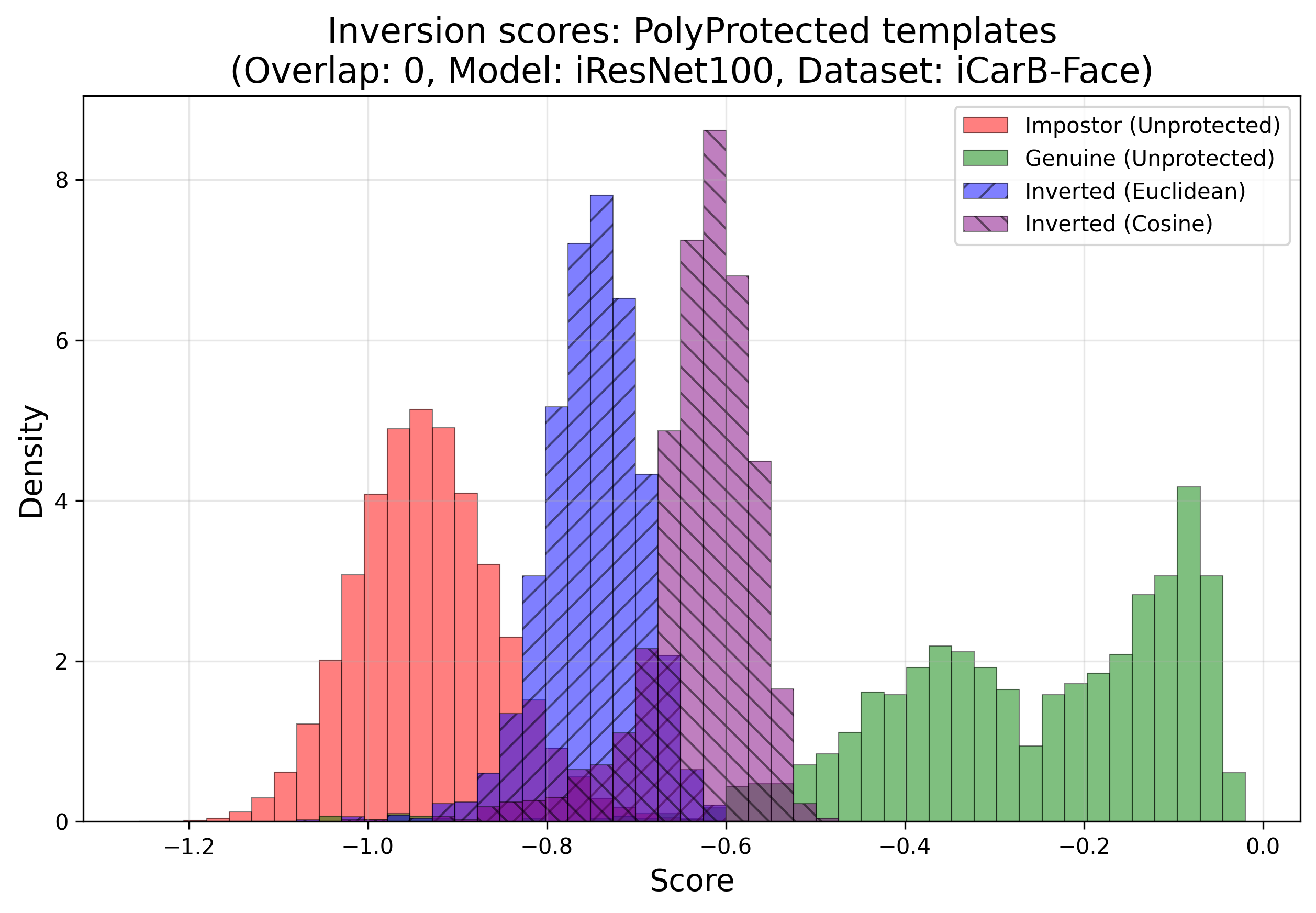}}
\vfil
\subfloat{\includegraphics[width=0.25\paperwidth]{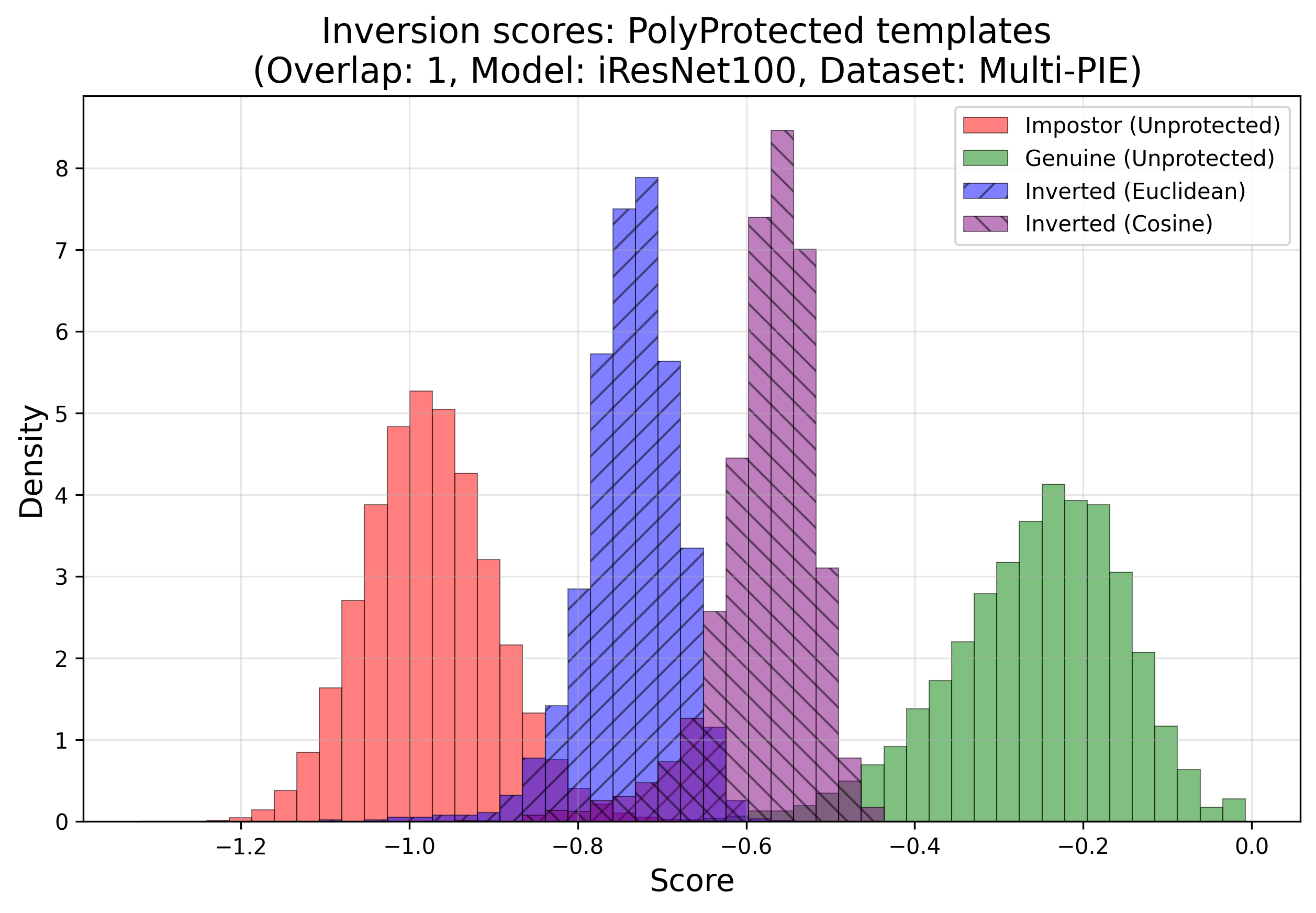}}
\hfil
\subfloat{\includegraphics[width=0.25\paperwidth]{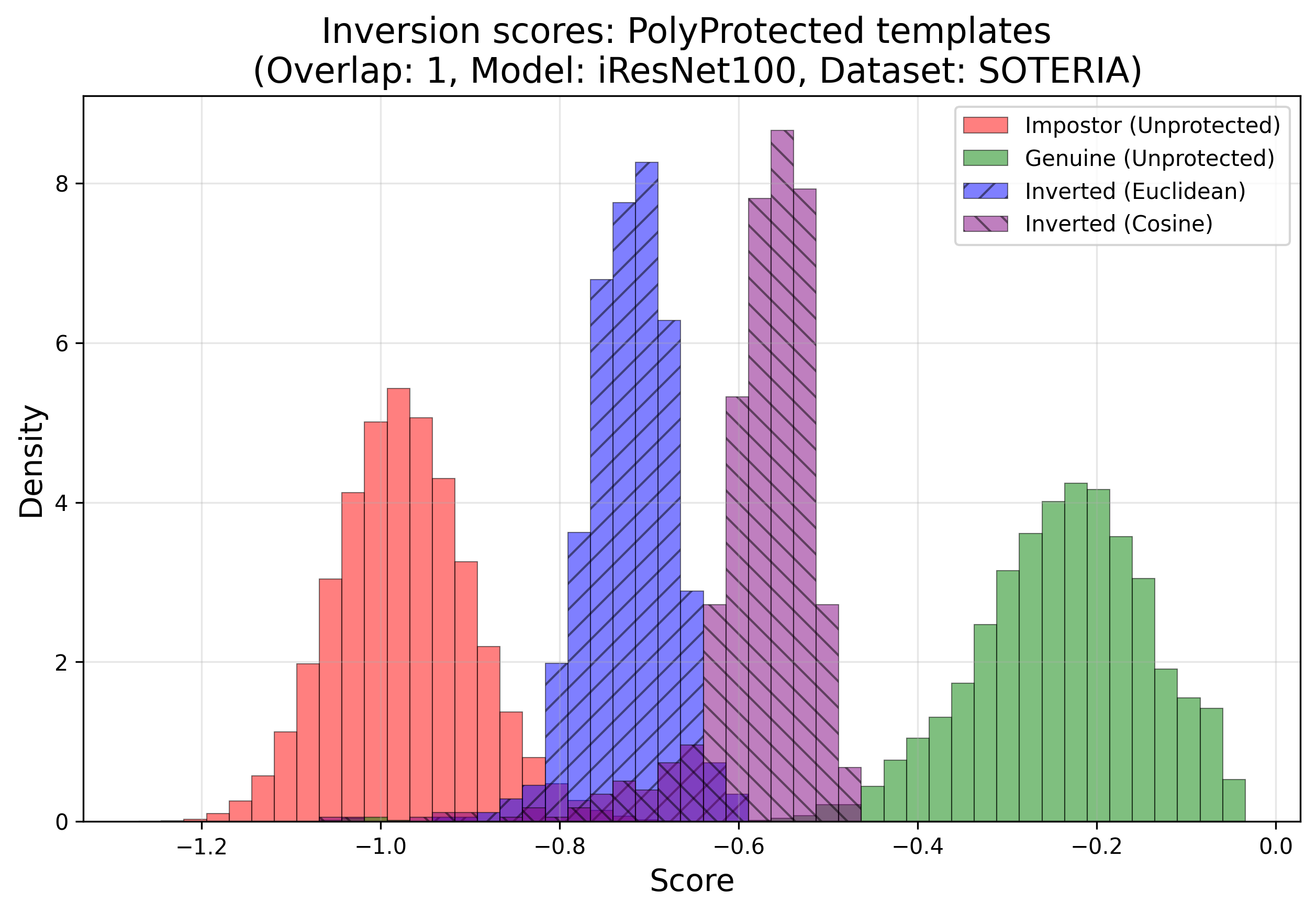}}
\hfil
\subfloat{\includegraphics[width=0.25\paperwidth]{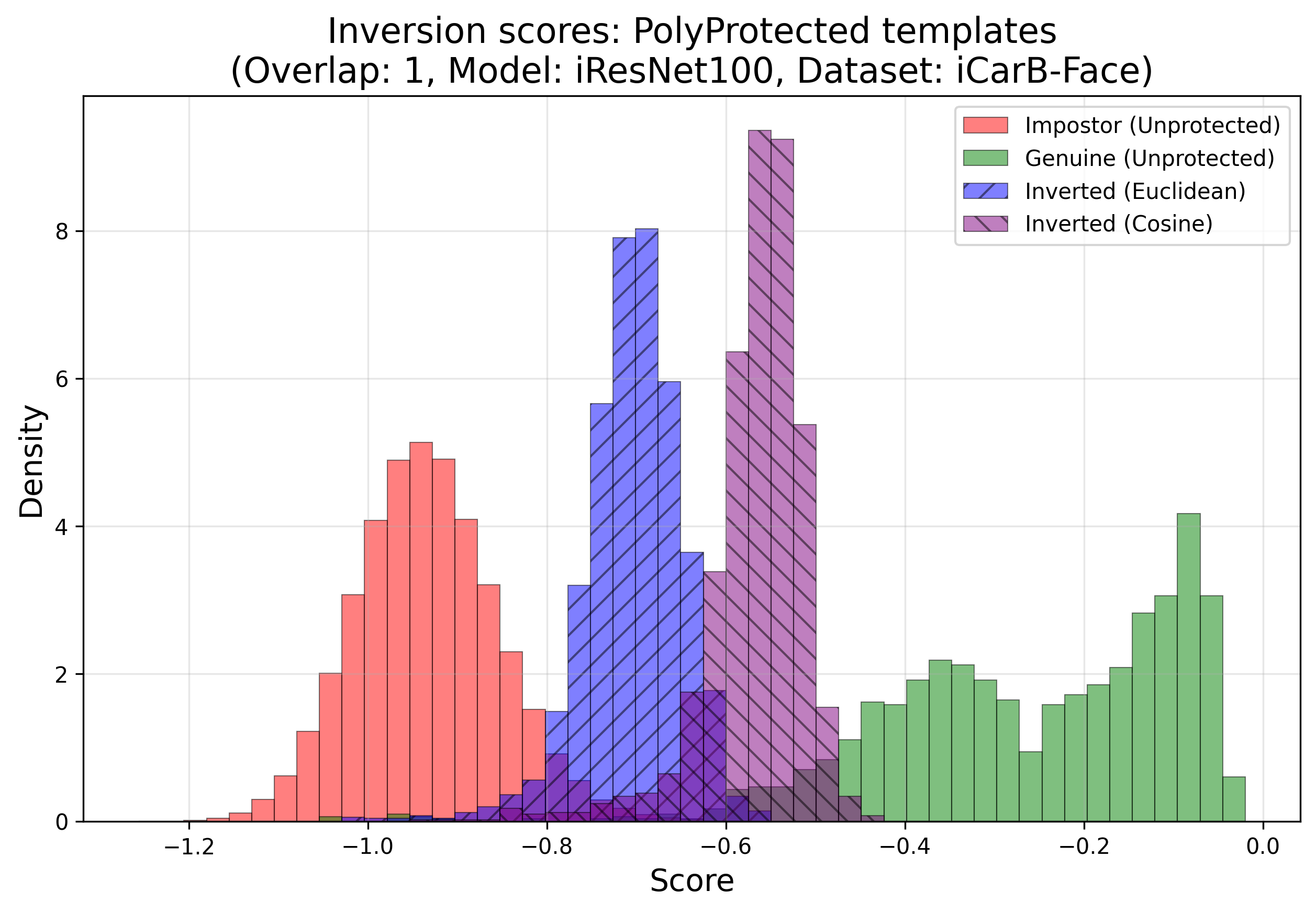}}
\vfil
\subfloat{\includegraphics[width=0.25\paperwidth]{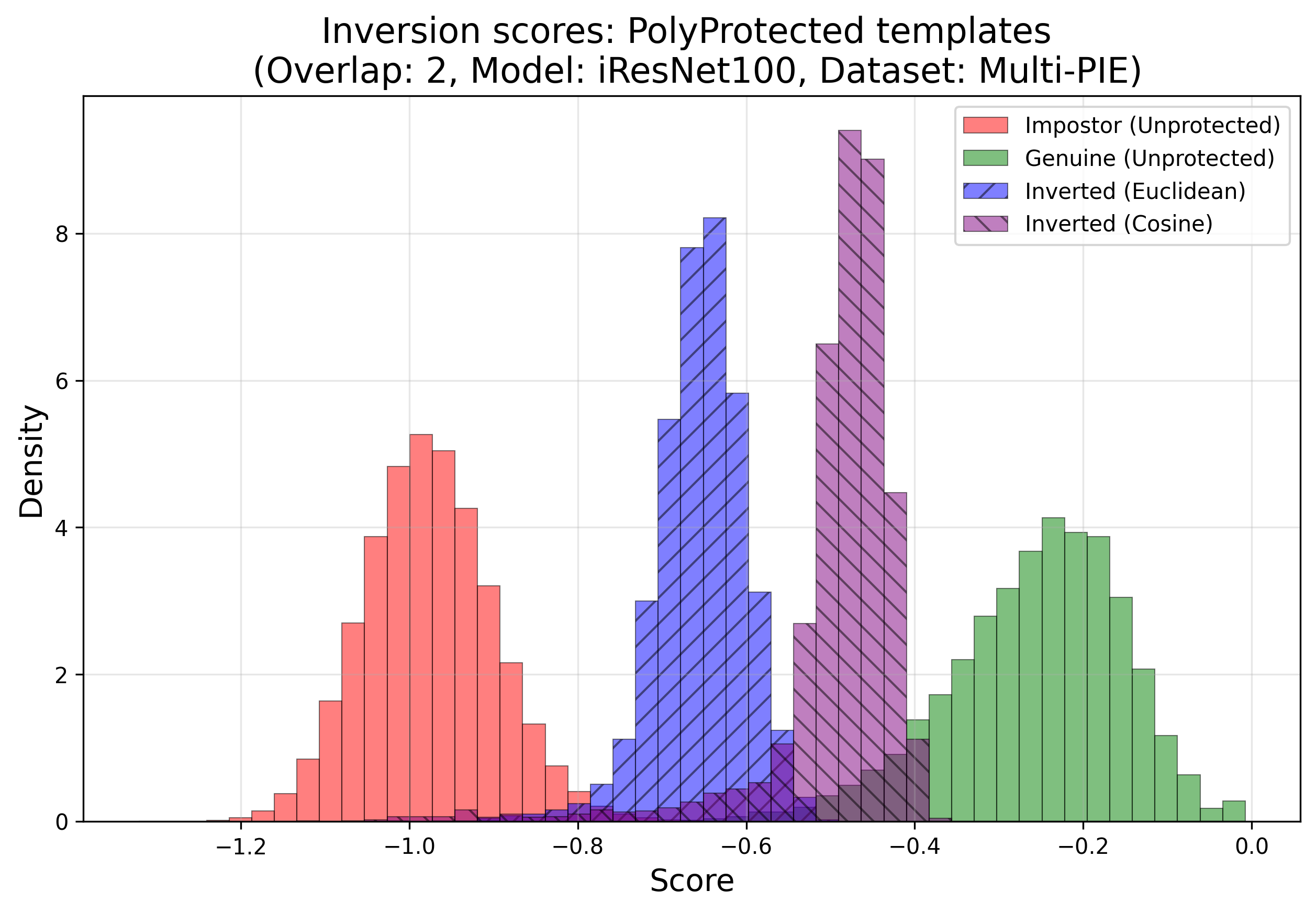}}
\hfil
\subfloat{\includegraphics[width=0.25\paperwidth]{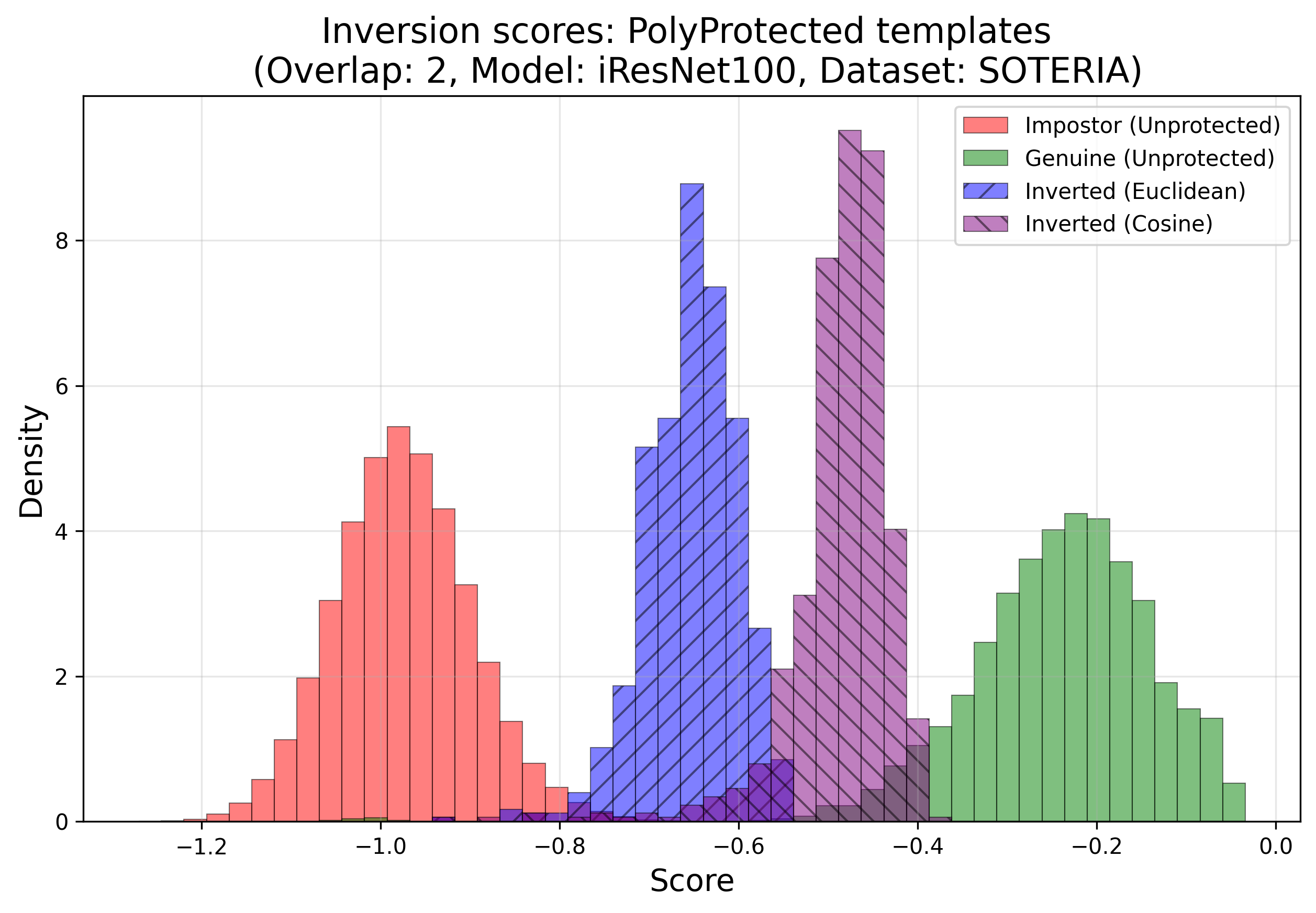}}
\hfil 
\subfloat{\includegraphics[width=0.25\paperwidth]{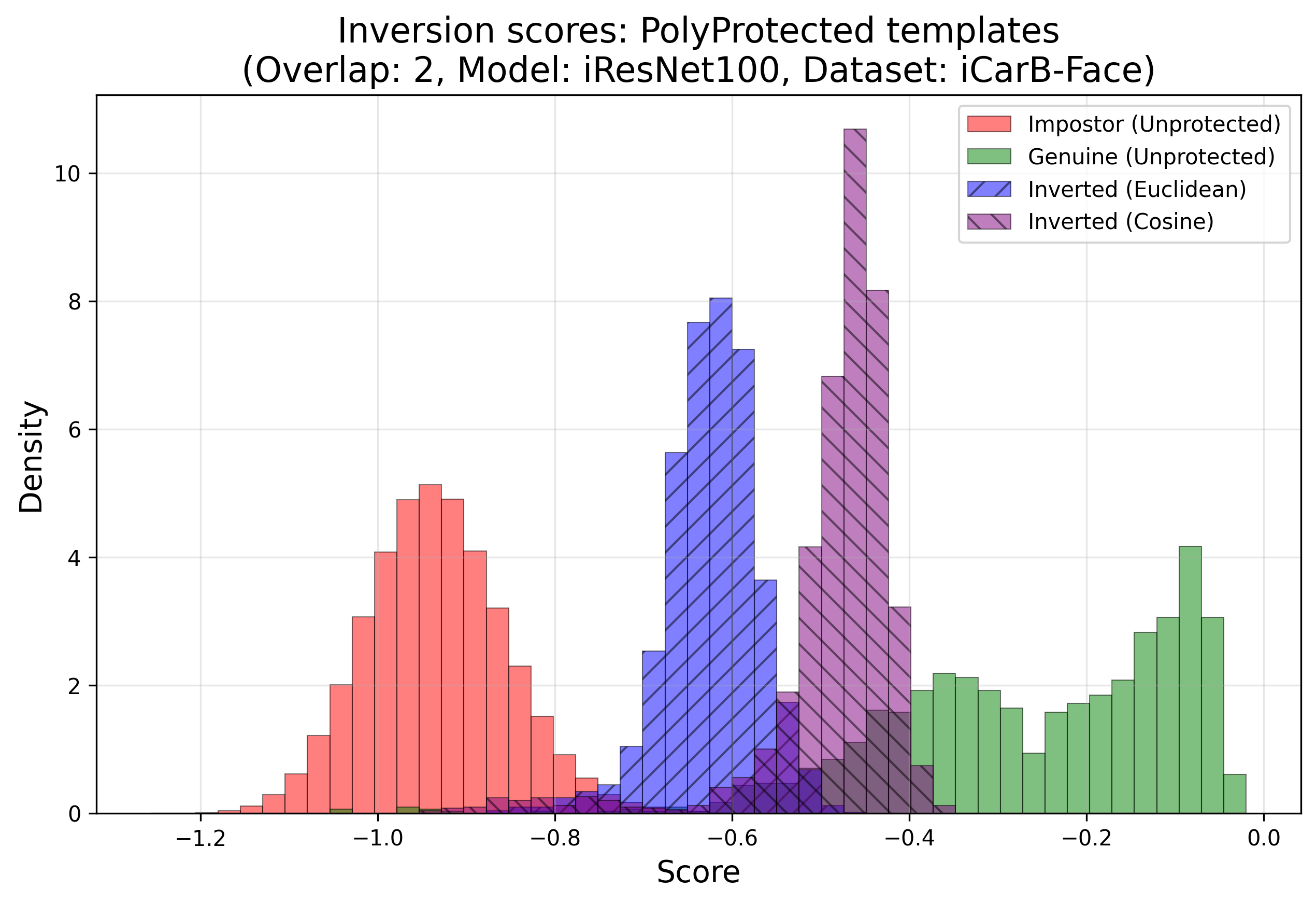}}
\vfil
\subfloat{\includegraphics[width=0.25\paperwidth]{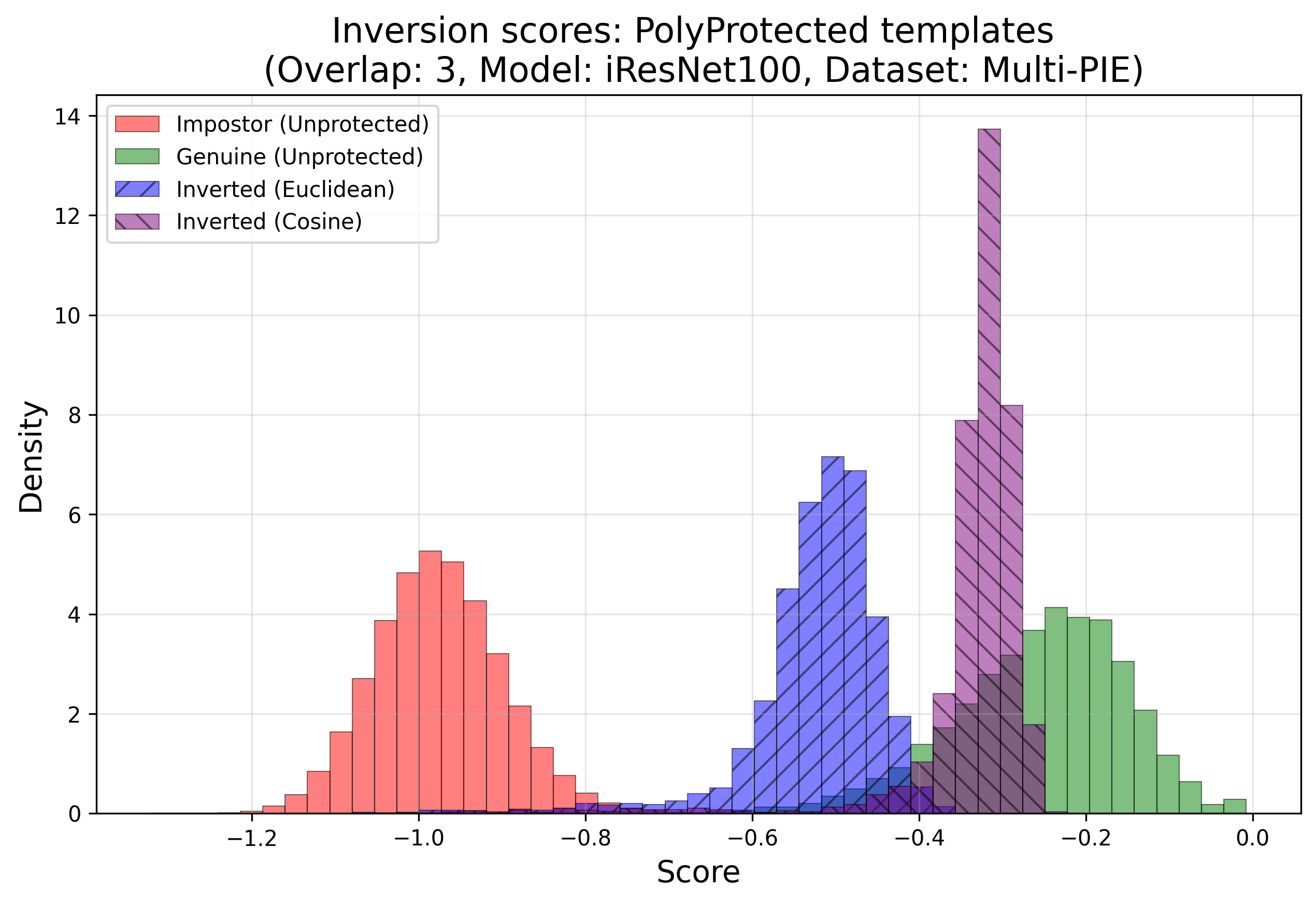}}
\hfil
\subfloat{\includegraphics[width=0.25\paperwidth]{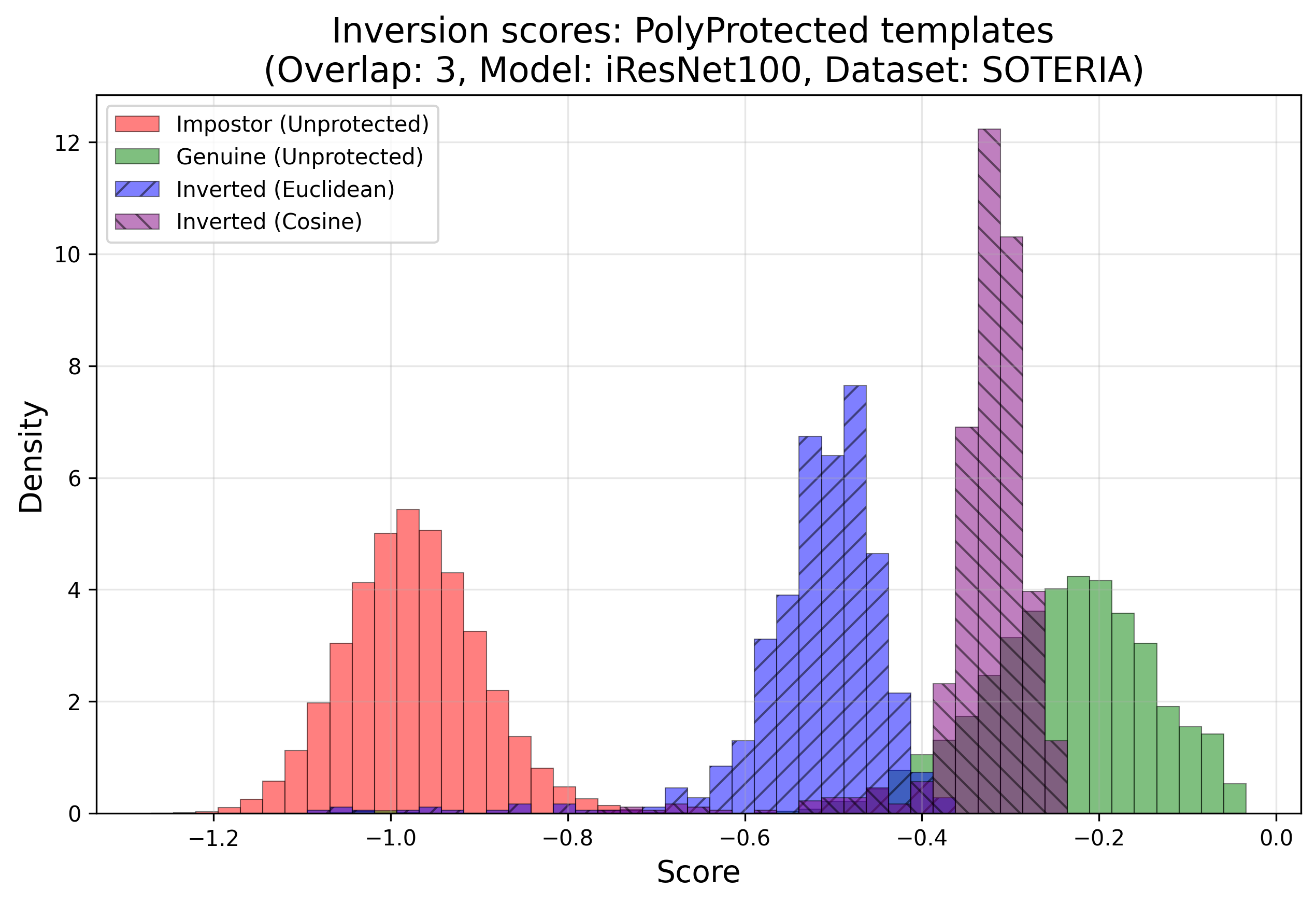}}
\hfil 
\subfloat{\includegraphics[width=0.25\paperwidth]{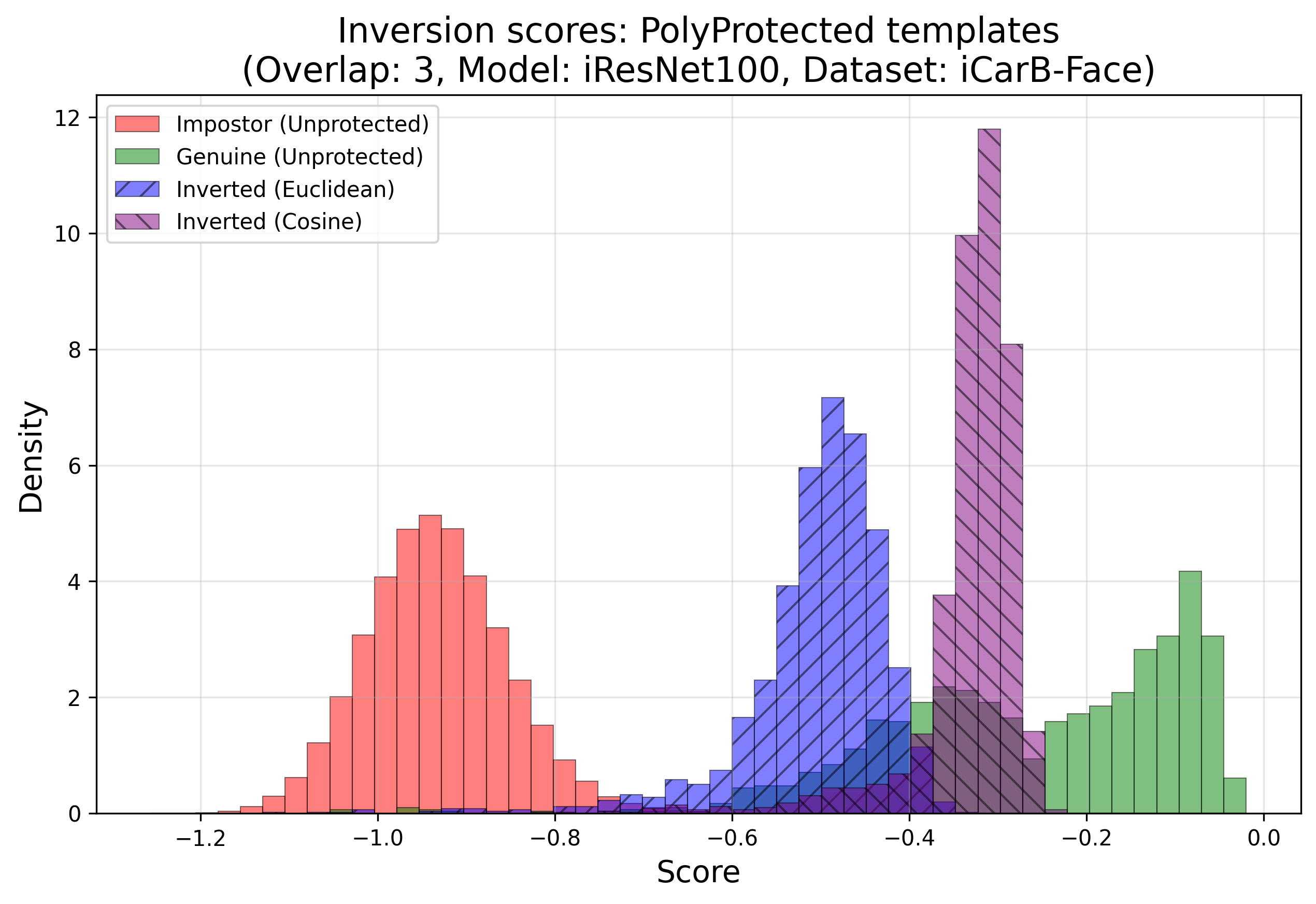}}
\vfil
\subfloat{\includegraphics[width=0.25\paperwidth]{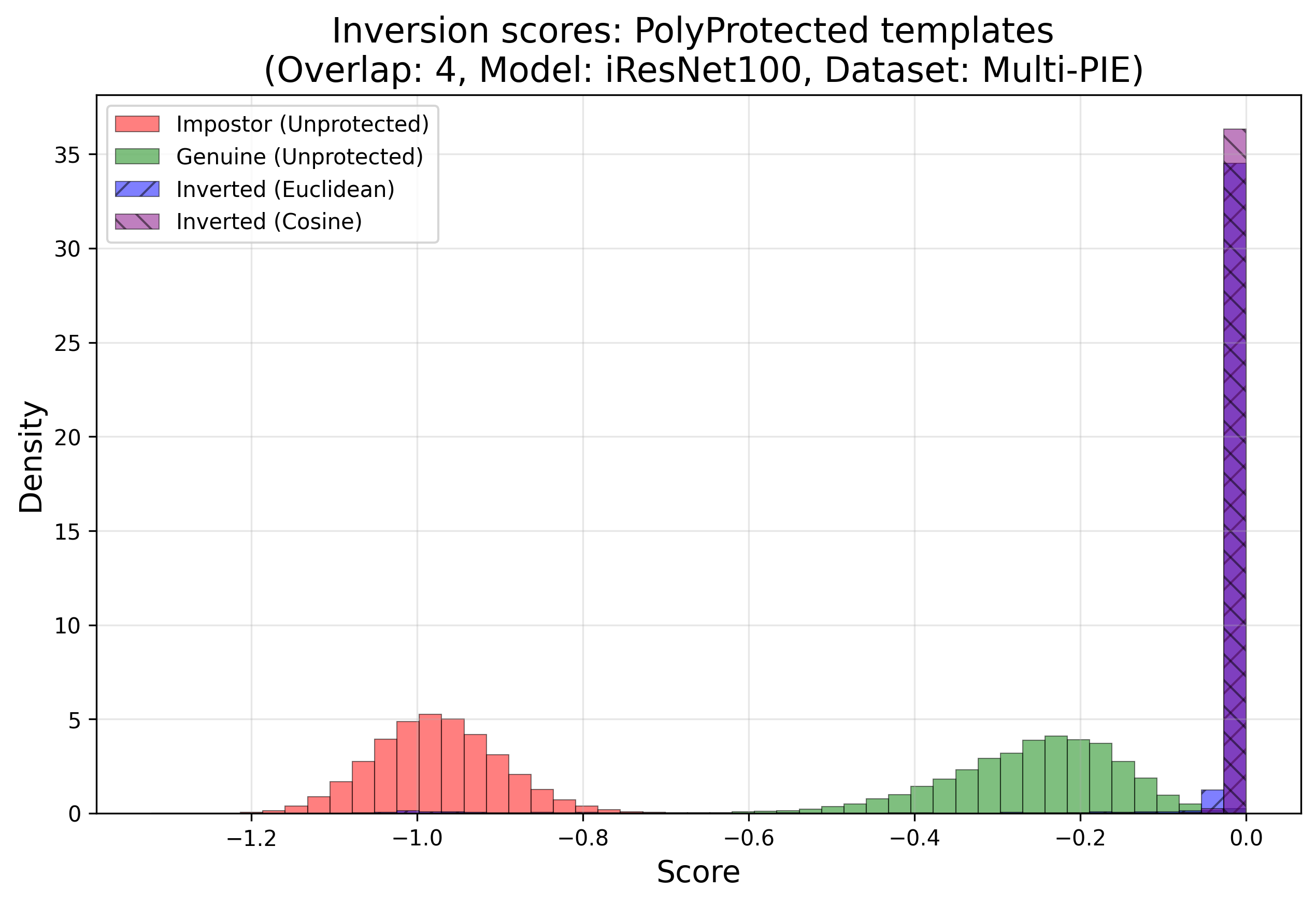}}
\hfil 
\subfloat{\includegraphics[width=0.25\paperwidth]{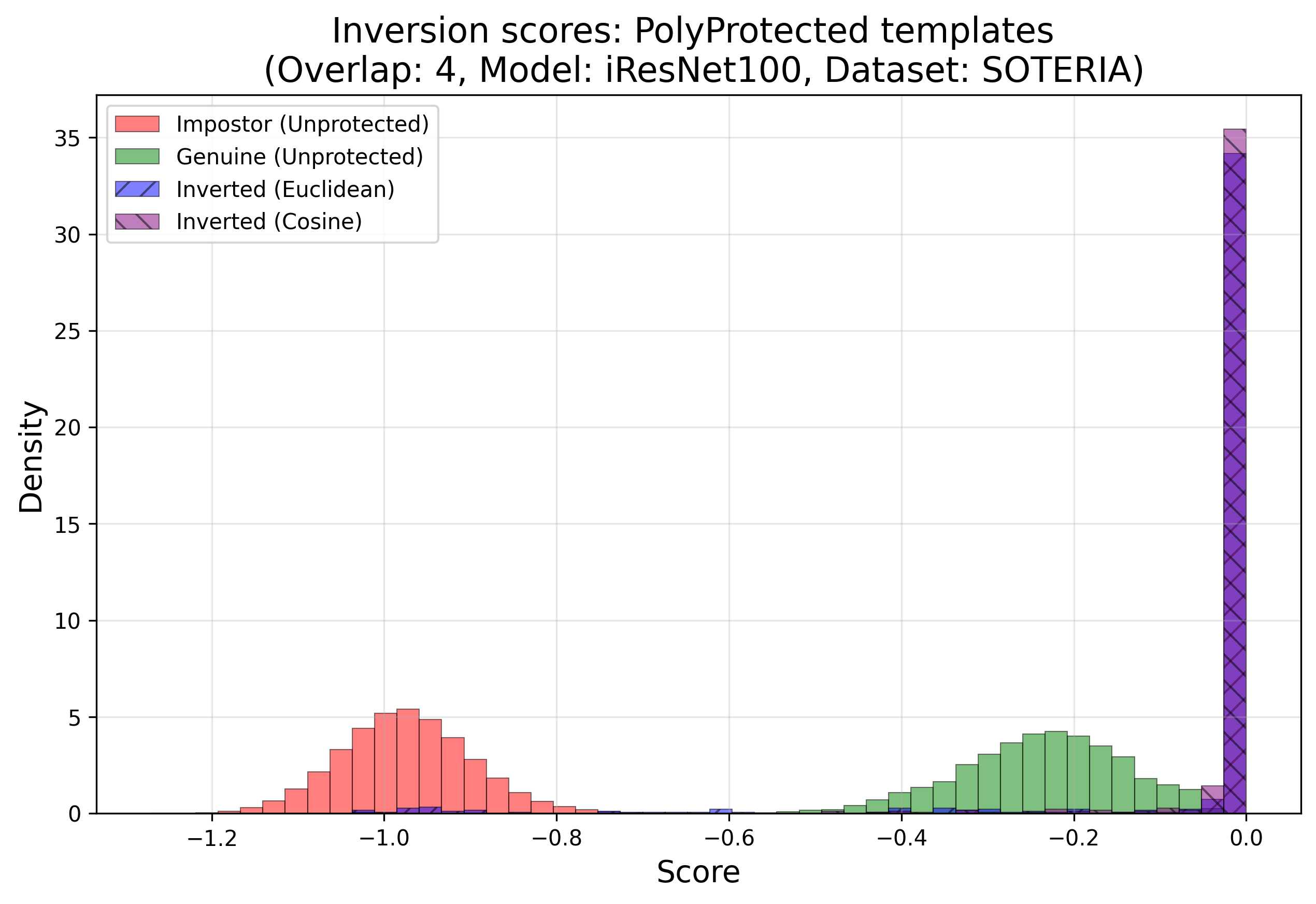}}
\hfil
\subfloat{\includegraphics[width=0.25\paperwidth]{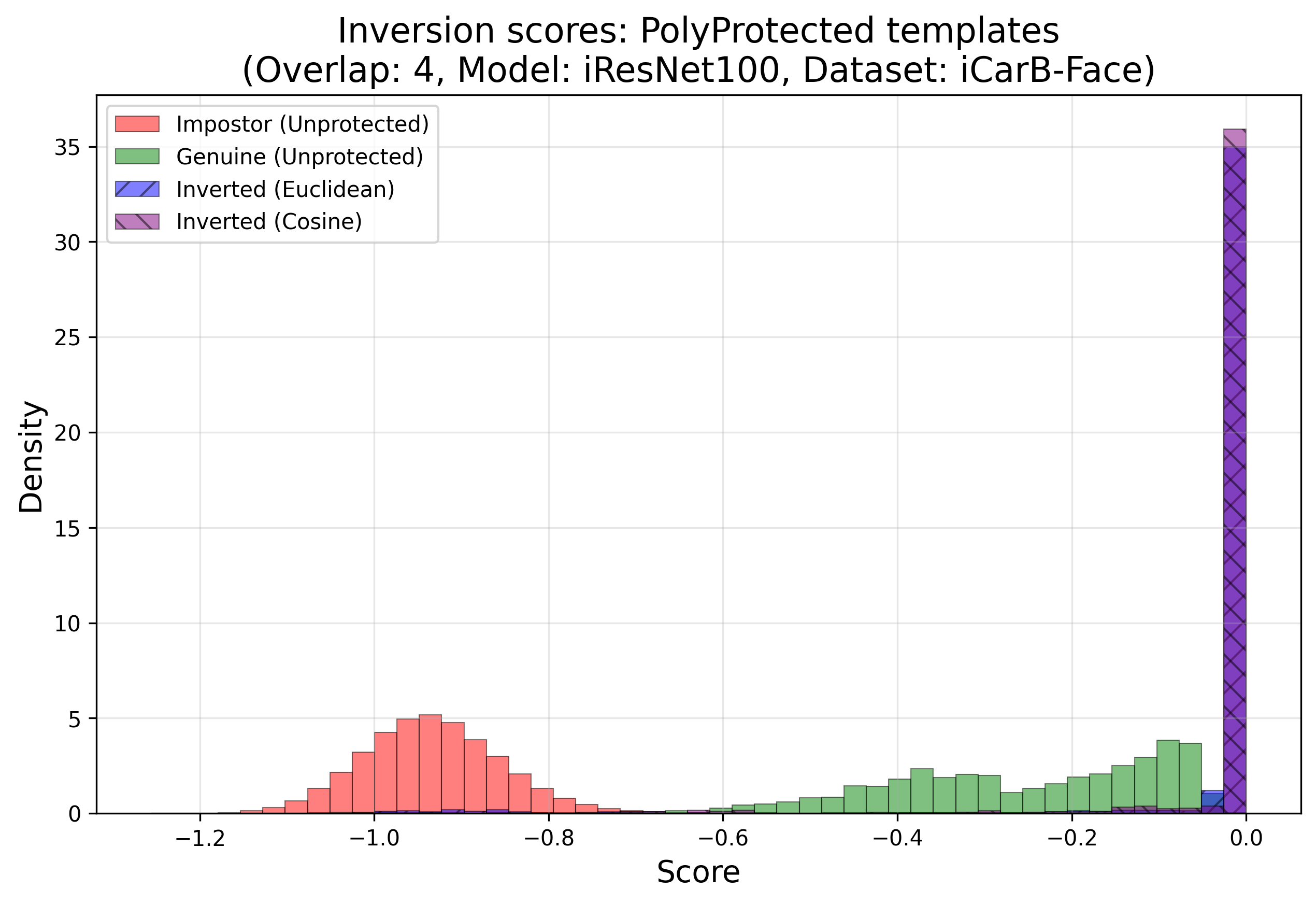}}
\caption{Inversion scores for PolyProtected templates generated from normalized iResNet100 face embeddings using different amounts of overlap, when the inversion is performed using two different numerical solvers: one based on Euclidean distance, and the other one based on cosine distance.  The closer the inversion scores are to the unprotected template genuine distribution, the more effective the inversion attack.}
\label{fig:solvers}
\end{figure*}

Fig. \ref{fig:solvers} shows histograms representing the inversion scores (i.e., cosine distances\footnote{Multiplied by -1 to turn them into \textit{similarity} scores.}) between the inverted iResNet100 templates (i.e., solutions for $V$ found by each solver, $V^{*}$), and the true $V$s from which the corresponding $P$s (i.e., the templates being inverted) were created.  These histograms were generated by launching an inversion attack on each protected template 10 times, using 10 different initial guesses for the solvers, and concatenating the resulting inversion scores.  Each inversion score indicates how close, in terms of cosine distance, the inverted template is to the original face embedding.  The figures also include histograms for the genuine and impostor scores (also cosine distances) computed on the unprotected face embeddings.  Ideally, the inversion scores should lie as close as possible to the impostor distribution, which would indicate that the inverted templates are as different from their corresponding face embeddings as are embeddings from different identities.

From Fig. \ref{fig:solvers}, it is clear that, in general, the numerical solver based on cosine distance results in higher inversion scores than the solver from \cite{kh22}, which is based on Euclidean distance.  In other words, the templates inverted using the cosine solver tend to be more similar (lower cosine distance) to their corresponding face embeddings.  This is good for the attacker, since it would allow them to more easily use the inverted template to impersonate the underlying identity in the unprotected face recognition system.  On the other hand, this is bad for PolyProtect, because its irreversibility is \textit{lower} with this new solver than with the Euclidean solver employed in \cite{kh22}.  Having said that, it is important to note that the solver based on Euclidean distance actually represents a ``truer'' picture of irreversibility, since this type of solver tries to recover both the \textit{magnitude} and \textit{direction} of the underlying $n$-dimensional embedding, whereas the cosine solver attempts to recover only the \textit{direction} of the embedding.  However, since the success of an inversion attack is judged in terms of cosine distance, technically the \textit{direction} of the embedding is all we need to recover.  So, it makes sense that, for this particular definition of a successful inversion, the \textit{worst-case} irreversibility evaluation should be based on the cosine solver.       

\begin{figure*}[!h]
\centering
\subfloat{\includegraphics[width=0.25\paperwidth]{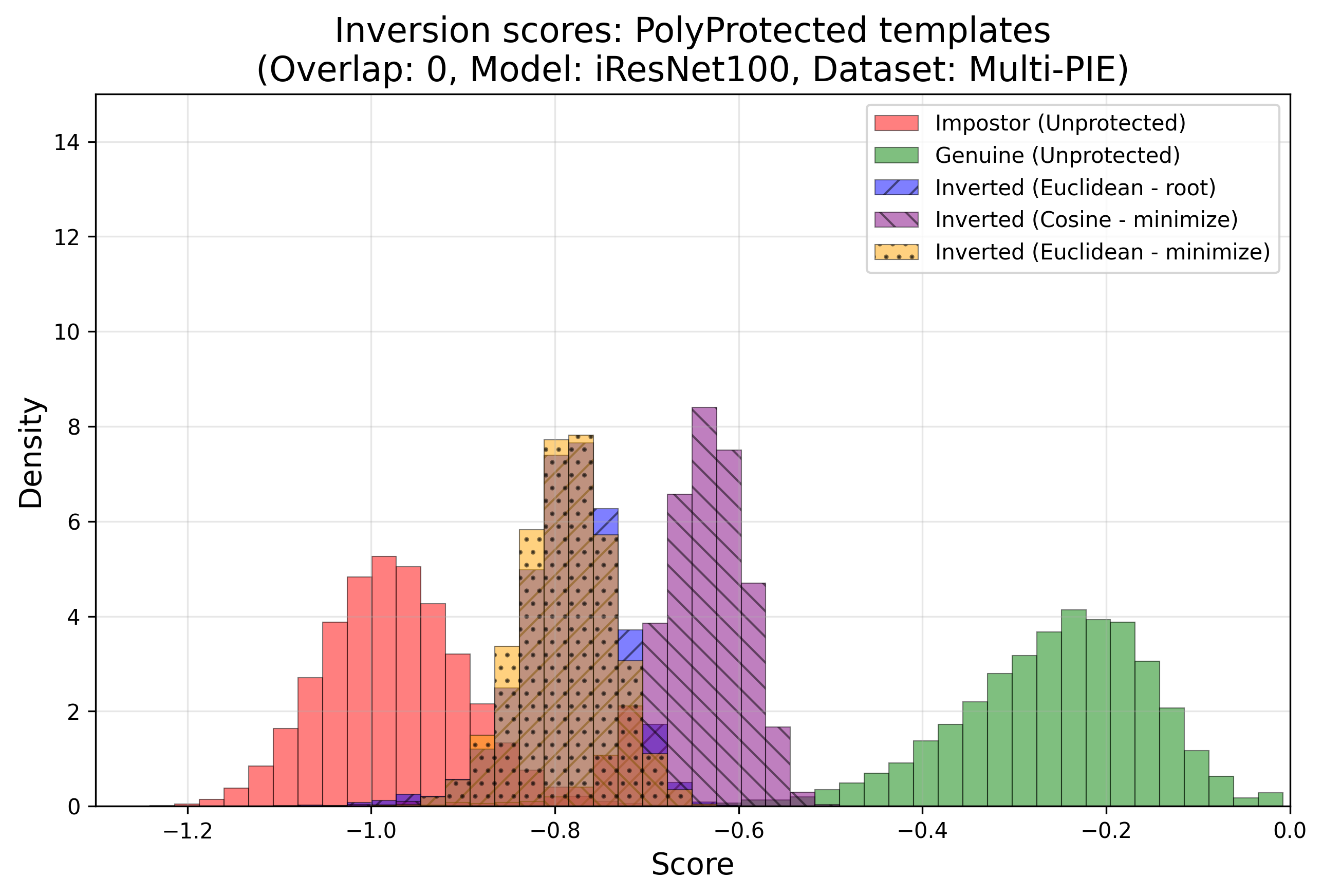}}
\hfil 
\subfloat{\includegraphics[width=0.25\paperwidth]{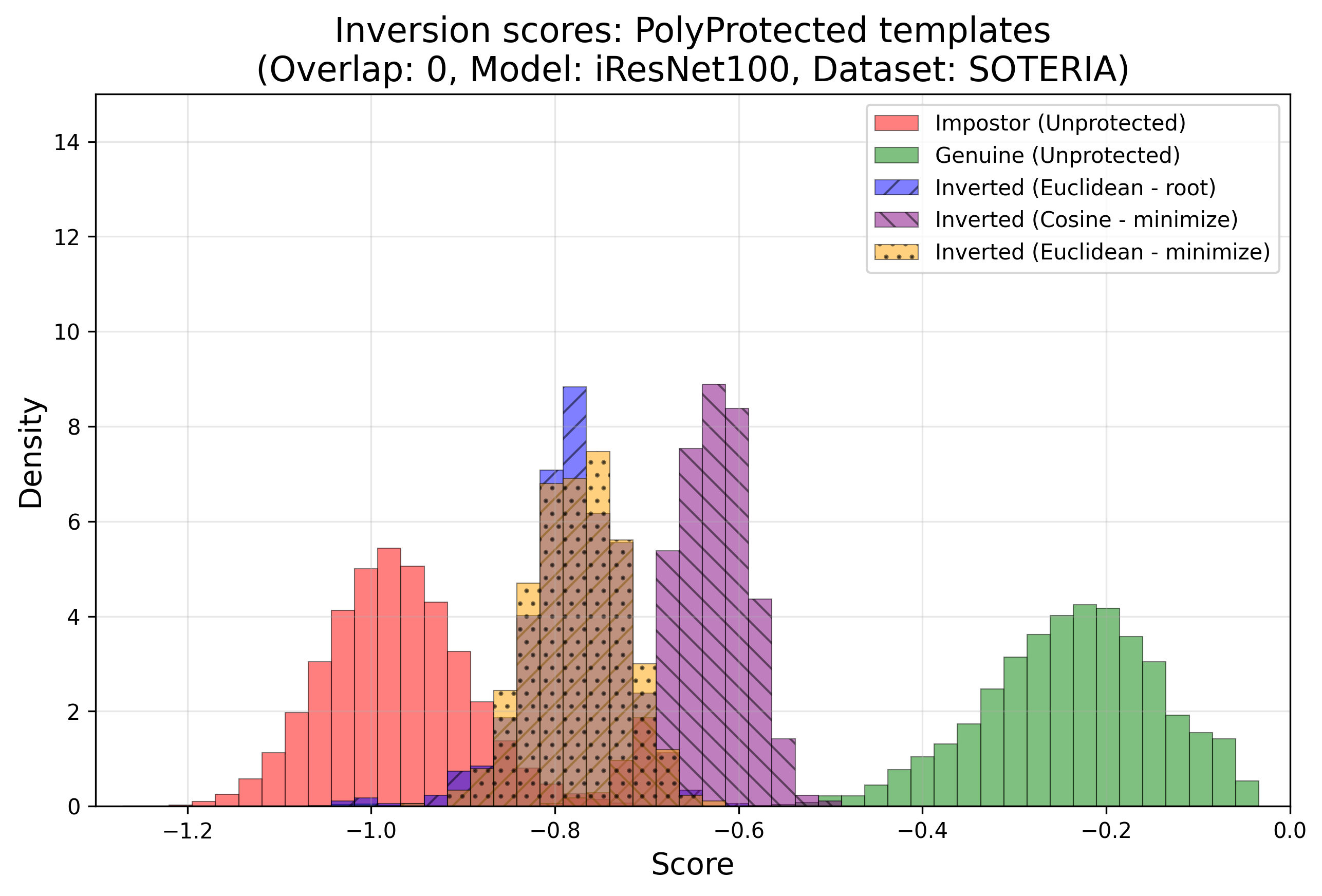}}
\hfil
\subfloat{\includegraphics[width=0.25\paperwidth]{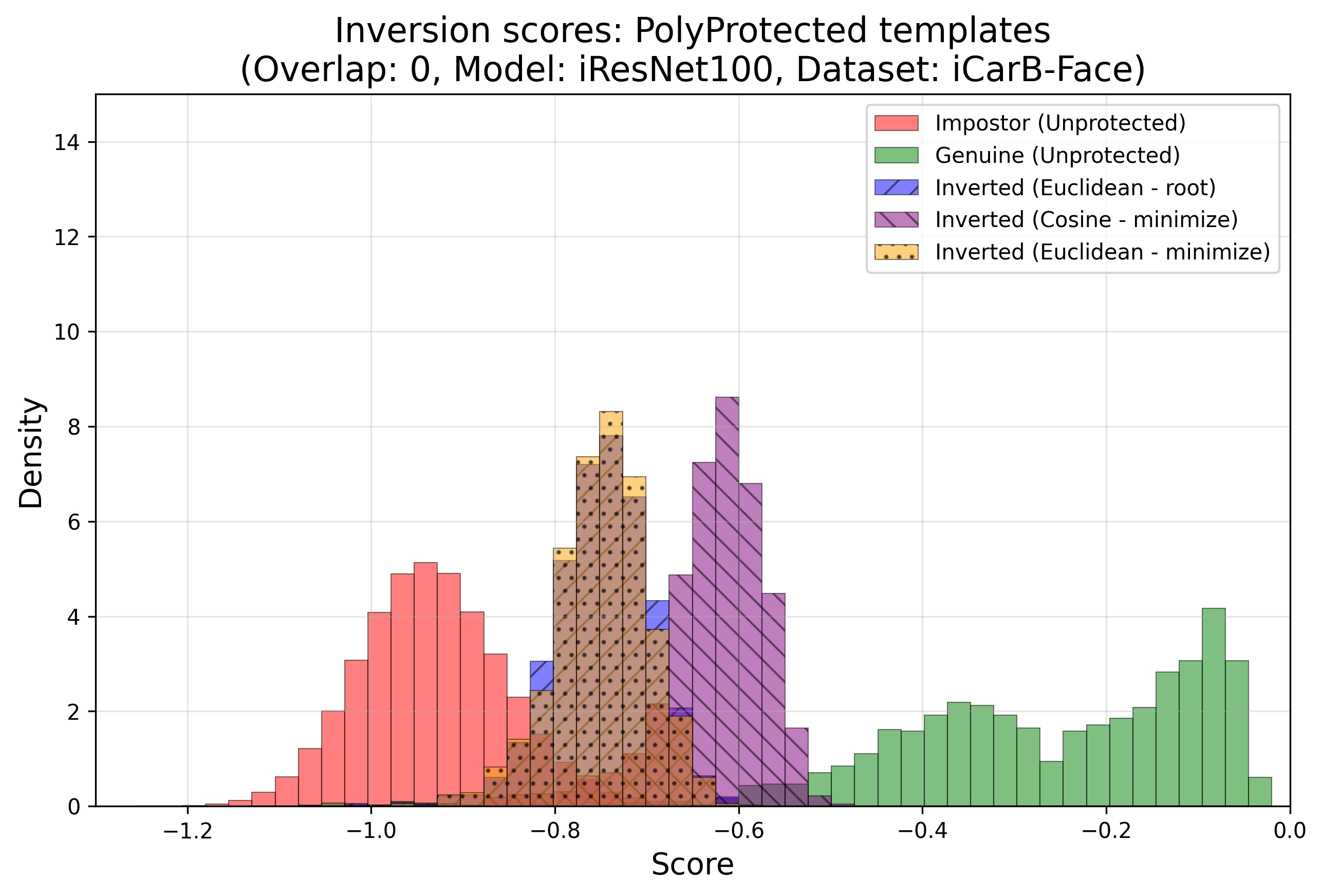}}
\vfil
\subfloat{\includegraphics[width=0.25\paperwidth]{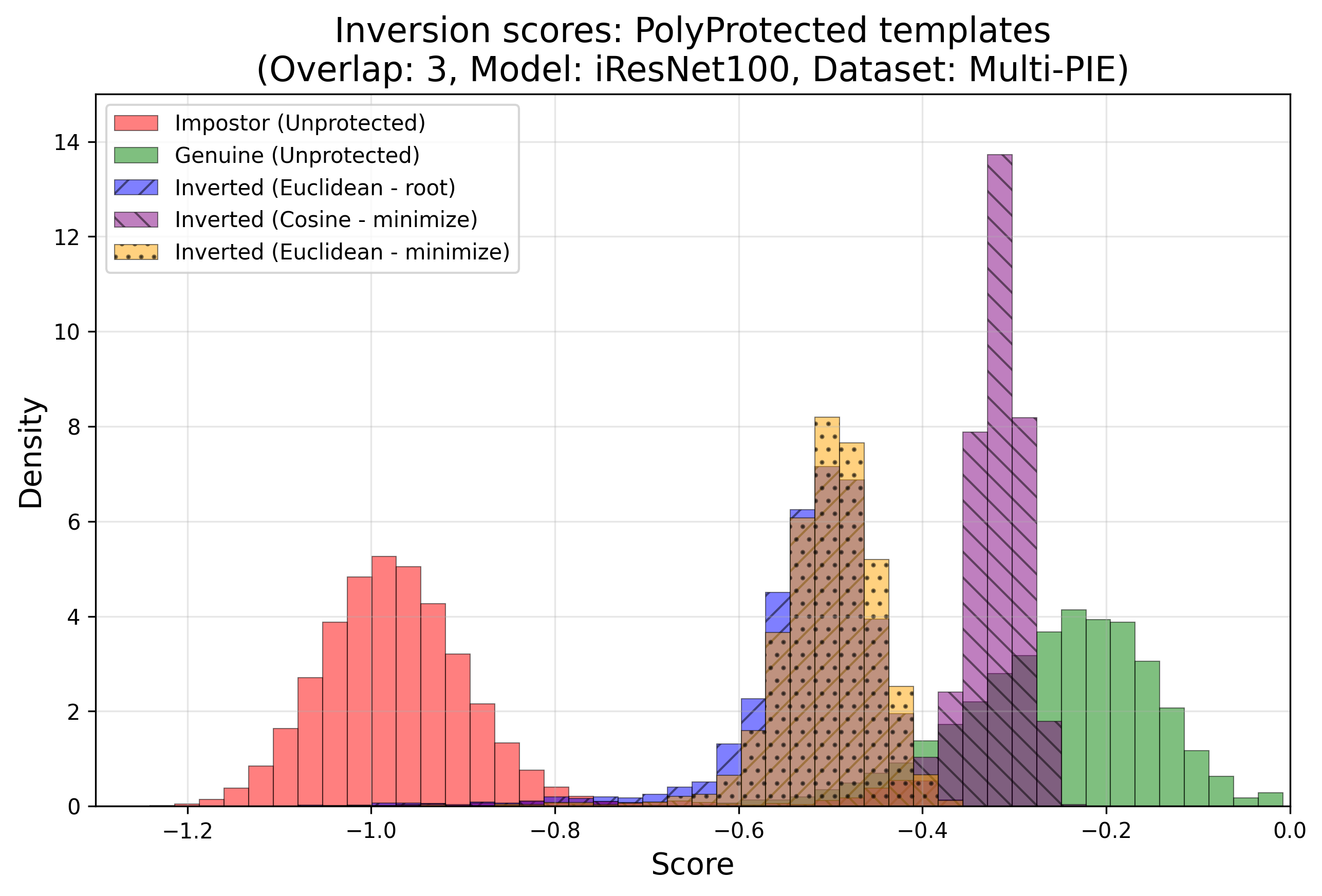}}
\hfil
\subfloat{\includegraphics[width=0.25\paperwidth]{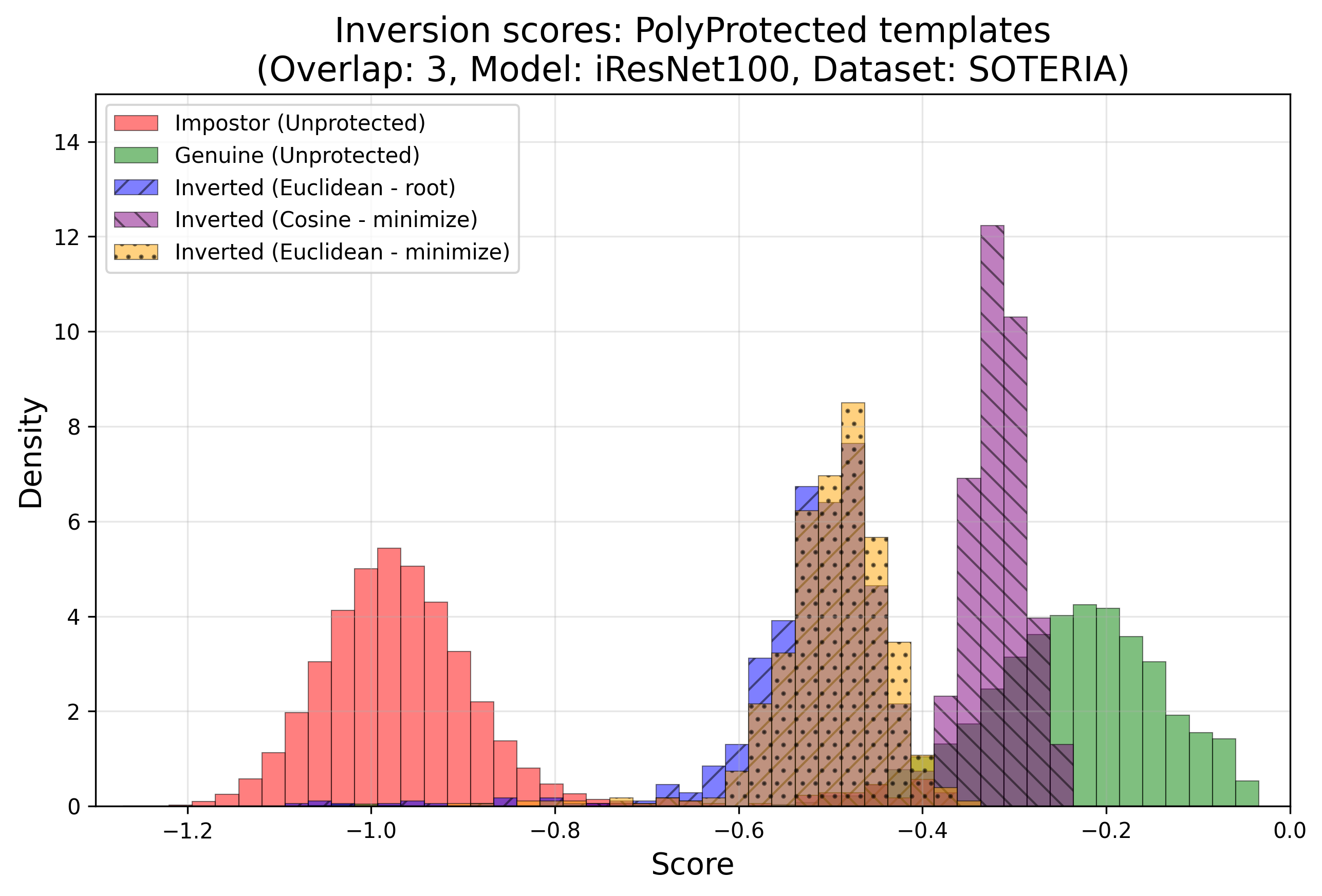}}
\hfil
\subfloat{\includegraphics[width=0.25\paperwidth]{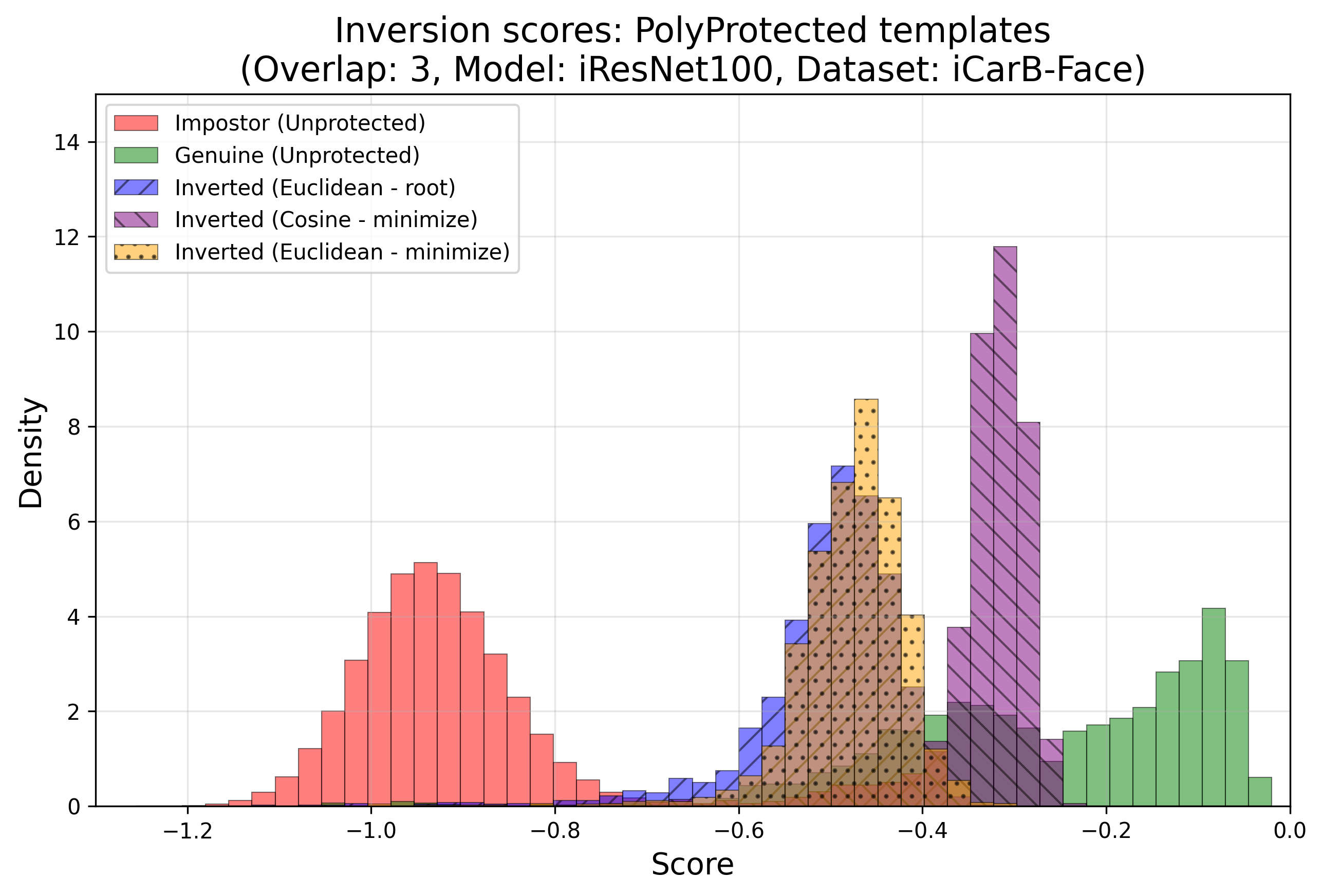}}
\caption{Inversion scores for PolyProtected templates generated from normalized iResNet100 face embeddings, when the inversion is performed using the same numerical solver, Python's \textit{scipy.optimize.minimize} function with the \textit{L-BFGS-B} method, but based on minimizing different distances: either cosine or Euclidean.  For reference, we also show inversion score histograms obtained using Python's  \textit{scipy.optimize.root} function with the \textit{lm} method, which is suitable for Euclidean distance but not for cosine distance.}
\label{fig:solver_comparison}
\end{figure*}

To demonstrate that the aforementioned improvement in the attacker's inversion capabilities when using the cosine solver can indeed be attributed to the choice of distance function (i.e., cosine rather than Euclidean), as opposed to the choice of solver itself, Fig. \ref{fig:solver_comparison} compares the inversion score histograms obtained when using the \textit{same solver} with \textit{different distance functions}, i.e., the \textit{scipy.optimize.minimize} solver with the \textit{L-BFGS-B} method, set up to minimize either the cosine distance (as in Fig. \ref{fig:solvers}) or the Euclidean distance (between the PolyProtected version of the solution, $V^{*}$, and the protected template we are trying to invert, $P$).  For reference, we also show the inversion score histograms obtained using the original solver from \cite{kh22}, i.e., \textit{scipy.optimize.root} with the \textit{lm} method, which is essentially based on Euclidean distance.  Note that this solver is not suitable for minimizing the cosine distance between $PolyProtect(V^{*})$ and $P$, because the \textit{lm} method operates on each dimension of the two vectors to try to reduce the ``error'' between them.  This makes sense when we consider the difference between the two vectors in terms of Euclidean distance, where the magnitude of each dimension matters; however, cosine distance considers the two vectors holistically and simply quantifies the difference between their overall directions in terms of a single scalar, which does not lend itself to the \textit{lm} formulation.  Due to space constraints, Fig. \ref{fig:solver_comparison} shows results only for iResNet100 and two overlap parameters (0 and 3), but the same observations can be made for EdgeFace and other overlaps. 

From Fig. \ref{fig:solver_comparison}, it is clear that it is the choice of \textit{distance} function, not the choice of solver, which determines the feasibility of the inversion attack.  More specifically, for the same solver (\textit{scipy.optimize.minimize}), the inversion scores are higher when minimizing \textit{cosine} distance as opposed to Euclidean distance (i.e., the purple histogram is further to the right than the yellow, dotted histogram).  Furthermore, we see that Euclidean-distance-based inversion scores remain approximately the same regardless of whether we use the \textit{scipy.optimize.minimize} or the \textit{scipy.optimize.root} solver.  So, we stand by our earlier conclusion that a numerical solver based on cosine distance allows for more successful inversion attacks than a solver based on Euclidean distance.

We now return to Fig. \ref{fig:solvers} for another observation, in addition to the one made earlier regarding the two solvers.  Fig. \ref{fig:solvers} also shows that inversion scores increase as the amount of overlap (used in the PolyProtect transform) increases (i.e., the inversion score histograms shift to the right).  This means that PolyProtected templates generated using larger overlaps are easier to invert than those generated using smaller overlaps.  This trend was already observed in the earlier PolyProtect work \cite{kh22, s25}, and is due to the same reason as that used to explain why larger overlaps lead to better accuracy in the protected domain (Section \ref{sec:accuracy_eval}), i.e., because larger overlaps result in more information about the original embedding being retained in the PolyProtected template, which, in the context of irreversibility, makes it easier to recover the embedding from its protected template.  So, we confirm the observation from \cite{kh22, s25}, that there is a trade-off between the irreversibility and accuracy of PolyProtected templates depending on the amount of overlap applied in the transform, i.e., a larger overlap results in higher recognition accuracy but lower irreversibility, and vice-versa.  In Section \ref{sec:key_selection}, we will demonstrate that this trade-off can be effectively mitigated using our new key selection algorithm, which also serves to significantly improve the irreversibility of PolyProtected templates even in the worst-case scenario where a numerical solver based on cosine distance is used for the inversion attack.

\section{Improving the Irreversibility of PolyProtect: Key Selection Algorithm}
\label{sec:key_selection}

In the PolyProtect evaluations presented thus far, the coefficients, $C$, and exponents, $E$, used to parameterise the PolyProtect polynomials, were \textit{randomly} generated for each subject (identity).  In this section, we propose an alternative method for selecting these subject-specific ``keys'', with the aim of generating PolyProtected templates that are more difficult to invert.  In particular, we propose a \textit{key selection algorithm}, which works as follows.  Assume we have a database of reference templates (embeddings) that we wish to protect.  The goal of the key selection algorithm is to choose $C$ and $E$ for those templates, such that their PolyProtected versions are irreversible under a numerical solver.  We assume that the cosine-distance-based solver from Section \ref{sec:irreversibility_eval} is used, since this represents the worst-case scenario.  So, for a particular subject's reference template, our algorithm begins by generating $C$ and $E$ randomly, as before.  These parameters are used to transform the template to its PolyProtected counterpart, and an inversion attack using the cosine solver is launched.  If the inversion is successful, new $C$ and $E$ parameters are generated, and the process is repeated until the inversion \textit{fails}.  The threshold used to define a successful inversion is set to a value well beyond anything likely to be used in a practical face recognition system -- in our case, the threshold was set at 20\% FMR\footnote{Alternative thresholds are explored in Section \ref{subsec:ks_thresh}.} -- because this way, if the inversion fails for such a loose threshold, we may expect it to fail for all stricter (more practical) thresholds (e.g., at 0.1\% or 0.01\% FMR).

\begin{figure*}[!h]
\centering
\subfloat{\includegraphics[width=0.25\paperwidth]{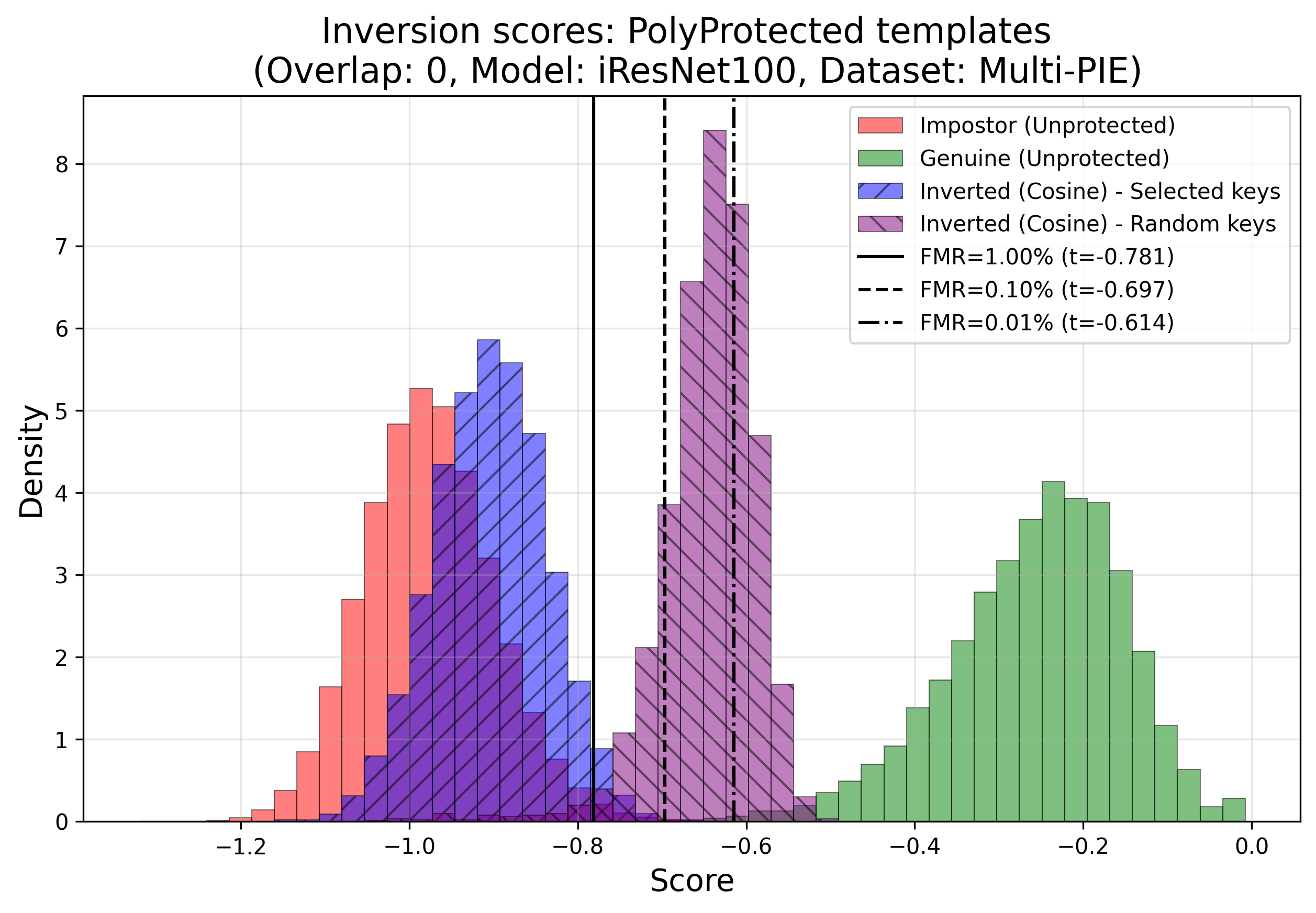}}
\hfil 
\subfloat{\includegraphics[width=0.25\paperwidth]{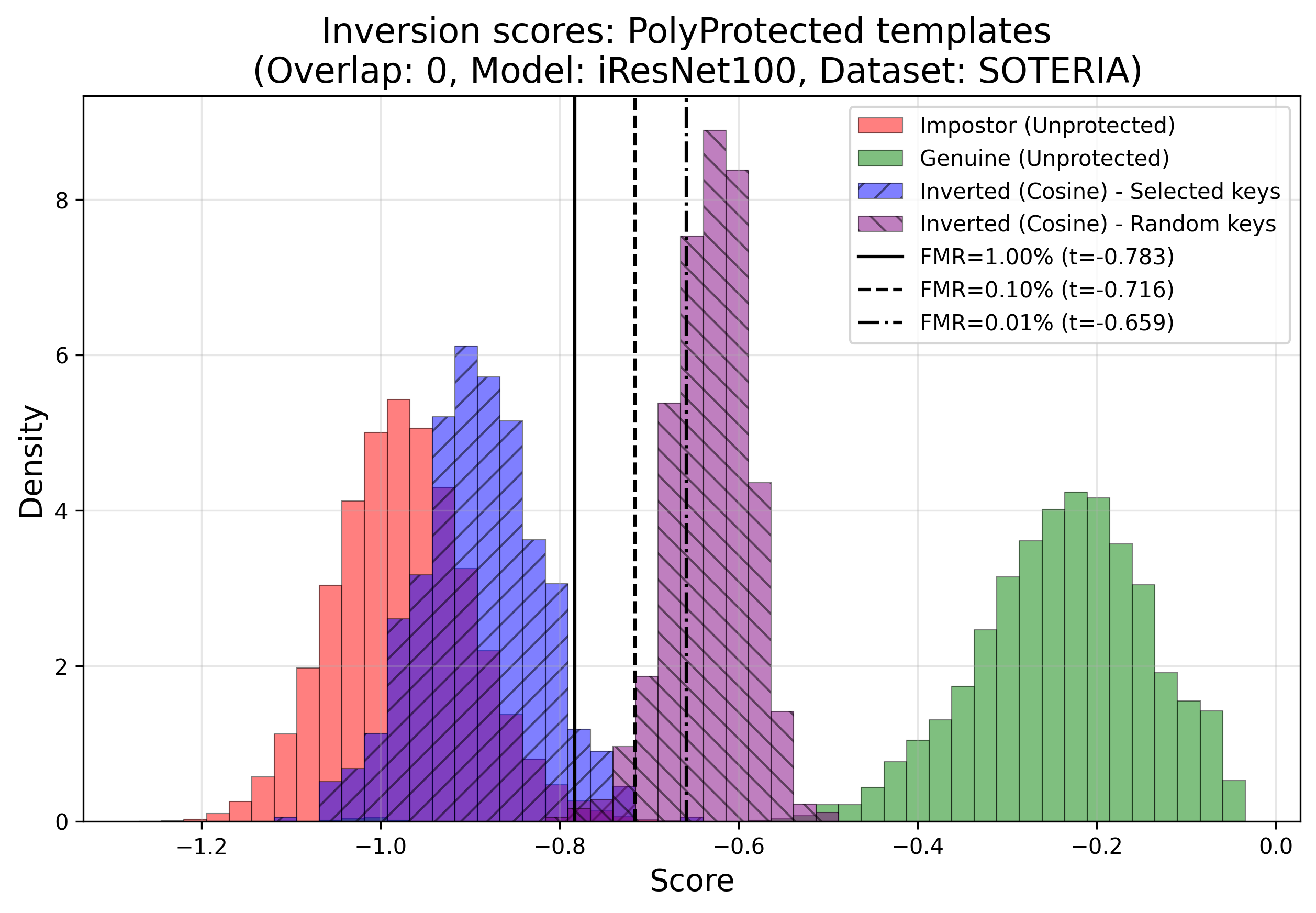}}
\hfil 
\subfloat{\includegraphics[width=0.25\paperwidth]{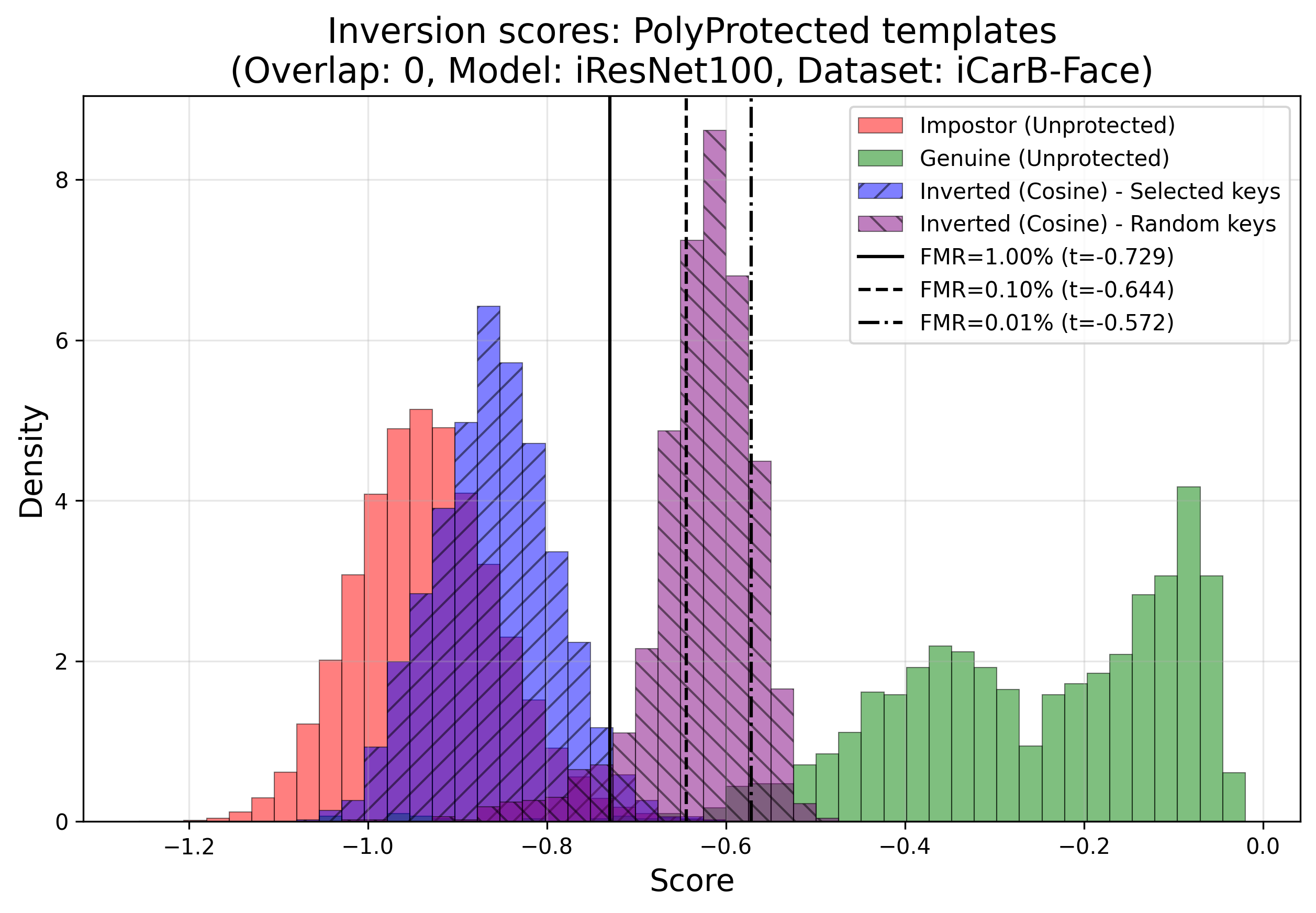}}
\vfil
\subfloat{\includegraphics[width=0.25\paperwidth]{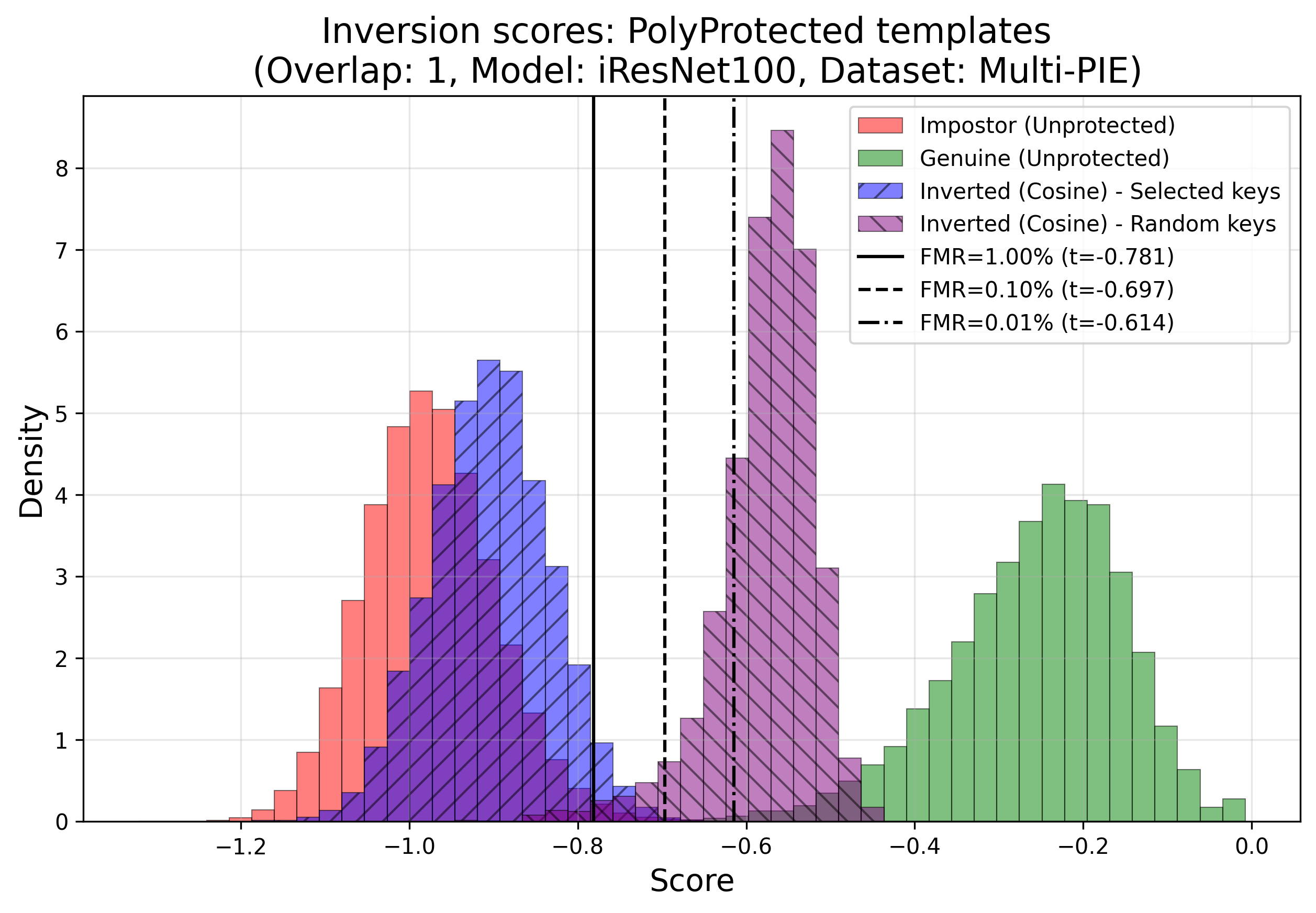}}
\hfil
\subfloat{\includegraphics[width=0.25\paperwidth]{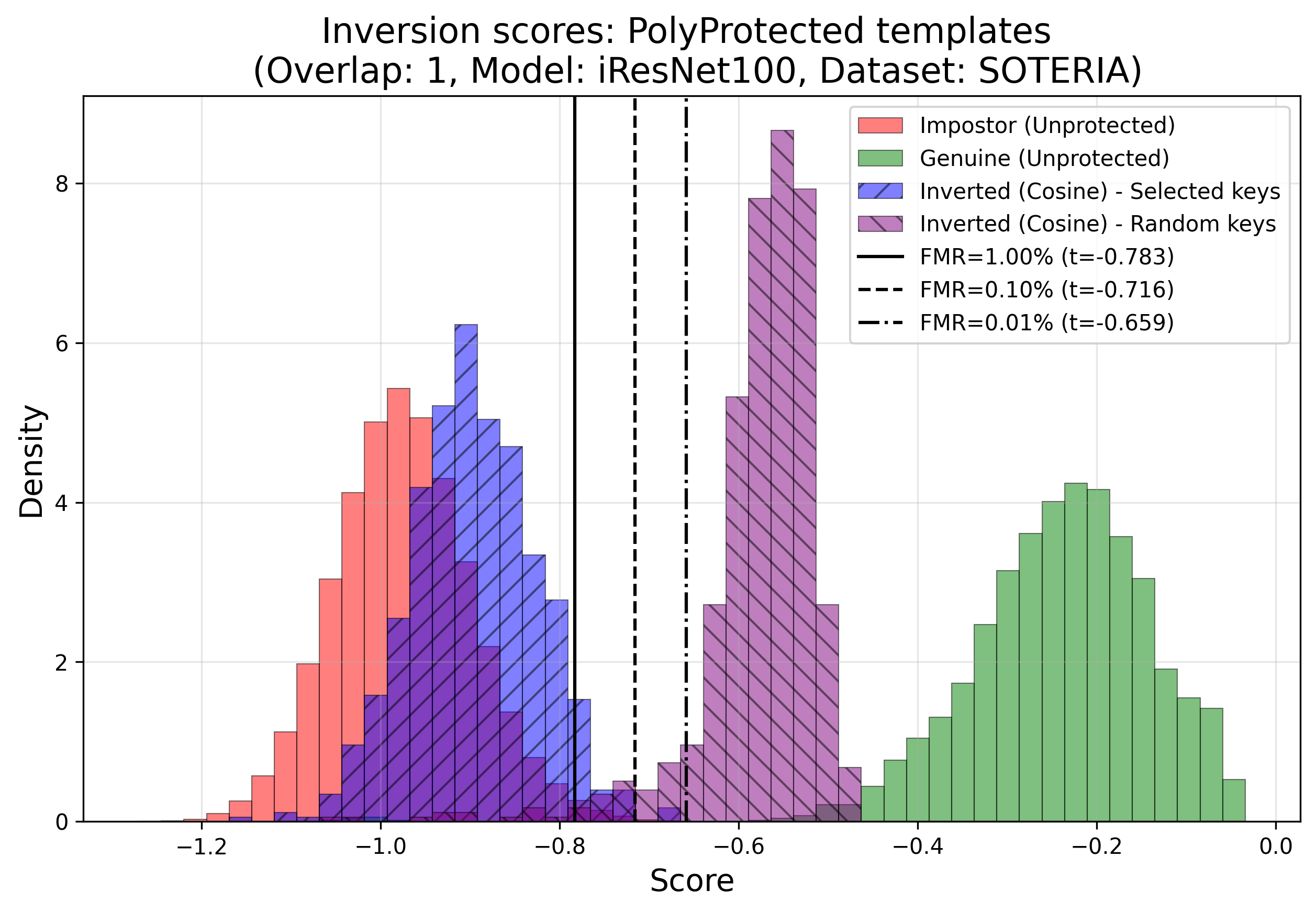}}
\hfil
\subfloat{\includegraphics[width=0.25\paperwidth]{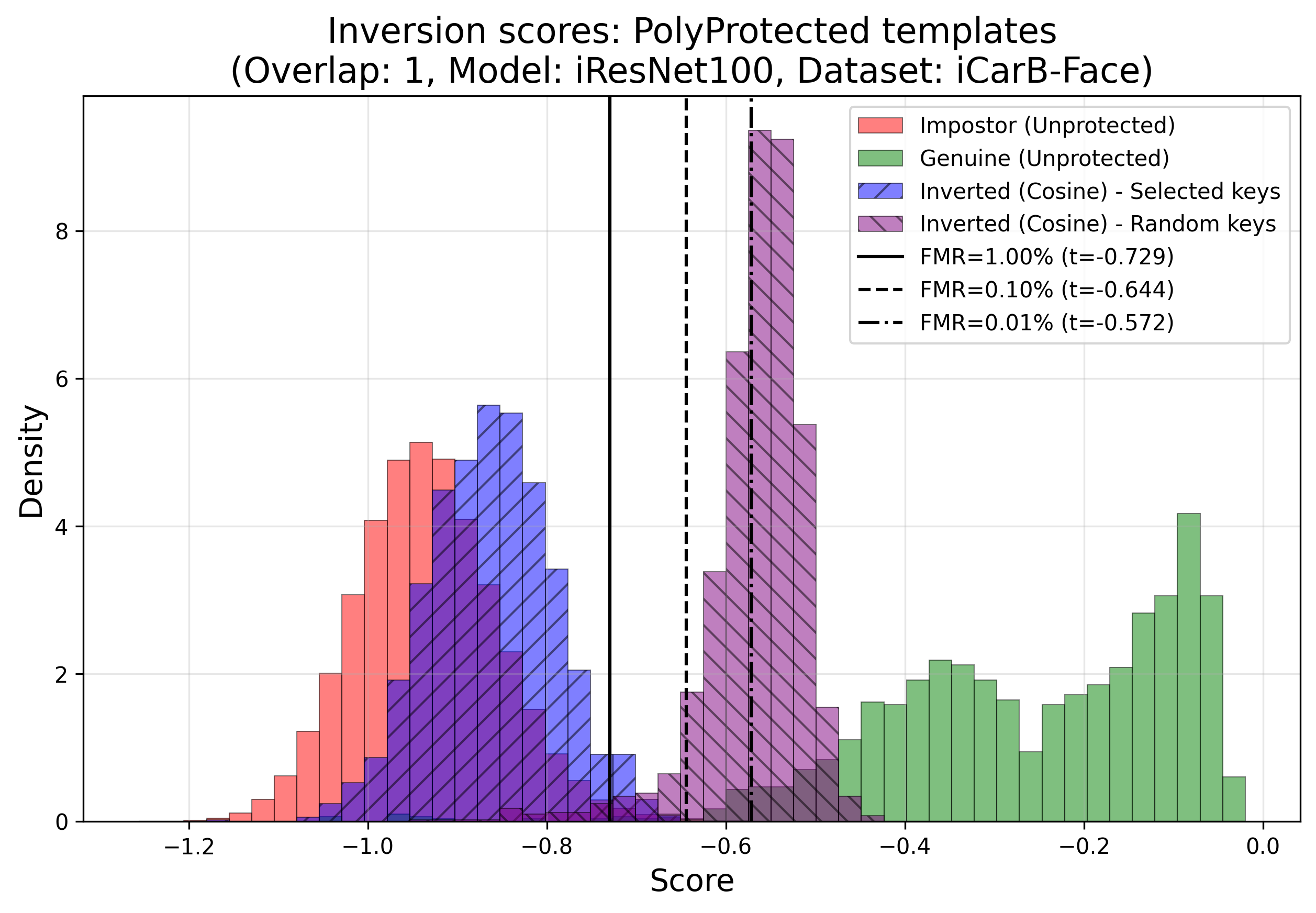}}
\vfil
\subfloat{\includegraphics[width=0.25\paperwidth]{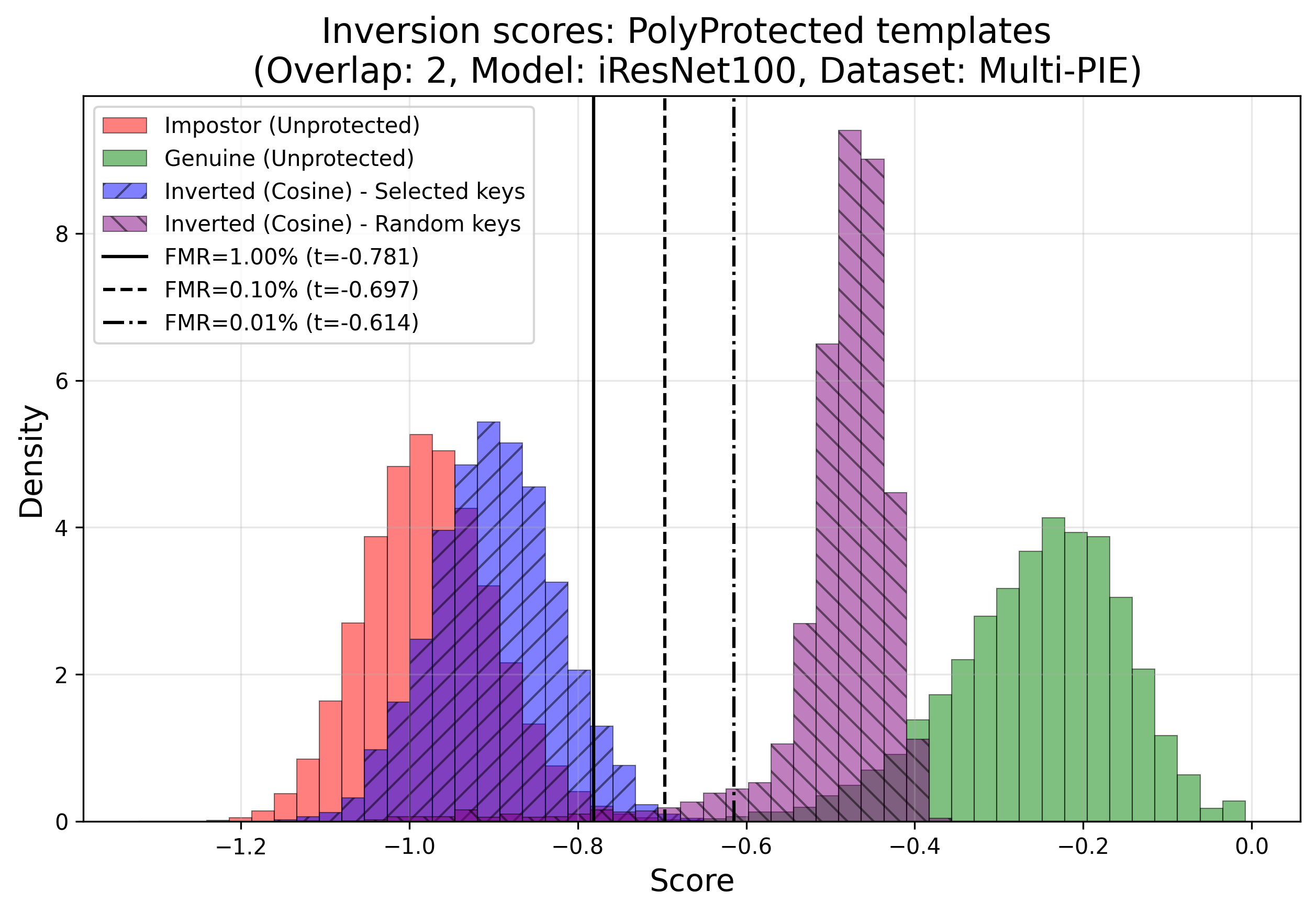}}
\hfil
\subfloat{\includegraphics[width=0.25\paperwidth]{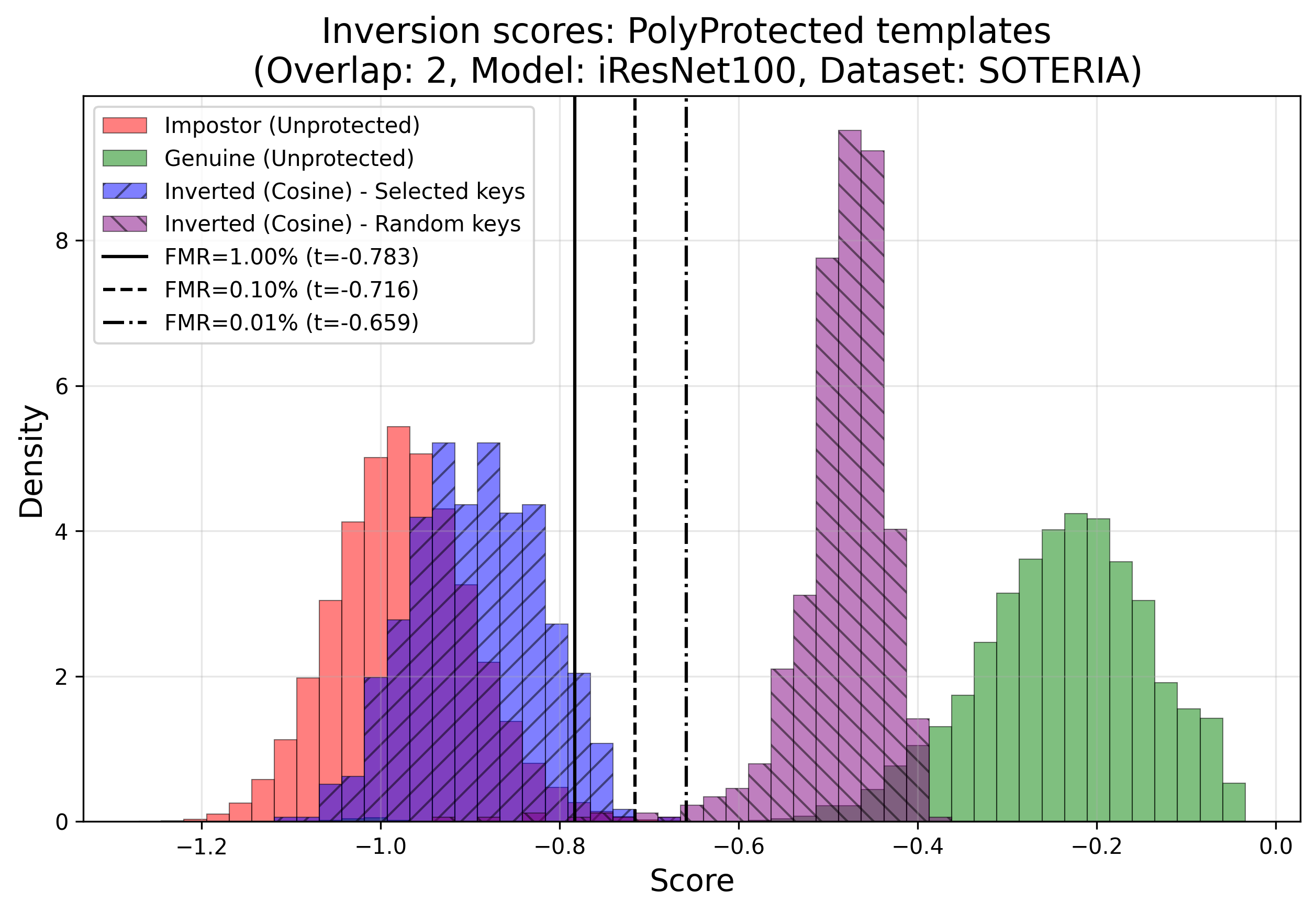}}
\hfil 
\subfloat{\includegraphics[width=0.25\paperwidth]{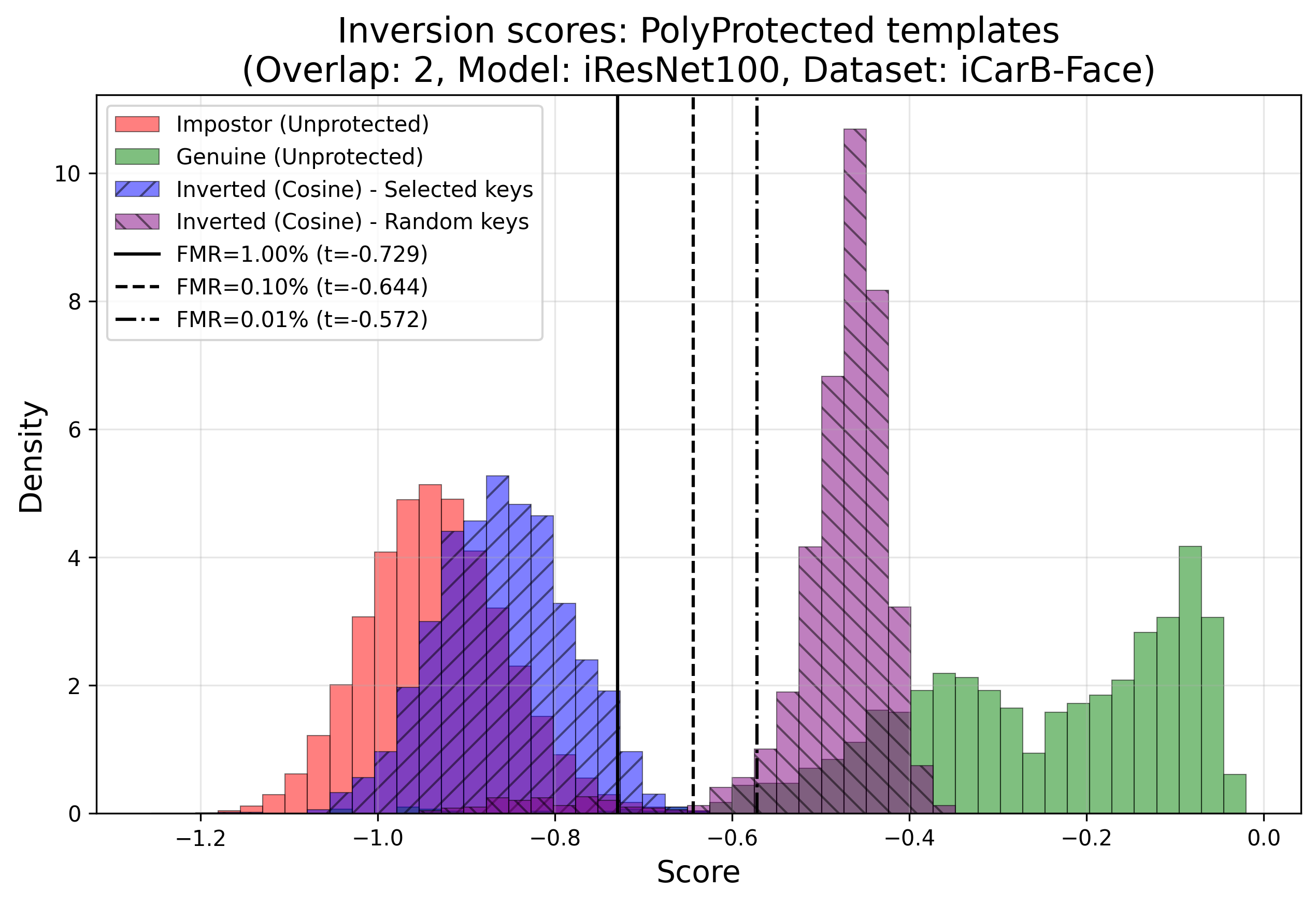}}
\vfil
\subfloat{\includegraphics[width=0.25\paperwidth]{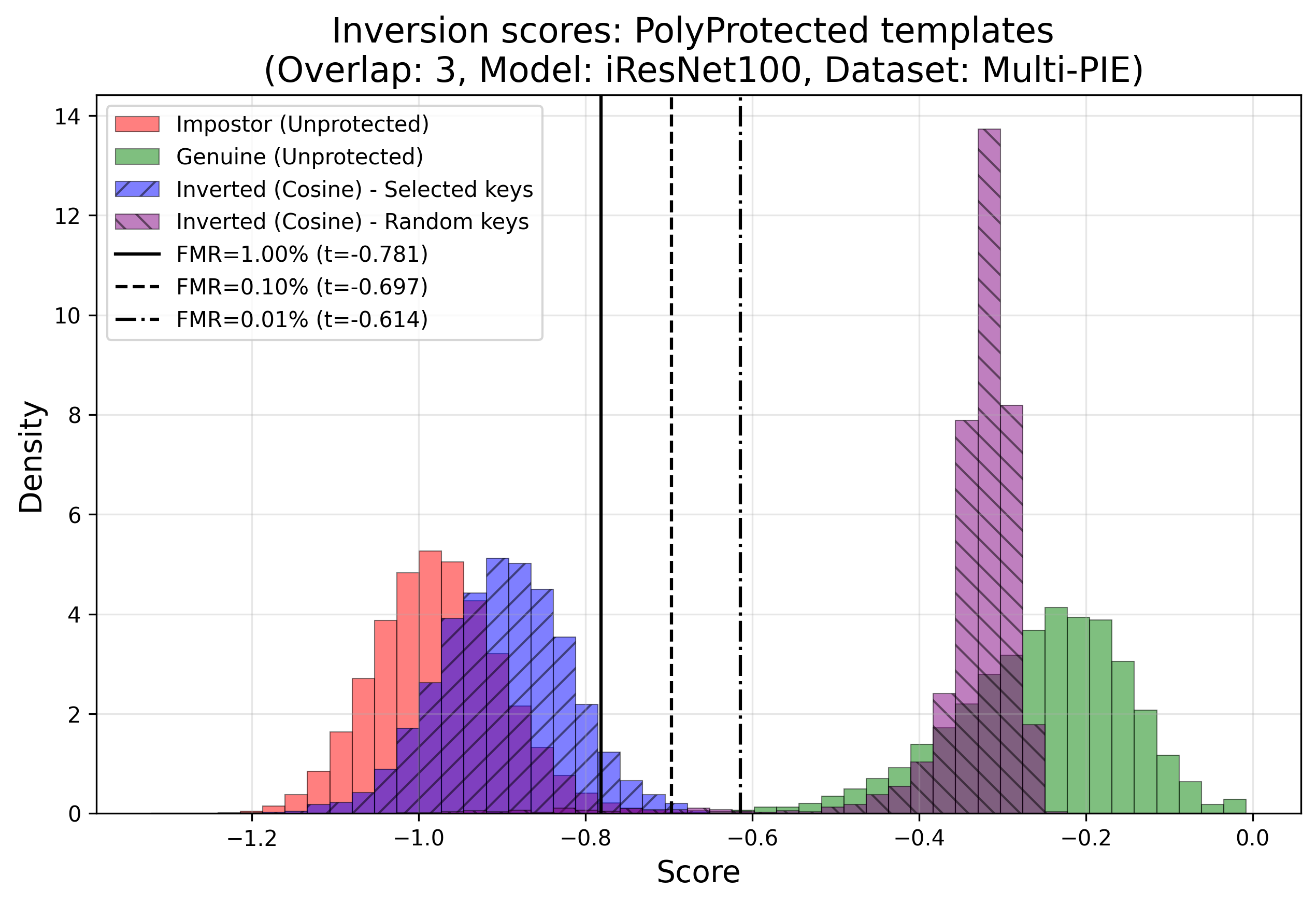}}
\hfil 
\subfloat{\includegraphics[width=0.25\paperwidth]{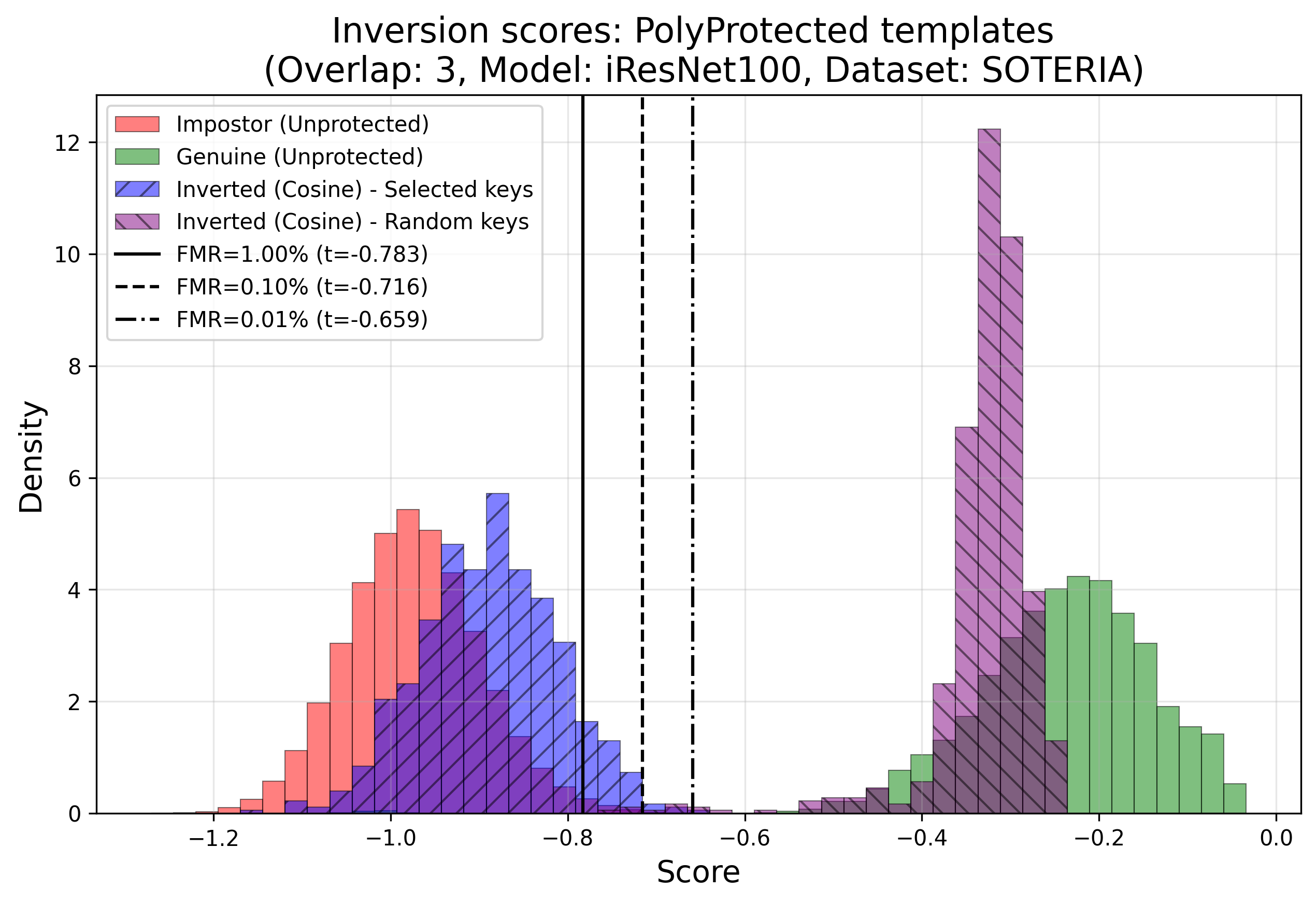}}
\hfil
\subfloat{\includegraphics[width=0.25\paperwidth]{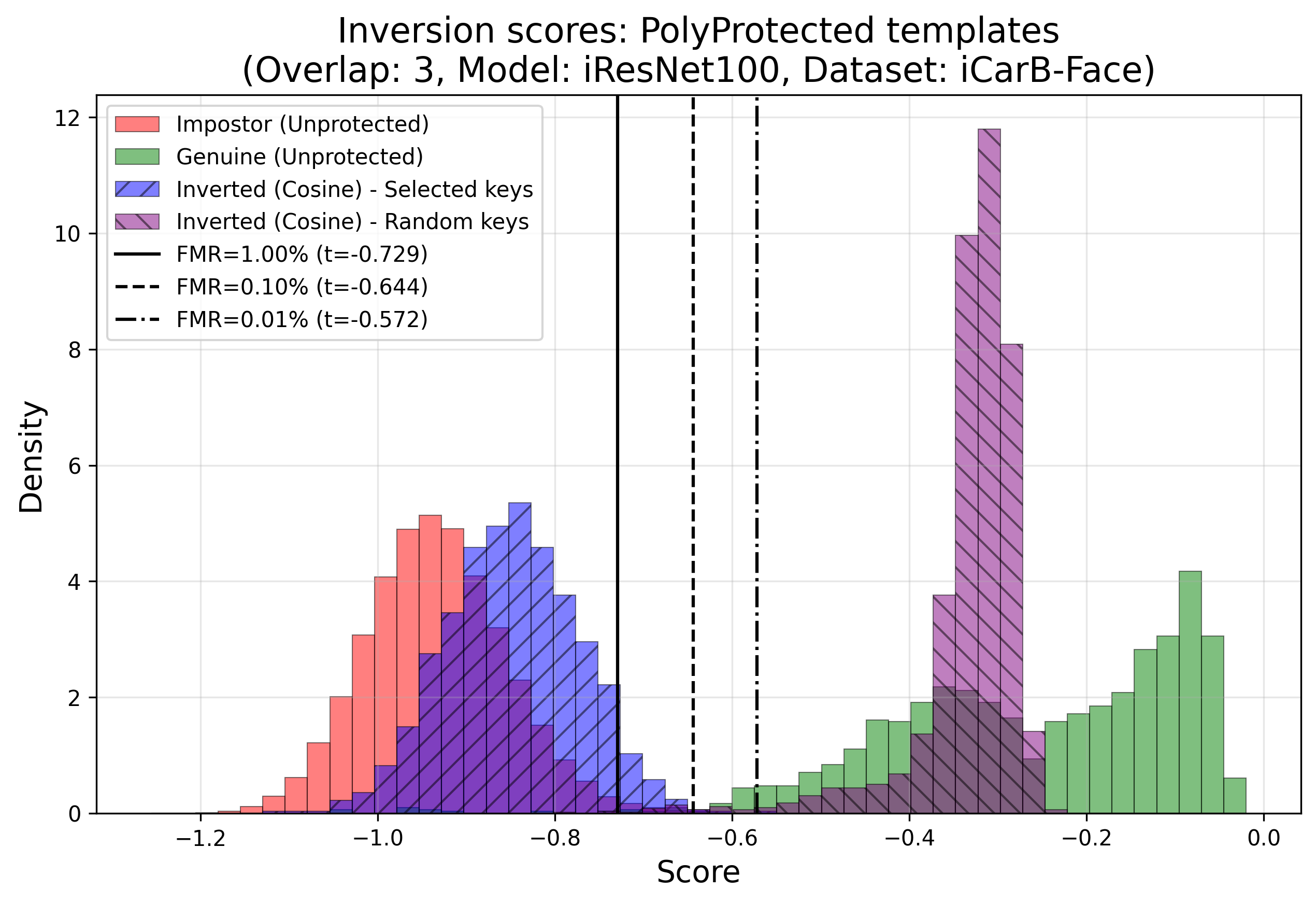}}
\caption{Inversion scores for PolyProtected templates generated from normalized iResNet100 face embeddings using different overlaps, when the keys ($C$ and $E$ parameters) are generated randomly versus using our key selection algorithm.  The inversion was performed using the cosine-based numerical solver.  The three vertical black lines represent different thresholds at which the inversion success rate (ISR) could be computed (e.g., Table \ref{tab:random_vs_ks}).}
\label{fig:ks}
\end{figure*}

Fig. \ref{fig:ks} shows the histograms of inversion scores for iResNet100 protected templates, based on the cosine solver, when the keys ($C$ and $E$) are selected randomly, as in Section \ref{sec:irreversibility_eval}, versus when our key selection algorithm is used.  Note that, in these plots, the genuine (green) and impostor (red) histograms are the same as in Fig. \ref{fig:solvers}, as is the histogram of inversion scores resulting from using random keys (purple).  The only new histogram is the one corresponding to inversion scores obtained when our key selection algorithm is used to generate the PolyProtected templates (blue).  To produce this histogram, 10 different (random) initial guesses for the solver were used to launch the inversion attack on the PolyProtected templates, and the resulting 10 sets of inversion scores were concatenated.  This was done because we cannot guarantee that the initial guesses used during the inversion step in the key selection algorithm (to select the $C$ and $E$ parameters) would be the same initial guesses used by an attacker attempting to invert the final, PolyProtected templates generated using the selected keys.

In Fig. \ref{fig:ks}, we observe that, for all PolyProtect overlaps and all face datasets, the blue histogram is located further to the left than the purple histogram.  In other words, the blue histogram boasts lower inversion scores, and it overlaps quite significantly with the unprotected template impostor score distribution.  This tells us that our key selection algorithm is very effective at improving the irreversibility of PolyProtect, i.e., when the $C$ and $E$ parameters are chosen more carefully, using our key selection algorithm, the resulting PolyProtected templates are more difficult to invert compared to when these keys are simply generated randomly.  Moreover, while random key generation results in varying degrees of irreversibility depending on the amount of overlap used in the PolyProtect transform (i.e., the purple histogram shifts to the right as the overlap increases), when our key selection algorithm is employed the irreversibility is approximately the same regardless of the overlap.  This suggests that our key selection algorithm may also be able to mitigate the irreversibility versus accuracy trade-off across different overlaps (discussed later), which was mentioned in Section \ref{sec:irreversibility_eval}.  Note that results for overlap = 4 are not shown, because the key selection algorithm did not manage to find suitable keys for all templates from the Multi-PIE and iCarB-Face datasets.  This suggests that an overlap of 4 should be avoided, as already recommended in \cite{kh22}, since the resulting PolyProtected templates seem fairly easy to invert using a numerical solver (see Fig. \ref{fig:solvers}) and our key selection algorithm cannot be guaranteed to fix that.

Table \ref{tab:random_vs_ks} compares the inversion success rate (ISR) of PolyProtected templates generated using random keys ($C$ and $E$) versus keys chosen by our key selection algorithm.  As in \cite{kh22}, the ISR was computed in terms of the proportion of PolyProtected templates whose inversion score (cosine distance) is below a pre-defined threshold, in which case the inversion attack would be considered successful.  Due to space constraints, we present results at only two of the three thresholds illustrated in Fig. \ref{fig:ks}: at 0.1\% FMR and at 0.01\% FMR, computed on the unprotected face recognition system (genuine and impostor scores).  However, from Fig. \ref{fig:ks} it is clear that the ISR would be higher at larger thresholds (e.g., $\geq$ 1\% FMR) and lower at stricter thresholds (e.g., $<$ 0.01\% FMR).  Note that Table \ref{tab:random_vs_ks} reports ISR for both iResNet100 (Fig. \ref{fig:ks}) and EdgeFace.    

\begin{table}[!h]
\renewcommand{\arraystretch}{1.2}
\caption{Inversion success rate (ISR) at two FMR thresholds, for normalized iResNet100 and EdgeFace embeddings protected via PolyProtect with different overlaps, when the keys ($C$, $E$) are generated randomly (\textit{R}) VS using our key selection algorithm (\textit{KS}).  $\downarrow$ shows the ISR improvement due to \textit{KS}.\label{tab:random_vs_ks}} 
\centering
\begin{tabular}{|c|c|c|c|c|c|c|c|}
\hline
\multirow{3}{*}{Dataset} & \multirow{3}{*}{Overlap} & \multicolumn{6}{c|}{ISR (\%)} \\
\cline{3-8}
 & & \multicolumn{3}{c|}{@ 0.1\% FMR} & \multicolumn{3}{c|}{@ 0.01\% FMR} \\
\cline{3-8}
 & & \textit{R} & \textit{KS} & $\downarrow$ & \textit{R} & \textit{KS} & $\downarrow$ \\
\hline
\multicolumn{8}{l}{\textbf{iResNet100:}} \\
\hline
\multirow{4}{*}{Multi-PIE} & 0 & 85.6 & 0.0 & 85.6 & 29.9 & 0.0 & 29.9 \\
 & 1 & 95.9 & 0.1 & 95.8 & 80.7 & 0.0 & 80.7 \\
 & 2 & 96.7 & 0.2 & 96.5 & 94.0 & 0.0 & 94.0 \\
 & 3 & 98.3 & 0.5 & 97.8 & 97.6 & 0.0 & 97.6 \\
\hline
\multirow{4}{*}{SOTERIA} & 0 & 96.3 & 0.1 & 96.2 & 74.6 & 0.0 & 74.6 \\
 & 1 & 95.9 & 0.6 & 95.3 & 92.3 & 0.0 & 92.3 \\
 & 2 & 98.9 & 0.1 & 98.8 & 98.4 & 0.0 & 98.4 \\
 & 3 & 99.6 & 0.9 & 98.7 & 98.7 & 0.0 & 98.7 \\
\hline
\multirow{4}{*}{iCarB-Face} & 0 & 68.5 & 0.2 & 68.3 & 14.9 & 0.0 & 14.9 \\
 & 1 & 93.5 & 0.0 & 93.5 & 63.3 & 0.0 & 63.3 \\
 & 2 & 95.6 & 0.1 & 95.5 & 92.6 & 0.0 & 92.6 \\
 & 3 & 99.2 & 0.4 & 98.8 & 98.6 & 0.1 & 98.5 \\
\hline
\multicolumn{8}{l}{\textbf{EdgeFace:}} \\
\hline
\multirow{4}{*}{Multi-PIE} & 0 & 91.2 & 0.5 & 90.7 & 41.3 & 0.0 & 41.3 \\
 & 1 & 97.0 & 0.8 & 96.2 & 87.5 & 0.0 & 87.5 \\
 & 2 & 98.8 & 1.7 & 97.1 & 97.3 & 0.0 & 97.3 \\
 & 3 & 97.9 & 2.7 & 95.2 & 97.5 & 0.1 & 97.4 \\
\hline
\multirow{4}{*}{SOTERIA} & 0 & 90.9 & 3.4 & 87.5 & 81.7 & 0.3 & 81.4 \\
 & 1 & 98.6 & 2.0 & 96.6 & 97.6 & 0.3 & 97.3 \\
 & 2 & 97.7 & 3.7 & 94.0 & 96.9 & 0.9 & 96.0 \\
 & 3 & 100.0 & 4.4 & 95.6 & 99.6 & 0.9 & 98.7 \\
\hline
\multirow{4}{*}{iCarB-Face} & 0 & 78.3 & 0.2 & 78.1 & 10.5 & 0.0 & 10.5 \\
 & 1 & 95.1 & 0.3 & 94.8 & 59.3 & 0.0 & 59.3 \\
 & 2 & 98.4 & 0.2 & 98.2 & 96.4 & 0.0 & 96.4 \\
 & 3 & 99.2 & 0.8 & 98.4 & 97.7 & 0.1 & 97.6 \\
\hline
\end{tabular}
\end{table} 

From Table \ref{tab:random_vs_ks}, it is evident that our key selection algorithm drastically reduces the ISR for both iResNet100 and EdgeFace, across all overlaps, at both thresholds, and for all three datasets.  For example, at the less strict threshold (0.1\% FMR), this represents an improvement of 85.6 -- 97.8\% for iResNet100 and 90.7 -- 97.1\% for EdgeFace on Multi-PIE, 95.3 -- 98.8\% for iResNet100 and 87.5 -- 96.6\% for EdgeFace on SOTERIA, and 68.3 -- 98.8\% for iResNet100 and 78.1 -- 98.4\% for EdgeFace on iCarB-Face, depending on the amount of overlap used in the PolyProtect transform.  At the stricter threshold (0.01\% FMR), the improvement is generally a bit smaller, since the initial ISR (when random keys are used) is lower (as it is harder to find a match between the inverted template and the original embedding at this threshold); however, the improvement in ISR thanks to our key selection algorithm is still significant, especially considering that the ISR is now reduced to $<$ 1\% in all evaluation scenarios and is approximately 0 in most cases (i.e., the PolyProtected templates would be considered practically irreversible at this threshold, regardless of the overlap).  While the ISR seems to be, in general, higher for protected templates originating from EdgeFace embeddings, the effect of the key selection algorithm is substantial regardless of the face recognition model used to generate the underlying face embeddings.  Overall, these findings indicate that our key selection algorithm should definitely be used to select the $C$ and $E$ parameters of the PolyProtect polynomials, in order to ensure a high degree of irreversibility (low ISR) for the worst-case scenario of an inversion attack based on the cosine numerical solver.  Selecting these parameters in a purely random fashion is not recommended.

\subsection{Effect of key selection on accuracy}
\label{subsec:ks_accuracy}

\begin{figure}[!h]
\centering
\subfloat{\includegraphics[width=0.33\columnwidth]{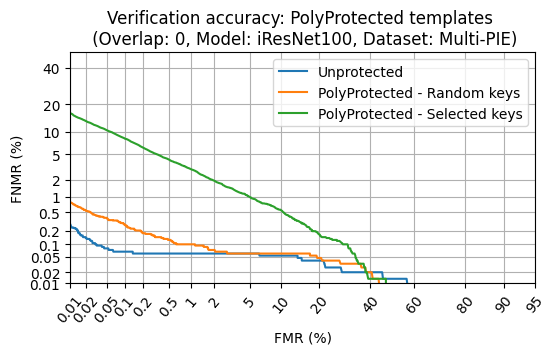}}
\hfil 
\subfloat{\includegraphics[width=0.33\columnwidth]{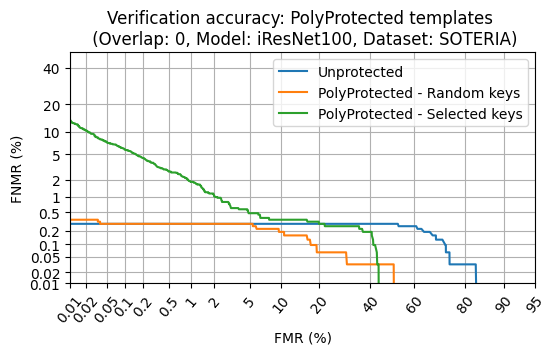}}
\hfil 
\subfloat{\includegraphics[width=0.33\columnwidth]{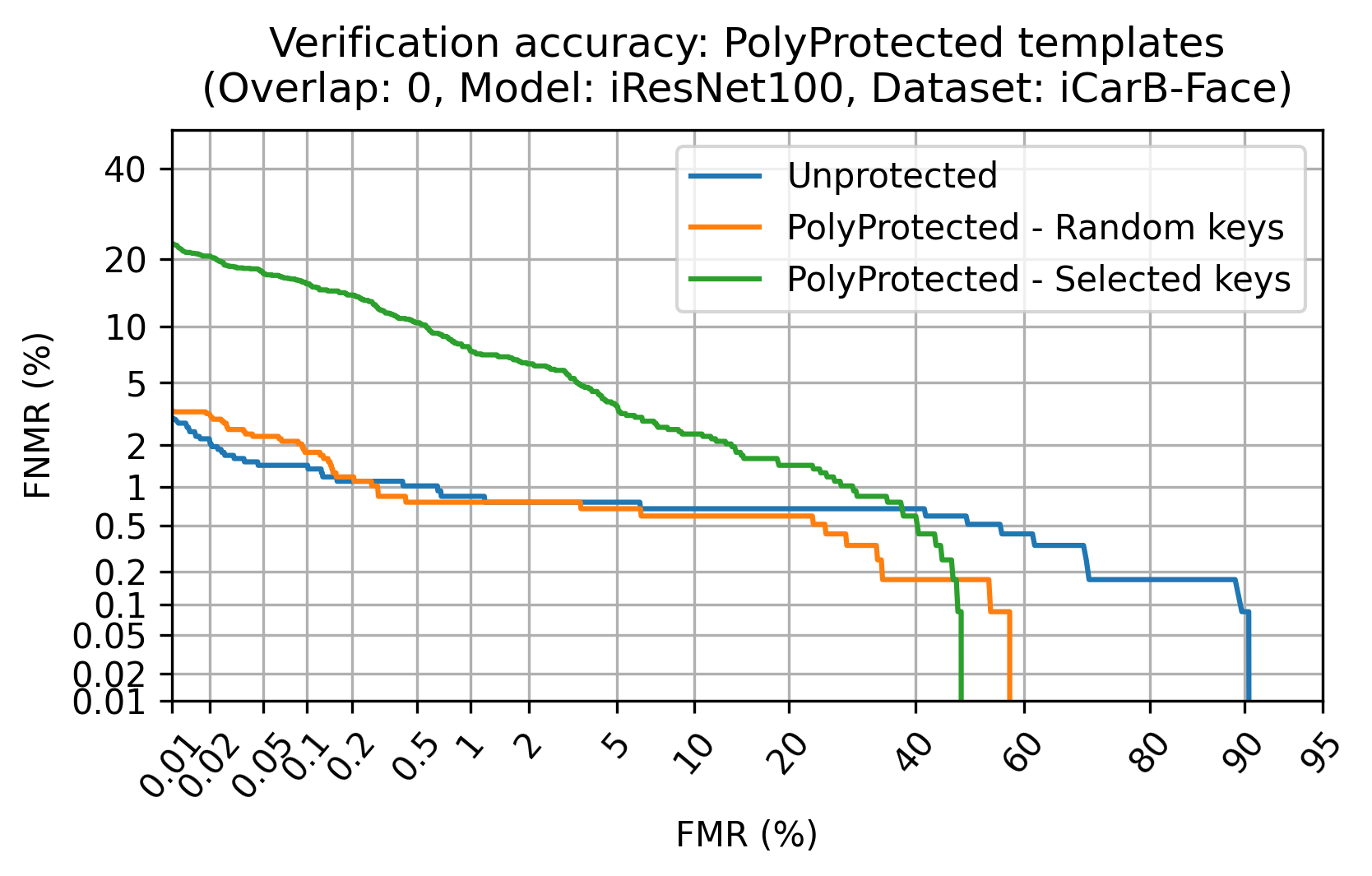}}
\vfil
\subfloat{\includegraphics[width=0.33\columnwidth]{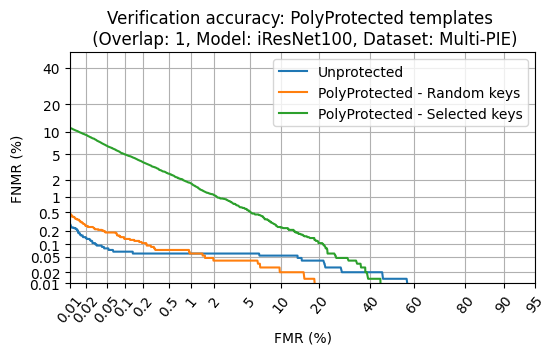}}
\hfil
\subfloat{\includegraphics[width=0.33\columnwidth]{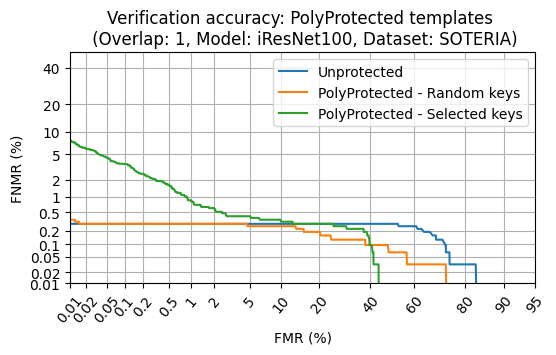}}
\hfil 
\subfloat{\includegraphics[width=0.33\columnwidth]{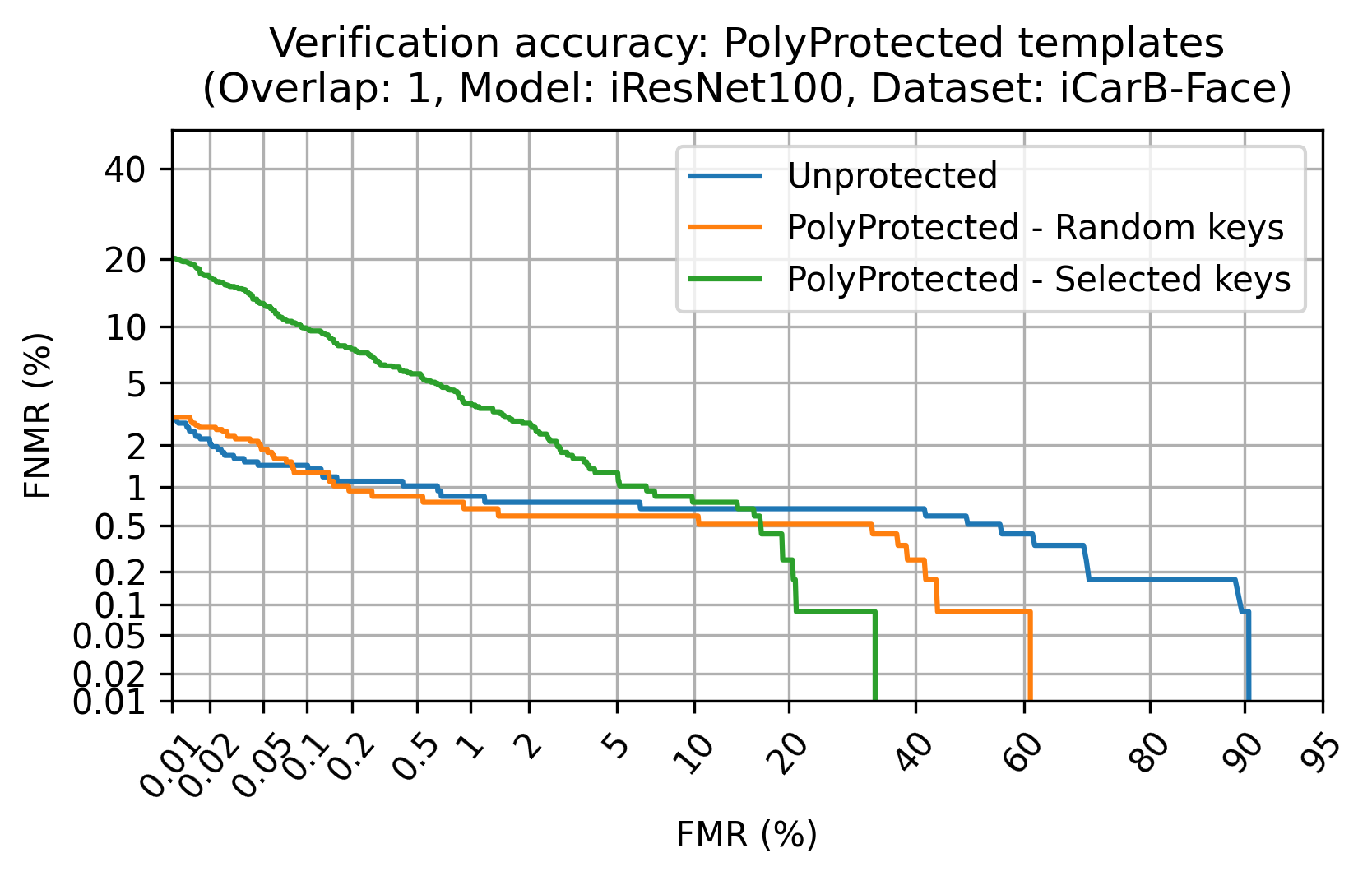}}
\vfil
\subfloat{\includegraphics[width=0.33\columnwidth]{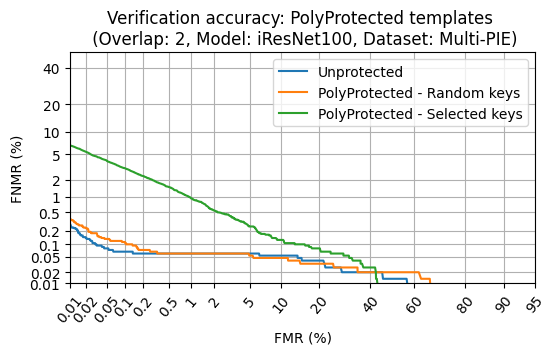}}
\hfil 
\subfloat{\includegraphics[width=0.33\columnwidth]{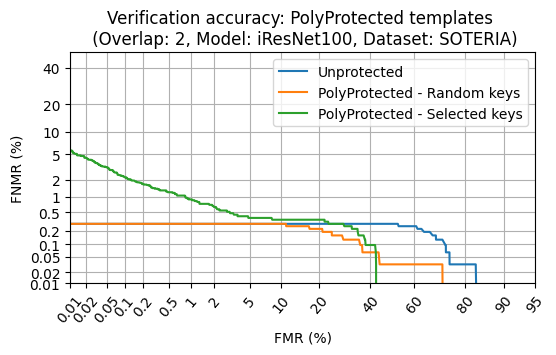}}
\hfil 
\subfloat{\includegraphics[width=0.33\columnwidth]{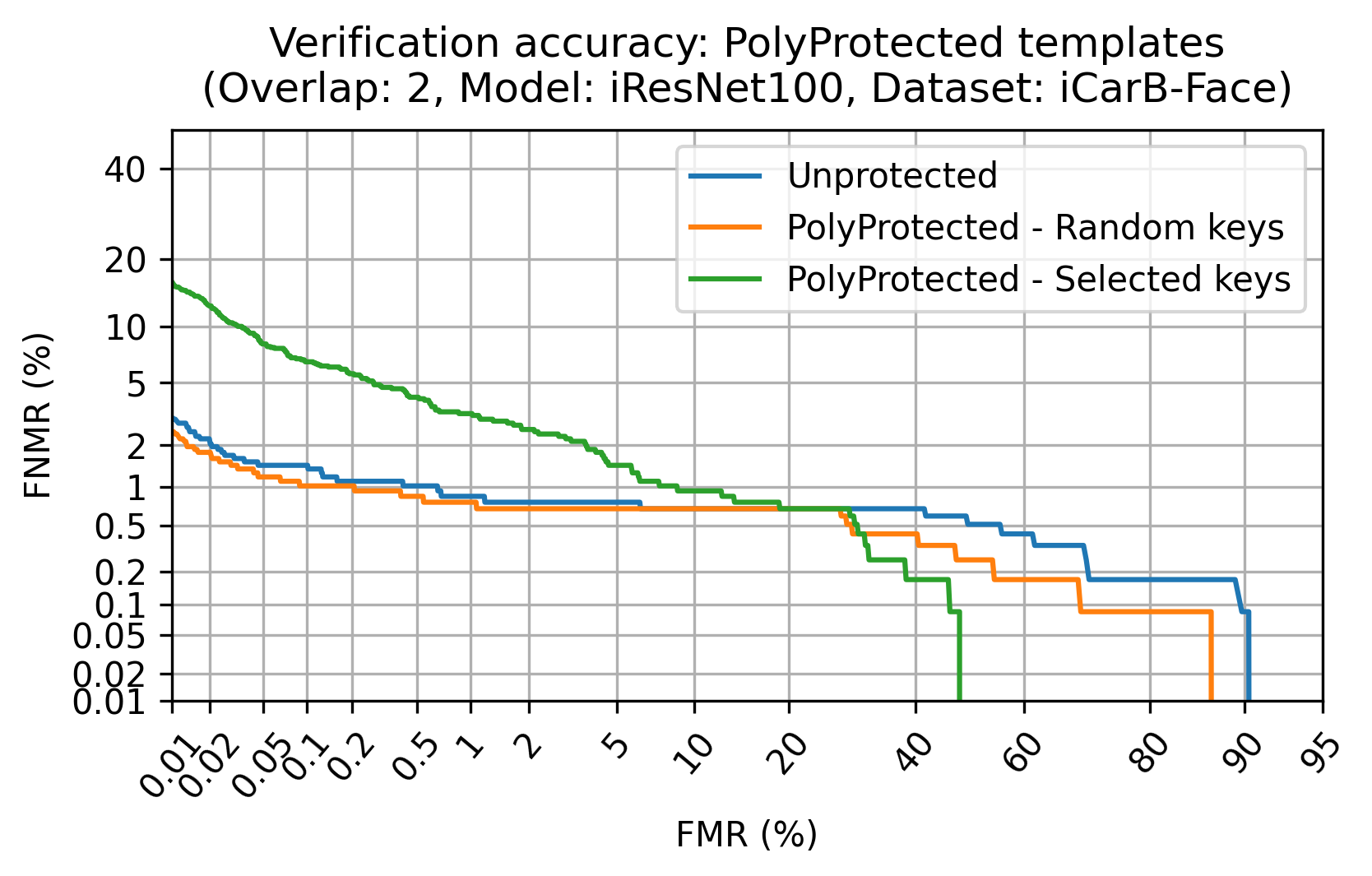}}
\vfil
\subfloat{\includegraphics[width=0.33\columnwidth]{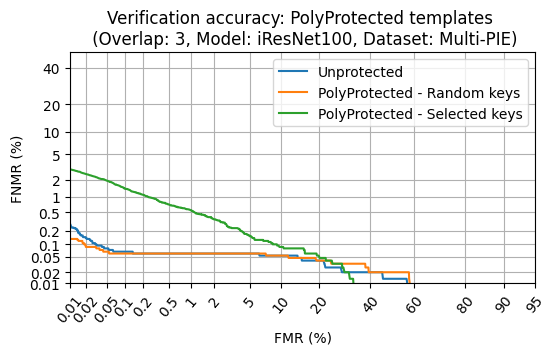}}
\hfil 
\subfloat{\includegraphics[width=0.33\columnwidth]{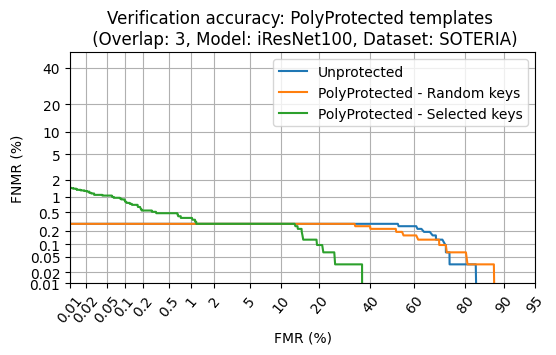}}
\hfil
\subfloat{\includegraphics[width=0.33\columnwidth]{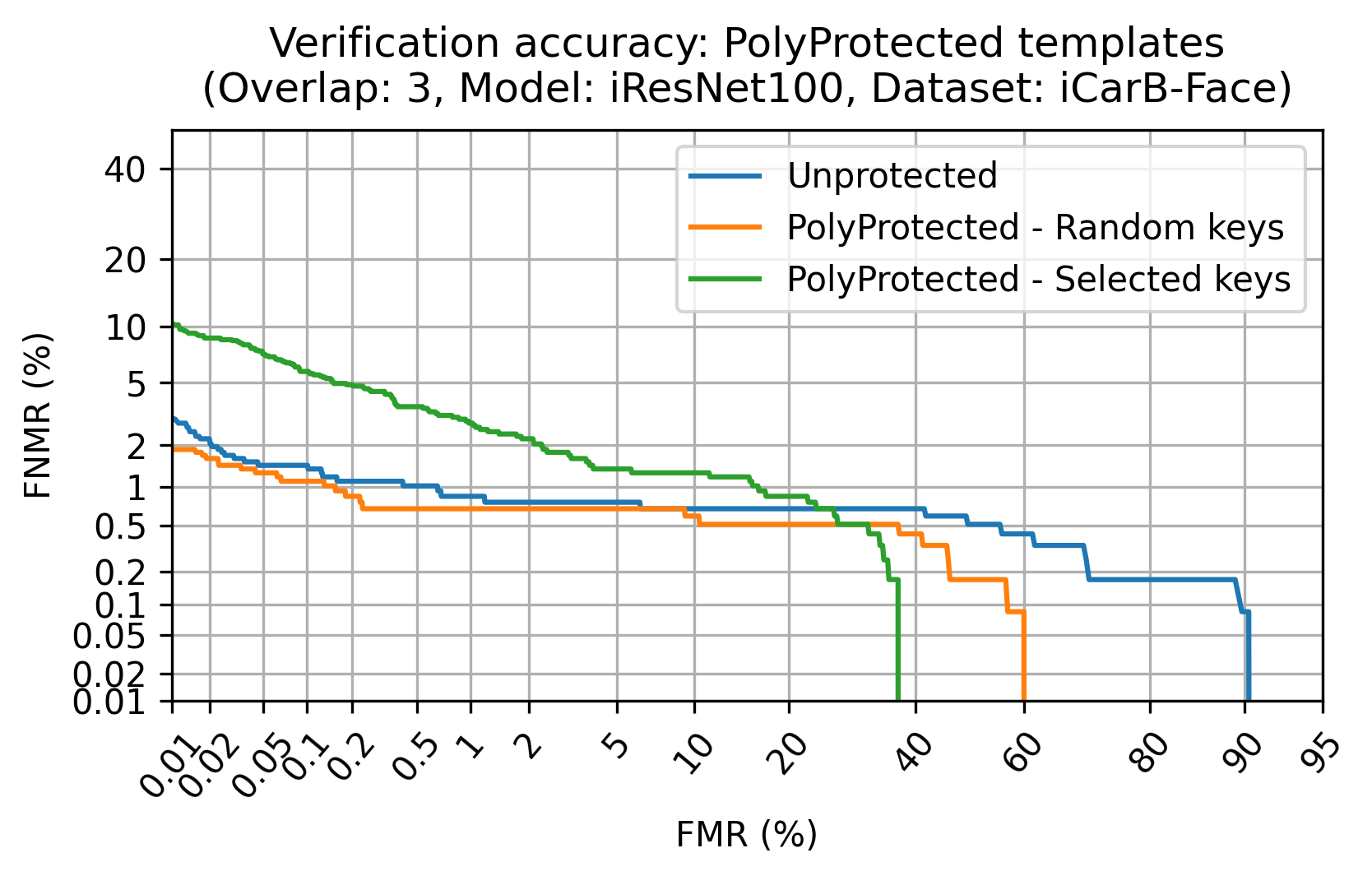}}
\vfil
\caption{Detection Error Trade-off (DET) plots comparing verification accuracy for PolyProtected templates generated from normalized iResNet100 face embeddings using different overlaps, when the keys ($C$ and $E$ parameters) are chosen randomly versus using our key selection algorithm.}
\label{fig:ks_accuracy}
\end{figure}

Our key selection algorithm is clearly effective at improving the irreversibility of PolyProtect, which was the aim of the work presented in this paper.  However, we know from earlier work \cite{kh22, s25} that PolyProtect incurs a trade-off between the irreversibility and accuracy of protected templates depending on the amount of overlap used in the transform.  Indeed, we also observed this in the accuracy analysis in Section \ref{sec:accuracy_eval} and the irreversibility analysis (based on random keys) in Section \ref{sec:irreversibility_eval}, i.e., as the overlap increases, the accuracy increases but the irreversibility decreases, and vice-versa.  So now the following question arises: Since our key selection algorithm is able to drastically improve, and approximately equalise, the irreversibility across all overlaps, what effect does this have on the recognition accuracy of the resulting PolyProtected templates?  Fig. \ref{fig:ks_accuracy} compares the verification accuracy obtained in the protected domain for iResNet100 embeddings, when PolyProtect's $C$ and $E$ parameters are generated using our key selection algorithm versus when they are generated randomly.

From Fig. \ref{fig:ks_accuracy}, we may conclude that our key selection algorithm degrades, to some extent, the recognition accuracy in the PolyProtected domain, compared to when the PolyProtected templates are generated using purely random $C$ and $E$ parameters.  Table \ref{tab:ks_acc} quantifies this accuracy degradation for both iResNet100 and EdgeFace, at the same two thresholds used to compute the ISR in Table \ref{tab:random_vs_ks}.

\begin{table}[!h]
\renewcommand{\arraystretch}{1.2}
\caption{Verification accuracy (FNMR) at two FMR thresholds, for normalized iResNet100 and EdgeFace embeddings protected via PolyProtect with different overlaps, when the keys ($C$, $E$) are generated randomly (\textit{R}) VS using our key selection algorithm (\textit{KS}).  $\downarrow$ shows the accuracy degradation due to \textit{KS}.\label{tab:ks_acc}} 
\centering
\begin{tabular}{|c|c|c|c|c|c|c|c|}
\hline
\multirow{3}{*}{Dataset} & \multirow{3}{*}{Overlap} & \multicolumn{6}{c|}{FNMR (\%)} \\
\cline{3-8}
 & & \multicolumn{3}{c|}{@ 0.1\% FMR} & \multicolumn{3}{c|}{@ 0.01\% FMR} \\
\cline{3-8}
 & & \textit{R} & \textit{KS} & $\downarrow$ & \textit{R} & \textit{KS} & $\downarrow$ \\
\hline
\multicolumn{8}{l}{\textbf{iResNet100:}} \\
\hline
\multirow{4}{*}{Multi-PIE} & 0 & 0.3 & 8.3 & 8.0 & 0.8 & 16.3 & 15.5 \\
 & 1 & 0.1 & 5.0 & 4.9 & 0.5 & 11.2 & 10.7 \\
 & 2 & 0.1 & 3.1 & 3.0 & 0.4 & 6.7 & 6.3 \\
 & 3 & 0.1 & 1.4 & 1.3 & 0.1 & 3.0 & 2.9 \\
\hline
\multirow{4}{*}{SOTERIA} & 0 & 0.3 & 5.9 & 5.6 & 0.4 & 13.8 & 13.4 \\
 & 1 & 0.3 & 3.6 & 3.3 & 0.4 & 7.9 & 7.5 \\
 & 2 & 0.3 & 2.2 & 1.9 & 0.3 & 5.7 & 5.4 \\
 & 3 & 0.3 & 0.9 & 0.6 & 0.3 & 1.5 & 1.2 \\
\hline
\multirow{4}{*}{iCarB-Face} & 0 & 1.8 & 15.8 & 14.0 & 3.4 & 23.0 & 19.6 \\
 & 1 & 1.3 & 9.8 & 8.5 & 3.1 & 20.4 & 17.3 \\
 & 2 & 1.0 & 6.5 & 5.5 & 2.5 & 16.1 & 13.6 \\
 & 3 & 1.1 & 5.8 & 4.7 & 1.9 & 10.4 & 8.5 \\
\hline
\multicolumn{8}{l}{\textbf{EdgeFace:}} \\
\hline
\multirow{4}{*}{Multi-PIE} & 0 & 0.4 & 9.3 & 8.9 & 0.8 & 18.5 & 17.7 \\
 & 1 & 0.2 & 6.7 & 6.5 & 0.5 & 12.8 & 12.3 \\
 & 2 & 0.2 & 4.0 & 3.8 & 0.4 & 8.2 & 7.8 \\
 & 3 & 0.1 & 1.5 & 1.4 & 0.3 & 3.2 & 2.9 \\
\hline
\multirow{4}{*}{SOTERIA} & 0 & 0.4 & 8.4 & 8.0 & 0.8 & 16.3 & 15.5 \\
 & 1 & 0.3 & 4.2 & 3.9 & 0.4 & 8.3 & 7.9 \\
 & 2 & 0.3 & 2.5 & 2.2 & 0.4 & 4.8 & 4.4 \\
 & 3 & 0.3 & 1.5 & 1.2 & 0.3 & 2.8 & 2.5 \\
\hline
\multirow{4}{*}{iCarB-Face} & 0 & 3.3 & 20.3 & 17.0 & 7.5 & 30.7 & 23.2 \\
 & 1 & 2.1 & 13.1 & 11.0 & 3.2 & 24.5 & 21.3 \\
 & 2 & 1.3 & 12.2 & 10.9 & 2.5 & 18.8 & 16.3 \\
 & 3 & 1.1 & 8.5 & 7.4 & 2.0 & 13.7 & 11.7 \\
\hline
\end{tabular}
\end{table} 

If we compare Table \ref{tab:ks_acc} to Table \ref{tab:random_vs_ks}, we see that the amount by which the accuracy decreases as a result of using our key selection algorithm is much less significant (in absolute terms) than the amount by which the ISR improves; e.g., at the less strict threshold (0.1\% FMR), the accuracy degradation is 1.3 -- 8.0\% for iResNet100 and 1.4 -- 8.9\% for EdgeFace on Multi-PIE, 0.6 -- 5.6\% for iResNet100 and 1.2 -- 8.0\% for EdgeFace on SOTERIA, and 4.7 -- 14.0\% for iResNet100 and 7.4 -- 17.0\% for EdgeFace on iCarB-Face, depending on the amount of overlap used in the PolyProtect transform, whereas the corresponding improvements in irreversibility are 85.6 -- 97.8\% for iResNet100 and 90.7 -- 97.1\% for EdgeFace on Multi-PIE, 95.3 -- 98.8\% for iResNet100 and 87.5 -- 96.6\% for EdgeFace on SOTERIA, and 68.3 -- 98.8\% for iResNet100 and 78.1 -- 98.4\% for EdgeFace on iCarB-Face.  At the stricter threshold (0.01\% FMR), the accuracy drop is larger, but the improvement in ISR is still much more drastic in comparison.  So, even though our key selection algorithm seems to have a negative effect on the accuracy attainable by the resulting PolyProtected templates, the increase in the irreversibility of those templates is significantly higher than this drop in accuracy.  Admittedly, a direct comparison between the decrease in accuracy and increase in irreversibility may not make sense in every application, since an \textit{x}\% accuracy drop may have different practical consequences compared to an \textit{x}\% improvement in irreversibility.  Furthermore, without targeting a specific application, it is impossible to conclude whether a certain accuracy degradation is ``worth'' the corresponding (and much larger) ISR improvement provided by our key selection algorithm -- this is a trade-off that must be considered within the context of each application.  Nevertheless, in global terms, we would still recommend using the key selection algorithm over simply generating the $C$ and $E$ parameters randomly, since the latter scenario is likely to result in poor irreversibility (and thus poor template protection) against the worst-case inversion attack based on a cosine-distance numerical solver.  

Another observation from comparing Tables \ref{tab:random_vs_ks} and \ref{tab:ks_acc} is that our key selection algorithm is able to effectively mitigate, and in some cases (e.g., at the 0.01\% FMR threshold) practically eliminate, the accuracy versus irreversibility trade-off that exists \textit{across different overlaps} used in the PolyProtect transformation.  In particular, Table \ref{tab:random_vs_ks} shows that the key selection algorithm allows us to concentrate the irreversibility across different overlaps around a small ISR, and Table \ref{tab:ks_acc} shows that the largest overlap results in the best accuracy.  This suggests that, to ensure the best accuracy versus irreversibility trade-off, we should select the largest (sensible) overlap; however, the maximum should be avoided due to our earlier observations (in relation to Fig. \ref{fig:ks}) that it may be difficult to select keys that produce irreversible protected templates when an overlap of 4 is used in the PolyProtect transformation.  So, in our experimental set-up, this recommendation would translate to an overlap of 3, which resulted in the best accuracy (Table \ref{tab:ks_acc}) while still maintaining a high degree of irreversibility comparable to the lower overlaps (Table \ref{tab:random_vs_ks}).  This may be interpreted as a general recommendation for choosing the most appropriate overlap when employing our key selection algorithm; however, in practice the accuracy versus irreversibility trade-off should be evaluated separately for each application context, and the overlap should be carefully tuned based on the requirements of the target PolyProtected system.

To drive home this point about the accuracy versus irreversibility trade-off being effectively eliminated \textit{across different overlaps} (when using our key selection algorithm), consider the example of EdgeFace embeddings from the iCarB-Face dataset PolyProtected using overlap = 0.  This configuration resulted in the worst accuracy degradation when switching from \textit{random} to \textit{selected} PolyProtect keys: Table \ref{tab:ks_acc} shows an increase in the FNMR (at 0.01\% FMR) from 7.5\% to 30.7\%.  These figures are accompanied by ISR of 10.5\% and 0.0\%, respectively (Table \ref{tab:random_vs_ks}).  If we now choose overlap = 3 instead of overlap = 0, the accuracy in the selected key scenario improves considerably from FNMR of 30.7\% to 13.7\%, while the ISR remains close to 0 (0.1\%).  In other words, using a higher overlap significantly improves the accuracy, while having almost no effect on the irreversibility, meaning that the accuracy/irreversibility trade-off is effectively eliminated.  Compare this to the \textit{random} key scenario, where the smaller improvement in accuracy from 7.5\% to 2.0\% FNMR as we go from overlap = 0 to overlap = 3 is accompanied by the colossal degradation in ISR from 10.5\% to 97.7\%.  In other words, when we use random PolyProtect keys, there is a clear accuracy/irreversibility trade-off, where increasing the overlap improves the accuracy at the cost of degrading the irreversibility by a potentially significant amount.  Our key selection algorithm mitigates this trade-off by keeping the irreversibility approximately \textit{stable} (and concentrated around a small ISR) as the overlap increases.

\subsection{Effect of key selection on unlinkability}
\label{subsec:ks_unlink}

In this section, we explore the effect of our key selection algorithm on the unlinkability of PolyProtect, i.e., the ability to generate multiple protected templates from the same face identity (but different embedding instances) using different keys (coefficients and exponents), such that these templates do \textit{not} match.  To conduct this experiment, we began by generating two sets of 10 subject-specific keys for each identity in each of our three datasets, and for each of our two face recognition models.  One set of 10 keys was generated \textit{randomly}, and the other 10 keys were \textit{selected} using our key selection algorithm.  We then used the keys to produce 10 random-key PolyProtected templates and 10 selected-key PolyProtected templates for each identity.  The resulting protected templates were compared to each other in terms of cosine distance to generate \textit{mated} and \textit{non-mated} scores (corresponding to same-subject and different-subject template comparisons, respectively).  This was done separately for the 10 random and 10 selected keys.  Finally, we plugged these scores into the evaluation framework proposed in \cite{gb18}, to estimate the global (across the entire score range) unlinkability, $D_{\leftrightarrow}^{\mathit{sys}}$, of the iResNet100 and EdgeFace PolyProtected templates.  Note that $D_{\leftrightarrow}^{\mathit{sys}}$ lies in the range $[0, 1]$, where 0 corresponds to full \textit{unlinkability} (good) and 1 to full \textit{linkability} (bad).  Table \ref{tab:unlink} summarises our results.

\begin{table}[!h]
\renewcommand{\arraystretch}{1.2}
\caption{Unlinkability ($D_{\leftrightarrow}^{\mathit{sys}}$) for normalized iResNet100 and EdgeFace embeddings protected via PolyProtect with different overlaps, when the keys ($C$, $E$) are generated randomly (\textit{R}) VS using our key selection algorithm (\textit{KS}).} \label{tab:unlink} 
\centering
\begin{tabular}{|c|c|c|c|c|c|c|c|}
\hline
\multirow{3}{*}{Model} & \multirow{3}{*}{Overlap} & \multicolumn{6}{c|}{$D_{\leftrightarrow}^{\mathit{sys}}$} \\
\cline{3-8}
 & & \multicolumn{2}{c|}{Multi-PIE} & \multicolumn{2}{c|}{SOTERIA} & \multicolumn{2}{c|}{iCarB-Face} \\
\cline{3-8}
 & & \textit{R} & \textit{KS} & \textit{R} & \textit{KS} & \textit{R} & \textit{KS} \\
\hline
\multirow{4}{*}{iResNet100} & 0 & 0.21 & 0.20 & 0.21 & 0.23 & 0.19 & 0.23 \\
 \cline{2-8}
 & 1 & 0.21 & 0.20 & 0.21 & 0.21 & 0.19 & 0.25 \\
 \cline{2-8}
 & 2 & 0.21 & 0.21 & 0.21 & 0.22 & 0.19 & 0.23 \\
 \cline{2-8}
 & 3 & 0.21 & 0.23 & 0.21 & 0.23 & 0.19 & 0.26 \\
\hline
\multirow{4}{*}{EdgeFace} & 0 & 0.21 & 0.19 & 0.20 & 0.20 & 0.19 & 0.24 \\
 \cline{2-8}
 & 1 & 0.21 & 0.20 & 0.21 & 0.21 & 0.19 & 0.24 \\
 \cline{2-8}
 & 2 & 0.21 & 0.21 & 0.21 & 0.23 & 0.19 & 0.24 \\
 \cline{2-8}
 & 3 & 0.19 & 0.23 & 0.19 & 0.24 & 0.17 & 0.26 \\
\hline
\end{tabular}
\end{table} 

From Table \ref{tab:unlink}, we see that, on Multi-PIE and SOTERIA, the unlinkability of the protected templates is usually approximately maintained when the key selection algorithm is used compared to when the keys are generated randomly.  In some cases, we note that $D_{\leftrightarrow}^{\mathit{sys}}$ is slightly higher for selected keys, such as when using an overlap of 3 on EdgeFace embeddings, and most notably across all overlaps on iCarB-Face; however, considering that $D_{\leftrightarrow}^{\mathit{sys}}$ lies in the range $[0, 1]$, these increases in $D_{\leftrightarrow}^{\mathit{sys}}$ when going from random to selected keys seem fairly inconsequential (especially considering the significantly greater improvement in ISR that key selection affords us -- see Table \ref{tab:random_vs_ks}).  So, we may conclude that, in general, our key selection algorithm is unlikely to have a considerable impact on the unlinkability of PolyProtect.  

\subsection{Effects of varying key selection threshold}
\label{subsec:ks_thresh}

In all preceding key selection experiments, the threshold used to define a successful inversion was set at 20\% FMR, as noted at the beginning of Section \ref{sec:key_selection}.  In this section, we explore the effects of varying that threshold.  In particular, we consider how the choice of threshold in the key selection algorithm affects the irreversibility, accuracy, and unlinkability of the resulting PolyProtected templates.

Table \ref{tab:ks_thresh} shows the change in ISR (irreversibility), FNMR (accuracy), and $D_{\leftrightarrow}^{\mathit{sys}}$ (unlinkability), as the \textit{key selection threshold}, which is used to define a ``good'' key (leading to an \textit{irreversible} protected template), varies between 5\% FMR, 10\% FMR, and 20\% FMR (our original choice).  Note that ISR and FNMR were computed at our strictest match threshold, set at 0.01\% FMR.  Due to space constraints, we report results only on the largest dataset, Multi-PIE, and only for overlap = 3, since this was the configuration recommended in Section \ref{subsec:ks_accuracy} as it provides the best accuracy/irreversibility trade-off (in general).  However, the same effects can be observed on the other two datasets (SOTERIA and iCarB-Face) and across all overlaps, which can be confirmed by reproducing our experiments with the help of our open-source code.  

\begin{table}[!h]
\renewcommand{\arraystretch}{1.2}
\caption{Effects of varying the key selection threshold, used to define a ``good'' key, on the irreversibility (ISR), accuracy (FNMR), and unlinkability ($D_{\leftrightarrow}^{\mathit{sys}}$) properties of the resulting PolyProtected iResNet100 and EdgeFace templates (using overlap = 3).} \label{tab:ks_thresh} 
\centering
\begin{tabular}{|c|c|c|c|c|}
\hline
\multirow{2}{*}{Model} & \multirow{2}{*}{Property} & \multicolumn{3}{c|}{Key selection threshold, at:} \\
\cline{3-5}
 & & 5\% FMR & 10\% FMR & 20\% FMR \\
\hline
\multirow{3}{*}{iResNet100} & ISR (\%) & 0.24 & 0.03 & 0.03 \\
 \cline{2-5}
 & FNMR (\%) & 2.68 & 2.52 & 2.96 \\
 \cline{2-5}
 & $D_{\leftrightarrow}^{\mathit{sys}}$ & 0.23 & 0.23 & 0.23 \\
\hline
\multirow{3}{*}{EdgeFace} & ISR (\%) & 0.24 & 0.15 & 0.03 \\
 \cline{2-5}
 & FNMR (\%) & 2.98 & 3.40 & 3.21 \\
 \cline{2-5}
 & $D_{\leftrightarrow}^{\mathit{sys}}$ & 0.23 & 0.23 & 0.23 \\
\hline
\end{tabular}
\end{table} 

From Table \ref{tab:ks_thresh}, it is clear that changing the threshold used to define a successful inversion in the key selection algorithm has the greatest effect on the \textit{irreversibility} of the resulting protected templates.  Most notably, we see that increasing the key selection threshold from 5\% to 20\% FMR incurs an \textit{eight-fold reduction} in the ISR, for both iResNet100 and EdgeFace protected templates.  On the other hand, although we see a slight increase in the FNMR as the threshold increases, the change is not as dramatic as for the ISR, and for unlinkability there is no perceptible difference across the three thresholds.  It is for these reasons that we opted for the largest considered key selection threshold, at 20\% FMR, since the result is an appreciable improvement in the ISR of the protected templates with minimal impact on the accuracy and unlinkability.  Note that we also experimented with larger thresholds (up to 50\% FMR), but since the ISR is already close to 0 with the 20\% threshold, and larger thresholds incurred a longer key selection time without a significant improvement in the ISR, we concluded that a key selection threshold at 20\% FMR was sufficient.

\subsection{Computational cost of key selection}
\label{subsec:ks_cost}

This section estimates the computational cost of our proposed key selection algorithm, in terms of the \textit{average time per subject} to select the first ``good'' key (Table \ref{tab:ks_cost}) and the \textit{average number of failed keys per subject} before this good key is found (Table \ref{tab:ks_num_keys}).  Due to space restrictions, and for increased legibility, these tables show results for only the largest of our three datasets, Multi-PIE, but the relevant trends are the same on the other datasets.  Since the time and effort to select a key depends on the size of the protected template, which is determined by PolyProtect's ``overlap'' parameter, we report the computational cost for the smallest overlap of 0 and the largest (recommended) overlap of 3.  Tables \ref{tab:ks_cost} and \ref{tab:ks_num_keys} also show how the computational cost of key selection is affected by the key selection thresholds analysed in Section \ref{subsec:ks_thresh}.

\begin{table}[!h]
\renewcommand{\arraystretch}{1.2}
\caption{Average time to find a ``good'' subject-specific key for protecting iResNet100 and EdgeFace embeddings via PolyProtect with different overlaps, using our key selection algorithm with three FMR-based thresholds. \label{tab:ks_cost}} 
\centering
\begin{tabular}{|c|c|c|c|c|}
\hline
\multirow{2}{*}{Model} & \multirow{2}{*}{Overlap} & \multicolumn{3}{c|}{Average time to select key (sec), at:} \\
\cline{3-5}
 & & 5\% FMR & 10\% FMR & 20\% FMR \\
\hline
\multirow{2}{*}{iResNet100} & 0 & 143.31 & 210.70 & 400.04 \\
 \cline{2-5}
 & 3 & 796.09 & 1,057.59 & 1,819.50 \\
\hline
\multirow{2}{*}{EdgeFace} & 0 & 146.00 & 226.71 & 385.06 \\
 \cline{2-5}
 & 3 & 737.48 & 1,023.08 & 1,530.63 \\
\hline
\end{tabular}
\end{table} 

\begin{table}[!h]
\renewcommand{\arraystretch}{1.2}
\caption{Average number of failed keys per subject before finding a ``good'' key, for our key selection algorithm with three FMR-based thresholds, when the keys are used to protect iResNet100 and EdgeFace embeddings via PolyProtect with different overlaps. \label{tab:ks_num_keys}} 
\centering
\begin{tabular}{|c|c|c|c|c|}
\hline
\multirow{2}{*}{Model} & \multirow{2}{*}{Overlap} & \multicolumn{3}{c|}{Average no. failed keys, at:} \\
\cline{3-5}
 & & 5\% FMR & 10\% FMR & 20\% FMR \\
\hline
\multirow{2}{*}{iResNet100} & 0 & 90.62 & 132.88 & 249.12 \\
 \cline{2-5}
 & 3 & 188.25 & 255.53 & 441.58 \\
\hline
\multirow{2}{*}{EdgeFace} & 0 & 102.12 & 153.66 & 267.11 \\
 \cline{2-5}
 & 3 & 227.75 & 317.36 & 477.20 \\
\hline
\end{tabular}
\end{table} 

Tables \ref{tab:ks_cost} and \ref{tab:ks_num_keys} shows two important trends.  Firstly, the amount of time to find a ``good'' key using our key selection algorithm, as well as the number of failed keys before this good key is found, increases for larger PolyProtect overlaps.  This is because larger overlaps result in higher-dimensional protected templates (see Table \ref{tab:mappings}), which contain more information about the original face embeddings and are thus easier to invert.  Therefore, it takes longer to find a key (corresponding to the $C$ and $E$ parameters of the PolyProtect polynomials) that would produce an \textit{irreversible} protected template.  On the other hand, since protected templates generated using smaller overlaps are already fundamentally more difficult to invert due to the greater amount of information lost in the ``embedding'' $\rightarrow$ ``PolyProtected template'' transformation, it is easier to find a key that would produce an irreversible protected template.

The second notable trend from Tables \ref{tab:ks_cost} and \ref{tab:ks_num_keys} is that, the larger the threshold used to define a successful inversion in the key selection algorithm, the longer it takes to find a ``good'' key and the more ``bad'' keys we need to sift through to find this key.  For example, Table \ref{tab:ks_cost} shows that the time taken to select a key when using the 20\% FMR threshold is more than twice the time taken at the 5\% FMR threshold, and the same observation can be made for the number of failed keys in Table \ref{tab:ks_num_keys}.  This is due to the fact that the larger threshold imposes a stricter criterion on what it means for a protected template to be \textit{irreversible}, i.e., a larger threshold necessitates a greater cosine distance between the inverted template and the original face embedding, thereby making it more difficult to recover an embedding that matches the original, which translates to a \textit{more irreversible} protected template.  So, it makes sense that using a larger key selection threshold would make it more difficult to find a key that would allow for such a high degree of irreversibility, which results in a larger number of ``bad'' keys and thus a longer key selection time.

Overall, the results in Tables \ref{tab:ks_cost} and \ref{tab:ks_num_keys} raise the following important question: Would the computational cost of the proposed key selection algorithm present a barrier to its adoption in practice?  For example, when selecting keys to protect either iResNet100 or EdgeFace embeddings via PolyProtect using an overlap of 3, Table \ref{tab:ks_cost} suggests that it would take between about 12 and 30 minutes (depending on the chosen key selection threshold) to select a ``good'' key for each person that is to be enrolled in the face recognition system.  Considering the significantly enhanced irreversibility provided by our key selection algorithm, compared to when the keys are selected randomly (see Table \ref{tab:random_vs_ks}), the aforementioned key selection times may be seen as an acceptable trade-off if the enrollment need not be conducted in real time.  This is especially true for the shorter wait time of approximately 12 minutes when using the smaller key selection threshold at 5\% FMR, where the ISR is still very low (below 1\%, as shown in Table \ref{tab:ks_thresh}) and may thus be considered reasonable despite being higher than what is achievable with the larger threshold at 20\% FMR (for which the key selection computational cost is much higher).  However, if enrollment must be practically instantaneous, and especially if the subjects need to be enrolled sequentially with as little inter-subject delay as possible, waiting even 12 minutes per subject for key selection would likely be unacceptable.  One solution to speed up key selection could be to check multiple keys in parallel, as opposed to iterating through the key space sequentially (which was our conservative approach).  Another solution could be to further reduce the key selection threshold (e.g., to 1\% FMR), which would incur greater ISR sacrifice but which may present a reasonable trade-off for a shorter key selection time.  Potentially the best solution could be to explore the key space offline to find a set of ``good'' keys ahead of time, then randomly assign these keys to subjects during enrollment.  This would translate to zero wait time for key selection \textit{during} enrollment, which would render the online computational cost of the key selection algorithm more practically acceptable.  The feasibility of this last approach is examined and proved in our follow-up paper \cite{khm26}, where we additionally investigate the selection of \textit{system}-specific (instead of \textit{subject}-specific) PolyProtect keys.

\section{Conclusions and Future Work}
\label{sec:conclusion}

The main aim of this work was to improve the irreversibility of protected face templates (embeddings), generated using the PolyProtect BTP method.  This was motivated by our finding that PolyProtected templates are easier to invert using a numerical solver based on cosine distance, compared to using a solver based on Euclidean distance (when the definition of a successful inversion is likewise based on cosine distance).  To make PolyProtected templates harder to invert with this new solver, we proposed a key selection algorithm, which tries to choose ``keys'' (i.e., coefficients and exponents of the PolyProtect polynomial) that are more likely to generate ``irreversible'' protected templates, compared to when these keys are selected purely randomly.  Our experiments showed that this algorithm can significantly improve the irreversibility of PolyProtected templates, and moreover that it is able to approximately equalise (concentrate around a small ISR) the irreversibility of PolyProtected templates generated using different ``overlap'' parameters.  This allows for more effective control of the irreversibility versus accuracy trade-off, known to exist across different overlaps.  To ensure that accuracy in the PolyProtected domain is as high as possible, we also suggested normalizing the face embeddings prior to transforming them using PolyProtect, so as to avoid potentially high intra-class variance caused by the (large) range in which the embedding elements initially lie. 

There are two main ideas for next steps.  Firstly, we aim to investigate the ``suitable key'' space of our key selection algorithm, to see if there are any patterns in the $C$ and $E$ parameters that are selected versus those that are rejected.  This would help us: (i) estimate the number of possible ``good'' keys, which would give us an idea of how many different PolyProtected templates it is possible to generate when our key selection algorithm is used (related to the renewability/unlinkability criterion of BTP methods); (ii) gain insight into how the key selection algorithm could be sped up to find ``good'' keys within a more practically acceptable time; and (iii) determine whether our key selection algorithm could be used to find suitable \textit{system}-specific keys (same key for all subjects enrolled in the same face recognition system), in addition to the \textit{subject}-specific keys (different keys for different subjects, which is the original setting for PolyProtect) explored in this paper.  Secondly, we plan to improve our key selection algorithm to increase accuracy in the protected domain.  One idea is to incorporate an accuracy check into the selection procedure -- at the moment, we select keys that produce ``irreversible'' PolyProtected templates, based on a threshold that determines a successful inversion, so we could imagine adding a similar check for accuracy.  This would help ensure an even better balance between the accuracy and irreversibility of PolyProtected templates.

\section*{Acknowledgments}
This work was funded by the Innosuisse project ``PRiMEAiD: Privacy-pReserving bioMetric idEntification for humAnitarian aid Distribution'' (Number: 116.346 IP-ICT).

\bibliographystyle{IEEEtran}
\bibliography{bibliography}

@incollection{bg26,
	title={{Biometric Template Protection: Why and How}},
	author={C. Busch and M. Gomez-Barrero and H. {Otroshi Shahreza}},
	editor={V. {Krivoku{\'c}a Hahn} and M. Gomez-Barrero and A. Ross and S. Marcel},
	booktitle={Handbook of Biometric Template Protection: Motivation, Methods and Metrics},
	pages={3--29},
	year={2026},
	publisher={Springer}
}

@article{z16,
	title={Inverting face embeddings with convolutional neural networks},
	author={A. Zhmoginov and M. Sandler},
	journal={arXiv preprint arXiv:1606.04189},
	year={2016}
}

@inproceedings{c17,  
	author={F. Cole and D. Belanger and D. Krishnan and A. Sarna and I. Mosseri and W. T. Freeman},  
	booktitle={2017 IEEE Conference on Computer Vision and Pattern Recognition (CVPR)},   
	title={{Synthesizing Normalized Faces from Facial Identity Features}},   
	year={2017},   
	pages={3386-3395},  
	doi={10.1109/CVPR.2017.361}
}

@article{m19,  
	author={G. Mai and K. Cao and P. C. Yuen and A. K. Jain},  
	journal={IEEE Transactions on Pattern Analysis and Machine Intelligence},   
	title={{On the Reconstruction of Face Images from Deep Face Templates}},   
	year={2019},  
	volume={41},  
	number={5},  
	pages={1188-1202},  
	doi={10.1109/TPAMI.2018.2827389}
}

@inproceedings{s22,
	title={{Face Reconstruction from Deep Facial Embeddings using a Convolutional Neural Network}},
	author={H. {Otroshi Shahreza} and V. {Krivoku{\'c}a Hahn}  and S. Marcel},
	booktitle={2022 IEEE International Conference on Image Processing (ICIP)},
	pages={1211--1215},
	year={2022},
	organization={IEEE}
}

@article{f20,
	title={{De-anonymizing Facial Recognition Embeddings}},
  	author={I. F{\'a}bi{\'a}n and G. G. Guly{\'a}s},
  	journal={Infocommunications Journal},
  	volume={12},
  	number={2},
  	pages={50--56},
  	year={2020}
}

@article{t20,
  	title={{Beyond Identity: What Information Is Stored in Biometric Face Templates?}},
  	author={P. Terh{\"o}rst and D. F{\"a}hrmann and N. Damer and F. Kirchbuchner and A. Kuijper},
  	journal={arXiv preprint arXiv:2009.09918},
  	year={2020}
}

@article{h22,
	title={{Biometric Template Protection for Neural-Network-Based Face Recognition Systems: A Survey of Methods and Evaluation Techniques}},
	author={V. {Krivoku{\'c}a Hahn} and S. Marcel},
	journal={IEEE Transactions on Information Forensics and Security},
	volume={18},
	pages={639--666},
	year={2022},
	publisher={IEEE}
}

@book{h26,
	title={{Handbook of Biometric Template Protection: Motivation, Methods and Metrics}},
	editor={V. {Krivoku{\'c}a Hahn} and M. Gomez-Barrero and A. Ross and S. Marcel},
	year={2026},
	publisher={Springer}
}

@incollection{d26,
	title={{Feature Transformation-Based Biometrics Template Protection}},
	author={X. Dong and A. B. J. Teoh},
	editor={V. {Krivoku{\'c}a Hahn} and M. Gomez-Barrero and A. Ross and S. Marcel},
	booktitle={Handbook of Biometric Template Protection: Motivation, Methods and Metrics},
	pages={79--106},
	year={2026},
	publisher={Springer}
}

@inproceedings{d19,  
	author={X. Dong and K. Wong and Z. Jin and J. -L. Dugelay},  
	booktitle={2019 7th International Workshop on Biometrics and Forensics (IWBF)},   
	title={{A Cancellable Face Template Scheme Based on Nonlinear Multi-Dimension Spectral Hashing}},   
	year={2019},   
	pages={1-6},  
	doi={10.1109/IWBF.2019.8739179}
}

@article{kh22,
	author={V. {Krivoku{\'c}a Hahn} and S. Marcel},
	journal={IEEE Transactions on Biometrics, Behavior, and Identity Science}, 
	title={{Towards Protecting Face Embeddings in Mobile Face Verification Scenarios}}, 
	year={2022},
	volume={4},
	number={1},
	pages={117-134},
	doi={10.1109/TBIOM.2022.3140472}
}

@incollection{r26,
	title={{Biometric Cryptosystems}},
	author={C. Rathgeb and V. Fohr and B. Tams},
	editor={V. {Krivoku{\'c}a Hahn} and M. Gomez-Barrero and A. Ross and S. Marcel},
	booktitle={Handbook of Biometric Template Protection: Motivation, Methods and Metrics},
	pages={107--132},
	year={2026},
	publisher={Springer}
}

@inproceedings{g19,  
	author={B. P. Gilkalaye and A. Rattani and R. Derakhshani},  
	booktitle={2019 7th International Workshop on Biometrics and Forensics (IWBF)},   
	title={{Euclidean-Distance Based Fuzzy Commitment Scheme for Biometric Template Security}},   
	year={2019},
	pages={1-6},
	doi={10.1109/IWBF.2019.8739177}
}

@article{r22,
	title={Deep face fuzzy vault: Implementation and performance},
	author={C. Rathgeb and J. Merkle and J. Scholz and B. Tams and V. Nesterowicz},
	journal={Computers \& Security},
	volume={113},
	pages={102539},
	year={2022},
	publisher={Elsevier}
}

@incollection{b26,
	title={{Homomorphic Encryption for Biometric Template Protection}},
	author={V. N. Boddeti},
	editor={V. {Krivoku{\'c}a Hahn} and M. Gomez-Barrero and A. Ross and S. Marcel},
	booktitle={Handbook of Biometric Template Protection: Motivation, Methods and Metrics},
	pages={133--170},
	year={2026},
	publisher={Springer}
}

@article{m17,  
	author={Y. Ma and L. Wu and X. Gu and J. He and Z. Yang},  
	journal={IEEE Access},   
	title={{A Secure Face-Verification Scheme Based on Homomorphic Encryption and Deep Neural Networks}},   
	year={2017},  
	volume={5},   
	pages={16532-16538},  
	doi={10.1109/ACCESS.2017.2737544}
}

@inproceedings{b18,  
	author={V. N. Boddeti},  
	booktitle={2018 IEEE 9th International Conference on Biometrics Theory, Applications and Systems (BTAS)},   
	title={{Secure Face Matching Using Fully Homomorphic Encryption}},   
	year={2018},   
	pages={1-10},  
	doi={10.1109/BTAS.2018.8698601}
}

@article{e22,
	author={J. J. Engelsma and A. K. Jain and V. N. Boddeti},
	journal={IEEE Transactions on Biometrics, Behavior, and Identity Science}, 
	title={{HERS: Homomorphically Encrypted Representation Search}}, 
	year={2022},
	volume={4},
	number={3},
	pages={349-360},
	doi={10.1109/TBIOM.2021.3139866}
}

@incollection{kh26,
	title={{Using Neural Networks to Learn Biometric Template Protection}},
	author={V. {Krivoku{\'c}a Hahn} and M. Valenti and V. Talreja and N. Nasrabadi and T. S. Ng and A. B. J. Teoh},
	editor={V. {Krivoku{\'c}a Hahn} and M. Gomez-Barrero and A. Ross and S. Marcel},
	booktitle={Handbook of Biometric Template Protection: Motivation, Methods and Metrics},
	pages={171--201},
	year={2026},
	publisher={Springer}
}

@inproceedings{p16,  
	author={R. K. Pandey and Y. Zhou and B. U. Kota and V Govindaraju},  
	booktitle={2016 IEEE Conference on Computer Vision and Pattern Recognition Workshops (CVPRW)},   
	title={{Deep Secure Encoding for Face Template Protection}},   
	year={2016},
	pages={77-83},
	doi={10.1109/CVPRW.2016.17}
}

@inproceedings{j18,  
	author={A. K. Jindal and S. Chalamala and S. K. Jami},  
	booktitle={2018 IEEE/CVF Conference on Computer Vision and Pattern Recognition Workshops (CVPRW)},   
	title={{Face Template Protection Using Deep Convolutional Neural Network}},   
	year={2018},
	pages={575-5758},
	doi={10.1109/CVPRW.2018.00087}
}

@inproceedings{t19,  
	author={V. Talreja and M. C. Valenti and N. M. Nasrabadi},  
	booktitle={2019 IEEE 10th International Conference on Biometrics Theory, Applications and Systems (BTAS)},   
	title={{Zero-Shot Deep Hashing and Neural Network Based Error Correction for Face Template Protection}},   
	year={2019},
	pages={1-10},
	doi={10.1109/BTAS46853.2019.9185979}
}

@article{p21,  
	author={J. R. Pinto and M. V. Correia and J. S. Cardoso},  
	journal={IEEE Transactions on Biometrics, Behavior, and Identity Science},   
	title={{Secure Triplet Loss: Achieving Cancelability and Non-Linkability in End-to-End Deep Biometrics}},   
	year={2021},
	volume={3},
	number={2},
	pages={180-189},
	doi={10.1109/TBIOM.2020.3046620}
}

@article{m21,  
	author={G. Mai and K. Cao and X. Lan and P. C. Yuen},  
	journal={IEEE Transactions on Information Forensics and Security},   
	title={{SecureFace: Face Template Protection}},   
	year={2021},  
	volume={16},   
	pages={262-277},  
	doi={10.1109/TIFS.2020.3009590}
}

@inproceedings{s25,
	title={{Securing Face and Fingerprint Templates in Humanitarian Biometric Systems}},
	author={G. Stragapede and S. Merrick and V. {Krivoku{\'c}a Hahn} and J. Sukaitis and V. {Graf Narbel}},
	booktitle={2025 IEEE International Joint Conference on Biometrics (IJCB)},
	pages={1--10},
	year={2025},
	organization={IEEE}
}

@article{g24,
	title={Edgeface: Efficient face recognition model for edge devices},
	author={A. George and C. Ecabert and H. Otroshi Shahreza and K. Kotwal and S. Marcel},
	journal={IEEE Transactions on Biometrics, Behavior, and Identity Science},
	volume={6},
	number={2},
	pages={158--168},
	year={2024},
	publisher={IEEE}
}

@inproceedings{h16,
	title = {{Deep Residual Learning for Image Recognition}},
	booktitle = {Proceedings of the {{IEEE Conference}} on {{Computer Vision}} and {{Pattern Recognition}}},
	author = {K. He and X. Zhang and S. Ren and J. Sun},
	year = {2016},
	pages = {770--778}
}

@inproceedings{dg19,
	title={{ArcFace: Additive Angular Margin Loss for Deep Face Recognition}},
	author={J. Deng and J. Guo and N. Xue and S. Zafeiriou},
	booktitle={Proceedings of the IEEE/CVF Conference on Computer Vision and Pattern Recognition},
	pages={4690--4699},
	year={2019}}

@inproceedings{s15,
	title={{FaceNet: A Unified Embedding for Face Recognition and Clustering}},
	author={F. Schroff and D. Kalenichenko and J. Philbin},
	booktitle={Proceedings of the IEEE Conference on Computer Vision and Pattern Recognition},
	pages={815--823},
	year={2015}
}

@article{g10,
	title={{Multi-PIE}},
	author={R. Gross and I. Matthews and J. Cohn and T. Kanade and S. Baker},
	journal={Image and Vision Computing},
	volume={28},
	number={5},
	pages={807--813},
	year={2010},
	publisher={Elsevier}}

@inproceedings{r24,
	title={{A Novel and Responsible Dataset for Face Presentation Attack Detection on Mobile Devices}},
	author={N. Ramoly and A. Komaty and V. Krivoku{\'c}a Hahn and L. Younes and A.-M. Awal and S. Marcel},
	booktitle={2024 IEEE International Joint Conference on Biometrics (IJCB)},
	pages={1--9},
	year={2024},
	organization={IEEE}
}

@article{kh24,
	title={{in-Car Biometrics (iCarB) Datasets for Driver Recognition: Face, Fingerprint, and Voice}},
	author={V. {Krivoku{\'c}a Hahn} and J. Maceiras and A. Komaty and P. Abbet and S. Marcel},
	journal={arXiv preprint arXiv:2411.17305},
	year={2024}
}

@article{gb18,  
  author={Gomez-Barrero, M. and Galbally, J. and Rathgeb, C. and Busch, C.},  
  journal={IEEE Transactions on Information Forensics and Security},   
  title={{General Framework to Evaluate Unlinkability in Biometric Template Protection Systems}},   
  year={2018},  
  volume={13},  
  number={6},  
  pages={1406-1420},  
  doi={10.1109/TIFS.2017.2788000}}

@article{khm26,
	title={{Is it Possible to Generate Irreversible PolyProtected Templates from Face Embeddings using System-Specific Keys?}},
	author={V. {Krivoku{\'c}a Hahn} and J. Maceiras and S. Marcel},
	journal={arXiv preprint arXiv:2610.01385},
	year={2026}
}

\vfill

\end{document}